\documentclass{article}

\usepackage[main, final]{neurips_2026}

\usepackage[utf8]{inputenc} 
\usepackage[T1]{fontenc}    
\usepackage{hyperref}       
\usepackage{url}            
\usepackage{booktabs}       
\usepackage{amsfonts}       
\usepackage{nicefrac}       
\usepackage{microtype}      
\usepackage[table]{xcolor}  
\usepackage{graphicx}
\usepackage{tabularx}
\usepackage{multirow}
\usepackage{amsmath}
\usepackage{caption}
\usepackage{subcaption}

\newlength{\choppedwidth}
\newlength{\captiongap}
\newlength{\imagegap}
\newcommand{\insertchoppedlq}[1]{%
  \leavevmode
  \hbox{%
    \vbox{%
      \hbox{\includegraphics[width=0.49\choppedwidth]{figures/chopped/#1/LQ/1.jpg}}%
      \hbox{\includegraphics[width=0.49\choppedwidth]{figures/chopped/#1/LQ/4.jpg}}%
    }%
    \hspace{0.5\imagegap}%
    \vbox{%
      \hbox{\includegraphics[width=\choppedwidth]{figures/chopped/#1/LQ/16.jpg}}%
    }%
    \hspace{0.5\imagegap}%
    \vbox{%
      \hbox{\includegraphics[width=0.49\choppedwidth]{figures/chopped/#1/LQ/64.jpg}}%
      \hbox{\includegraphics[width=0.49\choppedwidth]{figures/chopped/#1/LQ/256.jpg}}%
    }%
  }%
}

\newcommand{\singleimage}[3][\choppedwidth]{%
  \vbox{%
    \hbox to #1{\hfil\includegraphics[width=#1]{#2}\hfil}%
    \vskip\captiongap%
    \hbox to #1{\hfil #3\hfil}%
  }%
}

\makeatletter
\newcommand{\insertcomparison}{%
  \@ifstar{\insertcomparison@show}{\insertcomparison@hide}%
}

\newcommand{\insertimageonly}[2][\choppedwidth]{%
  \vbox{%
    \hbox to #1{\hfil\includegraphics[width=#1]{#2}\hfil}%
  }%
}

\newcommand{\insertcomparison@hide}[1]{%
  \tiny
  \par\centering
  \hbox{%
    \vbox{%
      \hbox{\insertchoppedlq{#1}}%
    }%
    \hspace{\imagegap}%
    \insertimageonly{figures/chopped/#1/GT.jpg}%
    \hspace{\imagegap}%
    \insertimageonly{figures/chopped/#1/Burstormer.jpg}%
    \hspace{\imagegap}%
    \insertimageonly{figures/chopped/#1/HDRFlow.jpg}%
    \hspace{\imagegap}%
    \insertimageonly{figures/chopped/#1/BracketIRE.jpg}%
    \hspace{\imagegap}%
    \insertimageonly{figures/chopped/#1/FlickerFormer.jpg}%
    \hspace{\imagegap}%
    \insertimageonly{figures/chopped/#1/OURS.jpg}%
  }%
  \par
}

\newcommand{\insertcomparison@show}[1]{%
  \tiny
  \par\centering
  \hbox{%
    \vbox{%
      \hbox{\insertchoppedlq{#1}}%
      \vskip\captiongap%
      \hbox to 2\choppedwidth{\hfil LQ $\{ ^1_2 \space 3 \space ^4_5 \}$ \hfil}%
    }%
    \hspace{\imagegap}%
    \singleimage{figures/chopped/#1/GT.jpg}{GT}%
    \hspace{\imagegap}%
    \singleimage{figures/chopped/#1/Burstormer.jpg}{Burstormer~\cite{dudhane2023burstormer}}%
    \hspace{\imagegap}%
    \singleimage{figures/chopped/#1/HDRFlow.jpg}{HDRFlow~\cite{xu2024hdrflow}}%
    \hspace{\imagegap}%
    \singleimage{figures/chopped/#1/BracketIRE.jpg}{TMRNet~\cite{Zhang2025BracketIRE}}%
    \hspace{\imagegap}%
    \singleimage{figures/chopped/#1/FlickerFormer.jpg}{FlickerFormer~\cite{qu2026ittakestwo}}%
    \hspace{\imagegap}%
    \singleimage{figures/chopped/#1/OURS.jpg}{BRACE (OURS)}%
  }%
  \par
}
\makeatother

\newcommand{\insertmethodlegend}{%
  \tiny
  \par\centering
  \hbox{%
    \hbox to 2\choppedwidth{\hfil LQ $[ \ ^1_2 \ 3 \ ^4_5 \ ]$\hfil}%
    \hspace{\imagegap}%
    \hbox to \choppedwidth{\hfil GT\hfil}%
    \hspace{\imagegap}%
    \hbox to \choppedwidth{\hfil Burstormer~\cite{dudhane2023burstormer}\hfil}%
    \hspace{\imagegap}%
    \hbox to \choppedwidth{\hfil HDRFlow~\cite{xu2024hdrflow}\hfil}%
    \hspace{\imagegap}%
    \hbox to \choppedwidth{\hfil TMRNet~\cite{Zhang2025BracketIRE}\hfil}%
    \hspace{\imagegap}%
    \hbox to \choppedwidth{\hfil FlickerFormer~\cite{qu2026ittakestwo}\hfil}%
    \hspace{\imagegap}%
    \hbox to \choppedwidth{\hfil BRACE (OURS)\hfil}%
  }%
  \par
}

\newcommand{\best}[1]{\textcolor{red}{\textbf{#1}}}
\newcommand{\secondbest}[1]{\textcolor{blue}{#1}}

\title{Bricker to BRACE: A Bracket Exposure RAW Dataset and Restoration Model for Flicker-Banding}

\author{%
	\textbf{Zihan Zhou}$^{1}$\thanks{Equal contribution.},\enspace
    \textbf{Libo Zhu}$^{1}$\footnotemark[1],\enspace
    \textbf{Jue Gong}$^{1}$,\enspace
    \textbf{Zhiyi Zhou}$^{1}$,\\
    \textbf{Jiezhang Cao}$^{1}$\thanks{Corresponding authors: Yulun Zhang (yulun100@gmail.com) and Jiezhang Cao (caojiezhang@sjtu.edu.cn).},\enspace
    \textbf{Yong Guo}$^{1}$,\enspace
    \textbf{Yulun Zhang}$^{1}$\footnotemark[2]  \\
	$^{1}$Shanghai Jiao Tong University
}

\begin{document}

\maketitle

\setlength{\abovedisplayskip}{2pt}
\setlength{\belowdisplayskip}{2pt}

\vspace{-1em}

\begin{abstract}
  \emph{Flicker-Banding} (FB) arises from temporal aliasing between a camera's rolling shutter and a display's brightness modulation, degrading screen-captured image readability with color shifts and jagged patterns. Existing single-frame methods with simplified parametric stripe models cannot reliably distinguish these artifacts from genuine texture. To address this, we conduct a systematic analysis of complex FB morphologies and reveal their significant variation across exposure settings, motivating a multi-frame bracketed RAW restoration paradigm. We construct \textbf{Bricker}, a synthetic–real bracketed RAW dataset built via ray-tracing-based physical simulation and an automated multi-exposure capture tool. We further propose \textbf{BRACE}: Bracketed RAW Flicker-Banding Removal, a multi-frame restoration model that utilizes a frequency-aware banding prior and a multi-scale spatial cross-attention modulator (MSCAM) for cross-exposure spatial fusion. We also introduce the Stripe Frequency Consistency (SFC) metric to evaluate banding removal. Experiments demonstrate state-of-the-art performance on both synthetic and real benchmarks. Our dataset and code are available at: \href{https://github.com/ZZH-qwq/BRACE}{https://github.com/ZZH-qwq/BRACE}.
\end{abstract}

\section{Introduction}

\textbf{Flicker-Banding (FB) removal} aims to eliminate the diverse stripe artifacts that appear in screen-captured images, and is a challenging and practically important image restoration task~\cite{abe2020imperceptible}. Despite the widespread use of smartphones and continuous advances in mobile imaging, capturing emissive screens still produces characteristic degradations. Among these, Flicker-Banding is one of the most common and visually disruptive artifacts, and it further degrades the performance of downstream tasks such as high dynamic range (HDR) fusion and optical character recognition (OCR)~\cite{Zhu2025RIFLE,deegan2018ledflicker}.

FB stems from the temporal coupling between a camera's rolling shutter and the screen's brightness modulation scheme. Most smartphone cameras use CMOS sensors with rolling shutters~\cite{ieee2018p2020whitepaper}, while displays regulate brightness through time-varying mechanisms such as pulse-width modulation (PWM) or progressive scanning refresh~\cite{geffroy2006organic}. When the rolling shutter's exposure overlaps with the screen's temporal brightness variation, the invisible time-domain modulation is projected into the spatial domain, producing alternating bright and dark stripe artifacts in the captured image~\cite{Zhu2025RIFLE,Sumner2020ImatestFlicker}.

Existing image restoration methods based on single-frame input~\cite{liang2021swinir,sun2025pixel,gong2025osdhuman,gong2025haodiff} have achieved impressive results across a broad range of scenarios. However, single-frame restoration faces a more severe ill-posedness challenge: in the context of FB removal, when the screen content contains rich texture such as dense text or grid patterns, the model cannot reliably distinguish banding artifacts from the underlying content structure. In contrast, multi-frame restoration and exposure bracketing offer a more promising alternative~\cite{Hasinoff2016HDRPlus,Mildenhall2018KPN,Zhang2025BracketIRE}. Under different exposure settings, FB artifacts appear with distinct morphologies, contrasts, and intensities, while the screen content itself remains unchanged. This makes it substantially easier to decouple artifacts from true scene texture.

Exposure bracketing captures multiple frames of the same scene at different exposure time. It is commonly stored in RAW format to preserve the full sensor information, and has proven highly effective in HDR imaging, denoising, and deblurring~\cite{Zhang2025BracketIRE,Zhang2024NTIREBracket}. RAW data retains the linear sensor readout, which makes it easier to model how FB patterns vary across different exposures. However, the use of bracketed RAW capture for FB removal remains largely unexplored.

Current FB simulation methods~\cite{Zhu2025RIFLE,Zhu2026CLEAR} have made reasonable attempts at single-frame degradation modeling under simplified assumptions. However, these simulation methods cannot produce exposure-dependent stripe morphologies, nor can they reproduce the diverse complex FB patterns observed in real captures. Existing bracketed exposure datasets~\cite{Zhang2025BracketIRE} are designed for HDR content or low-light scenes and do not reflect the unique degradation characteristics of screen capture or the variation of stripe patterns across exposures. 
The absence of high-quality bracketed FB datasets, together with suitable simulation pipelines and real-world acquisition methods, constitutes the primary bottleneck for advancing multi-frame FB restoration research.

To address these challenges, we first analyze the formation of complex FB patterns and show their exposure-dependent variations in shape, intensity, and contrast, motivating exposure bracketing for complementary observations. Based on this analysis, we develop a ray-tracing-based screen imaging simulator and an automated multi-exposure capture tool, with which we construct \textbf{Bricker}, a paired synthetic and real \textbf{Br}acket exposure Fl\textbf{icker}-Banding dataset covering diverse display types, driving strategies, and capture conditions. We further propose \textbf{BRACE} for Bracketed RAW Flicker-Banding Removal, incorporating a Frequency-Aware Banding Prior (FABP) and a Multi-scale Spatial Cross-Attention Modulator (MSCAM) for frequency-aware and spatial cross-exposure fusion. In addition, we introduce a Stripe Frequency Consistency (SFC) metric to quantify banding removal. Experiments on synthetic and real benchmarks demonstrate state-of-the-art performance in restoration fidelity, perceptual quality, frequency consistency, and color accuracy.

In summary, our contributions are:

\begin{itemize}
  \item We systematically analyze the causes of color shifts, diamond-like patterns, compound periodic stripes, and other complex FB degradations, and characterize how they vary across exposure settings, establishing the necessity and feasibility of introducing complementary information through exposure bracketing for screen-captured images.

  \item We construct Bricker, a paired synthetic and real multi-frame RAW dataset covering diverse display types, driving strategies, and capture conditions based on a physics-driven ray-tracing simulation pipeline and an automated capture tool we develop.

  \item We propose BRACE, a multi-frame RAW restoration model that effectively exploits complementary cross-exposure information through a frequency-aware banding prior (FABP) and a multi-scale spatial cross-attention modulator (MSCAM). We also propose a Stripe Frequency Consistency (SFC) metric to quantitatively evaluate the degree of banding removal.

  \item Our method achieves state-of-the-art performance on both synthetic and real benchmarks, with leading results across restoration fidelity, perceptual metrics, frequency consistency, and color accuracy, indicating practical value for screen-capture restoration.
\end{itemize}

\begin{figure}
    \centering
    \includegraphics[width=\textwidth]{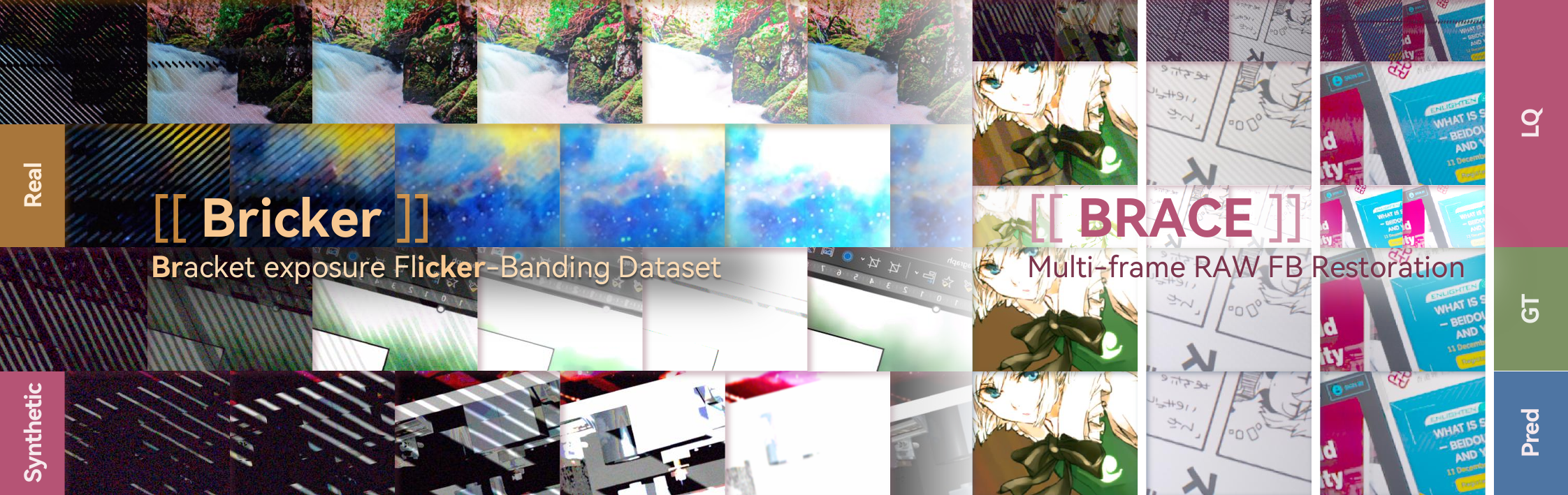}
    \vspace{-10pt}
    \caption{Overview of the \textbf{Bricker} dataset and \textbf{BRACE} model. \textbf{Left:} the Bricker dataset consists of paired synthetic and real images with diverse display types, driving strategies, and capture conditions. \textbf{Right}: the state-of-the-art Flicker-Banding Removal effect of our BRACE model.}
    \vspace{-12pt}
    \label{fig:title_graph}
\end{figure}

\newpage

\section{Related Work}
\vspace{-2mm}

\textbf{Flicker-Banding Removal.}
Flicker-Banding (FB) is a prominent structured degradation in screen-captured images. Unlike luminance stripes caused by ambient or AC-powered lighting flicker~\cite{qu2026ittakestwo,lin2023deflicker}, FB mainly results from temporal aliasing between the display's brightness modulation and the camera's rolling-shutter readout, affecting image readability and visual quality~\cite{Zhu2025RIFLE,deegan2018ledflicker}. Existing FB removal methods have explored explicit banding representations, including parameterized masks and gain maps~\cite{Zhu2025RIFLE,Zhu2026CLEAR}. These studies provide a foundation for screen-captured image restoration, yet FB patterns that depend on local screen brightness and image content remain underexplored.

\textbf{RAW Image Processing and Restoration.}
RAW image processing and restoration covers various computational photography tasks, including denoising, low-light enhancement, demoir\'eing, HDR synthesis, and color correction~\cite{Chen2018SID,Liu2019RawDenoising,Yang2023RealRawHDR,Xu2024RRID,Hasinoff2016HDRPlus}. Compared with ISP-processed sRGB images, RAW data better preserves the sensor's linear response and high-bit-depth measurements, avoiding demosaicking, tone mapping, and color correction~\cite{Hasinoff2016HDRPlus,Xu2024RRID,Yang2023RealRawHDR}. This makes the RAW domain more suitable for modeling noise, exposure variation, and sensor-sampling artifacts.

\textbf{Multi-Image Restoration.}
Multi-image restoration exploits complementary observations to mitigate the ill-posedness of single-image restoration. By acquisition strategy, existing methods include burst restoration with same-exposure frames~\cite{Bhat2021DeepBurstSR,Dudhane2022BurstIRE,dudhane2023burstormer,lee2025ntire,BurstDeflicker2025,qu2026ittakestwo}, dual-exposure restoration with noisy short-exposure and blurry long-exposure frames~\cite{Chang2022LSFNet,Zhang2022SelfIR}, and bracket restoration with multiple exposure levels~\cite{Zhang2025BracketIRE,Zhang2024NTIREBracket}. These paradigms impose cross-observation constraints on denoising, deblurring, and HDR synthesis, as do temporal priors in video restoration~\cite{chen2026diffst,chen2026dvface}.

\textbf{Bracket Image Restoration and Enhancement.}
Bracket image restoration and enhancement exploits complementary information from images captured under different exposure settings, making it a representative problem in multi-image restoration. Recent studies improve cross-exposure fusion through self-supervised pretraining~\cite{Zhang2025BracketIRE}, cross-frame interaction~\cite{Lin2024IREANet}, reference-guided feature aggregation~\cite{Xing2024RTIRE,Zhang2024NTIREBracket}, and frequency decomposition~\cite{Yang2024CRNet,Zhang2024NTIREBracket}. These designs enable bracket-based methods to address a range of restoration and enhancement tasks, including denoising, deblurring, HDR reconstruction, and super-resolution~\cite{Chang2022LSFNet,Zhang2022SelfIR,Zhang2025BracketIRE}.

\vspace{-2mm}
\section{Method}
\vspace{-3mm}

The complexity and diversity of FB phenomena, as systematically analyzed in Sec.~\ref{sec:problem}, make high-quality bracketed datasets essential for both training and evaluation. However, FB formation is tightly coupled to device and environmental factors, rendering large-scale, well-aligned real-world multi-frame data extremely difficult to obtain. Building on this analysis, we construct \textbf{Bricker}, a multi-frame RAW dataset with synthetic and real-world subsets covering diverse display types, driving strategies, and capture conditions (Sec.~\ref{sec:dataset}). To this end, we propose \textbf{BRACE}, a spatial--temporal model that fuses multi-frame spatial information to handle complex FB patterns, and a banding-sensitive metric, \textbf{SFC}, as detailed in Sec.~\ref{sec:model} and ~\ref{sec:sfc}.

\vspace{-1.5mm}
\subsection{Problem Formulation}
\label{sec:problem}
\vspace{-2mm}

\begin{figure}[t]
    \centering
    \includegraphics[width=\textwidth]{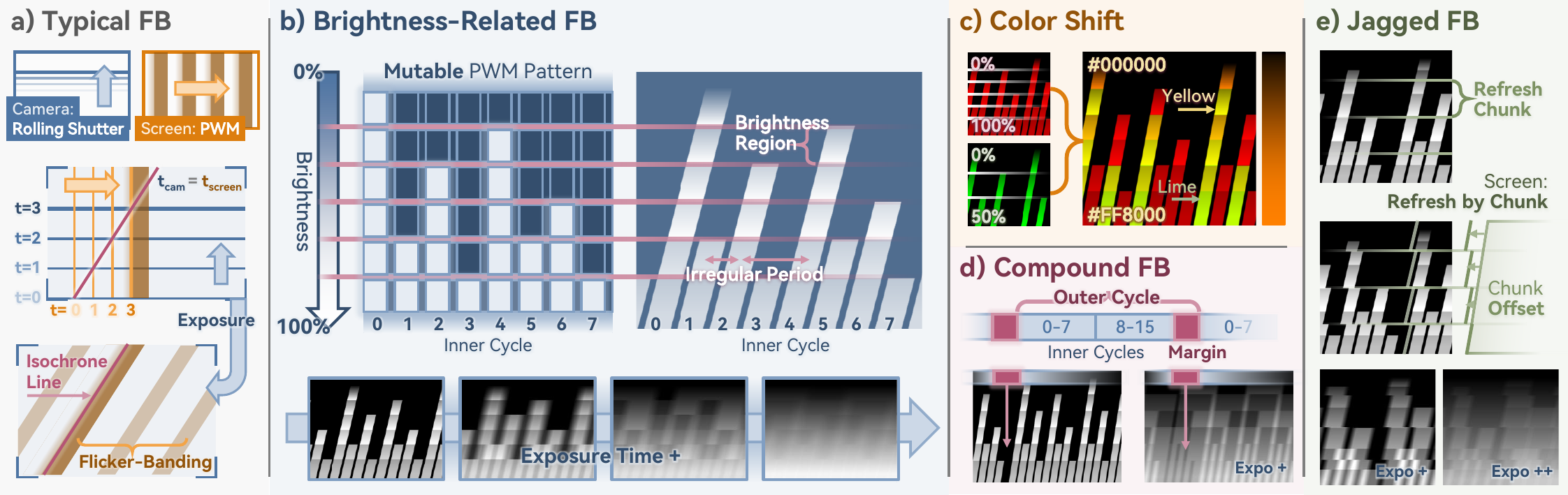}
    \vspace{-4.5mm}
    \caption{Complex FB morphologies commonly observed in real captures. The examples shown here are rendered using our physics-driven simulation pipeline.}
    \label{fig:complex_banding}
    \vspace{-5mm}
\end{figure}

FB originates from the temporal aliasing between a rolling-shutter sensor's sequential row exposure and the screen's time-varying brightness modulation via PWM or scanning refresh~\cite{Zhu2025RIFLE,xue2023screenid}. When these two timing processes overlap, temporal brightness variations on the display are projected into spatial striping patterns across the captured frame, as shown in Fig.~\ref{fig:complex_banding}.a. 

\textbf{The Brightness Dimming Mechanism.}
While existing simulation methods can produce banding with varied shapes and intensities through heuristic perturbation, they are not grounded in the physical display pipeline and therefore cannot capture artifacts that depend on local screen brightness~\cite{Zhu2025RIFLE,Zhu2026CLEAR}. The core issue is that a single global PWM cycle cannot provide each light-emitting unit with the full dynamic range required by the display content. To overcome this, real screens adopt region-specific dimming strategies: different image regions are assigned different PWM duty cycles or complex modulation sequences, thereby expanding the expressible brightness range (Fig.~\ref{fig:complex_banding}.b). This region-dependent modulation causes the period, contrast, and intensity of banding to vary non-uniformly across the frame according to local content. We briefly summarize the resulting FB morphologies below, with detailed analyses of color shift, compound FB, and jagged FB provided in Appendix~\ref{sec:app_complex_banding}.

\textbf{Color Shift.}
FB may appear with chromatic distortions, where the stripes are accompanied by spatially varying hue changes such as red, yellow, green, or unintended colors (Fig.~\ref{fig:complex_banding}.c).

\textbf{Compound FB.}
FB may contain multiple stripe scales simultaneously, forming composite patterns that mix coarse bands with finer periodic structures (Fig.~\ref{fig:complex_banding}.d).

\textbf{Jagged FB.}
FB may deviate from smooth parallel stripes and appear as stair-shaped, block-like, or diamond-shaped geometric patterns with abrupt phase discontinuities (Fig.~\ref{fig:complex_banding}.e).

\textbf{Significant Variation under Different Exposures.}
Since FB formation is tightly coupled to screen brightness modulation, its morphology, intensity, and contrast vary substantially with exposure settings~\cite{deegan2018ledflicker}. Short exposures produce high-contrast stripes, while long exposures can leave weak or invisible banding. Different brightness regions also exhibit distinct exposure-dependent changes in FB patterns and color shifts, which existing simulators fail to reproduce. Shutter times matched to the PWM period are often too long for practical capture, and the PWM frequency of a screen is generally unknown, so any nominal match still produces a beat; no practical single shutter time removes FB.

FB is therefore shaped by both exposure parameters and display driving strategies, and appears as complex, exposure-dependent artifacts that are difficult to separate from genuine texture. This non-aligned degradation also provides complementary information across bracketed exposures. These observations motivate a simulation system grounded in the physical imaging pipeline, together with a restoration model that exploits global frequency-direction priors and multi-frame fusion.

\vspace{-2.5mm}
\subsection{Dataset: Bricker}
\label{sec:dataset}
\vspace{-2.5mm}
\subsubsection{Synthetic dataset}
\vspace{-2mm}

Traditional approaches for generating bracketed-exposure datasets rely on stacking clear frames from HDR videos to simulate exposure and motion blur \cite{Zhang2025BracketIRE}. However, such methods cannot reproduce the unique characteristics of screen capture or the variation of FB patterns across bracketed exposures.

To generate a synthetic dataset with diverse and physically faithful FB morphologies, we construct a ray-tracing-based screen imaging simulation system that models the full physical imaging pipeline. 
Since FB artifacts concentrate in shorter-exposure frames, we define a 5-frame sequence indexed by $i \in \{1,\dots,5\}$, with $i=3$ as the reference frame. The exposure time of the $i$-th frame follows
\begin{equation}
    T_{i}^{\text{expo}} = 4^{\,i-3} \times T_{\text{ref}}^{\text{expo}},
\end{equation}
forming an underexposed--reference--overexposed progression. This design ensures that the reference frame provides texture information that is relatively clean and well-aligned with the ground truth, while the remaining frames supply FB degradations of differing morphology and intensity.

\begin{figure}[t]
    \centering
    \includegraphics[width=\textwidth]{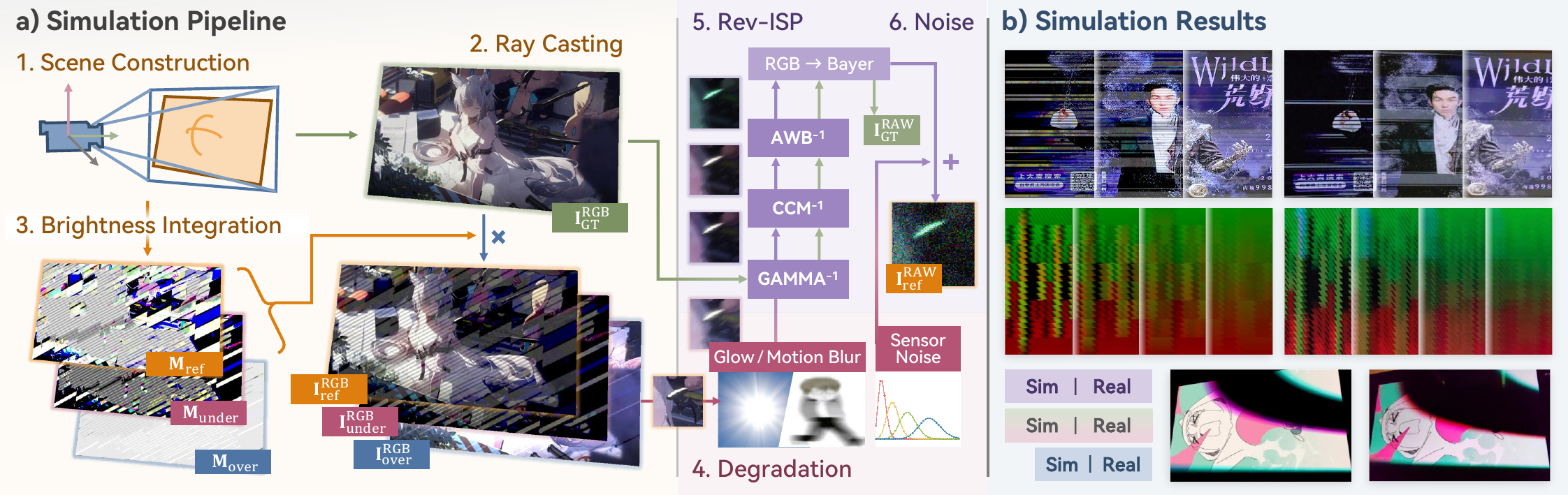}
    \vspace{-4mm}
    \caption{Overview of the proposed synthetic data generation pipeline. \textbf{a)} The physics-driven simulation pipeline. \textbf{b)} Comparison of FB patterns generated by our pipeline and real captures.}
    \vspace{-8mm}
    \label{fig:simulation_pipeline}
\end{figure}

The simulation pipeline, illustrated in Fig.~\ref{fig:simulation_pipeline}.a, consists of the following stages:

\textbf{Scene Construction.}
A virtual screen is placed in 3D space with randomized position and orientation (6DOF), with small inter-frame perturbations to emulate hand tremor. Camera and display parameters including rolling-shutter speed and display refresh scheme are also randomly sampled.

\textbf{Ray Casting.}
Based on the configured scene and camera parameters, we trace rays to compute, for each pixel, the intersection point on the screen and the corresponding timestamp within the exposure window, yielding the ground-truth RGB image $\mathbf{I}_{\text{GT}}^{\text{RGB}}$.

\textbf{Brightness Integration.}
Using the screen's refresh mechanism, we integrate the temporal brightness received by each pixel over its exposure, producing a brightness mask for each frame. Shown in Eq.~\ref{eq:brightness_integration}, $(x,y)$ indexes camera pixels and $(u,v)$ is the corresponding screen coordinate from ray intersection.

\begin{equation}
    \mathbf{M}_i(x,y) = \int_{t_i + y\,t_r}^{\,t_i + y\,t_r + T_i^{\text{expo}}} L\bigl(u(x,y), v(x,y), t\bigr)\, \mathrm{d}t,
    \label{eq:brightness_integration}
\end{equation}
where $t_i$ is the exposure start of the first row, $t_r$ is the row delay, and $L$ is the emitted brightness. The degraded RGB frame is then $\mathbf{I}_i^{\text{RGB}} = \mathbf{M}_i \odot \mathbf{I}_{\text{GT}}^{\text{RGB}}$, where $\odot$ denotes the element-wise product.

\textbf{Degradation.}
Frame-dependent artifacts such as motion blur and glow are applied to each frame.

\textbf{Reverse ISP.}
Both the ground-truth $\mathbf{I}_{\text{GT}}^{\text{RGB}}$ and degraded $\mathbf{I}_i^{\text{RGB}}$ are converted by the inverse ISP pipeline and then mosaicked into Bayer RAW. Since the ISP parameters are strongly device-dependent, we determine them from real DNG metadata (Appendix~\ref{sec:app_simulation}--\ref{sec:app_exif}). This stage produces the RAW-domain ground truth $\mathbf{I}_{\text{GT}}^{\text{RAW}}$ and the clean intermediate banding images $\mathbf{I}_{i,\text{clean}}^{\text{RAW}}$.

\textbf{Sensor Noise.}
Finally, simulated sensor noise is added to each $\mathbf{I}_{i,\text{clean}}^{\text{RAW}}$, yielding the final degraded RAW frames $\mathbf{I}_i^{\text{RAW}}$ for all bracketed exposure settings.

The masks $\mathbf{M}_i$ follow from the screen content, the camera-screen geometry, and the exposure time, so a predefined stripe template cannot reproduce them.
Fig.~\ref{fig:simulation_pipeline}.b compares the FB patterns generated by our pipeline with real captures, demonstrating the ability to reproduce complex morphologies as well as their variation across exposures under diverse real-world screen conditions.

\vspace{-2mm}
\subsubsection{Real-world dataset}
\vspace{-1.5mm}

Although our simulation pipeline strives for physical accuracy, several factors in real capture scenarios remain difficult to model exhaustively: non-ideal ambient reflections on the screen surface, lens distortion and aberration, sensor and display non-uniformity, slight defocus, and complex background interference. These collectively create a domain gap between synthetic and real distributions. A real-world dataset is therefore necessary to capture what simulation cannot, and to improve practical performance through fine-tuning under real capture conditions.

Existing mobile capture apps \cite{opencamera, proshot} only support fixed-ISO burst bracketing, so aligned data with varying exposures is hard to obtain. We therefore developed an Android-based tool that extends OpenCamera's Camera2 API to programmatically control exposure parameters and capture RAW sequences under arbitrary combinations (Appendix~\ref{sec:app_capture_tool}).

Using this tool, we collected a multi-frame bracketed RAW dataset spanning diverse screen content and capture conditions (Appendix~\ref{sec:app_statistics}). The tool's key strength is its fully automated parameter control and capture workflow, which substantially lowers collection effort and cost while ensuring quality and diversity; it may also benefit other tasks requiring multi-exposure data.

\begin{figure}
    \centering
    \includegraphics[width=\textwidth]{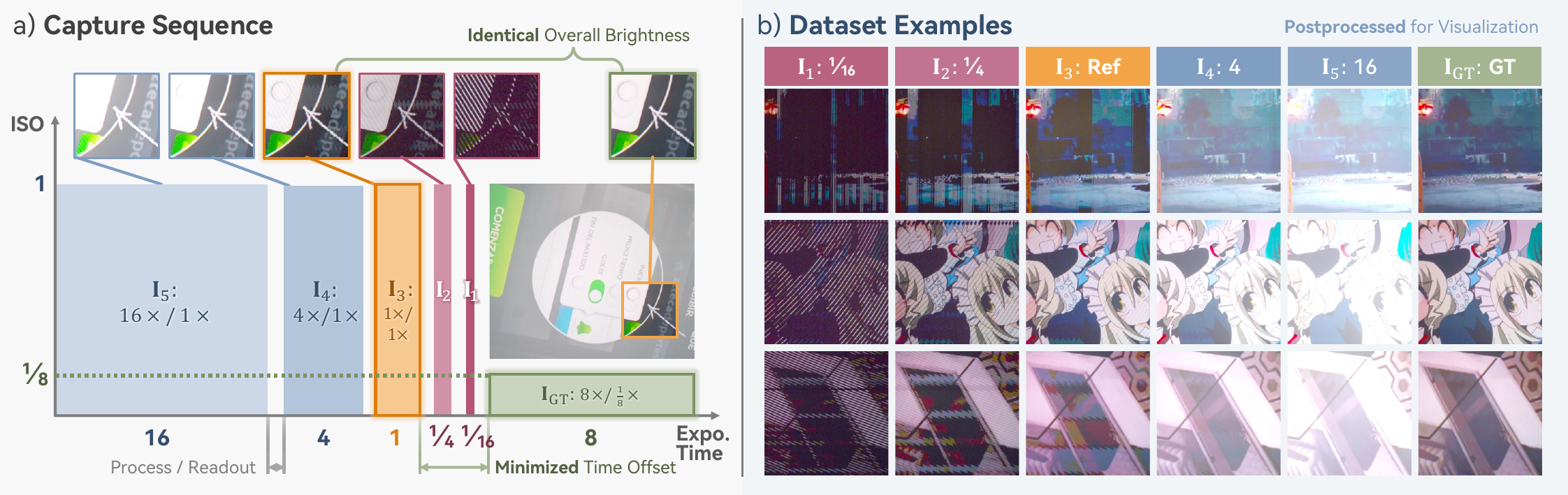}
    \vspace{-4mm}
    \caption{Overview of the real data collection process. \textbf{a)} The capture sequence and exposure parameters setting to align brightness and position between $\textbf{I}_\text{GT}$ and $\textbf{I}_\text{ref}$. \textbf{b)} Example captures from the real dataset, covering various content types and shooting conditions.}
    \label{fig:real_capture}
    \vspace{-5mm}
\end{figure}

Because FB artifacts diminish at longer exposures, we can obtain a near-artifact-free ground-truth frame by keeping $\text{ISO} \times T^{\text{expo}}$ constant while increasing $T^{\text{expo}}$ and proportionally lowering ISO. The capture sequence follows the exposure-time / ISO ratio pattern, as shown in Fig.~\ref{fig:real_capture}.a:
\begin{equation*}
    16\times\!/1\times,\; 4\times\!/1\times,\; 1\times\!/1\times\;(\text{reference}),\; \tfrac{1}{4}\times\!/1\times,\; \tfrac{1}{16}\times\!/1\times,\; 8\times\!/0.125\times\;(\text{GT}).
\end{equation*}
The camera is held nearly still and the GT frame is captured at about $1/100$ to $1/30$ s, which keeps it free of FB and aligned with the reference frame, whereas practical capture uses much shorter exposures. Shooting longer exposures first and the GT last further minimizes spatial drift, yielding reasonably well-aligned sequences even with handheld capture.

\vspace{-4mm}
\subsection{Model Design}
\label{sec:model}
\vspace{-2mm}

\subsubsection{Frequency-Aware Banding Prior (FABP)}
\label{sec:banding_frequency}
\vspace{-1mm}

Given a bracketed RAW sequence $\{\mathbf{I}_i^{\text{RAW}}\}_{i=1}^{T}$ with $T=5$ frames, where each $\mathbf{I}_i^{\text{RAW}} \in \mathbb{R}^{H \times W \times 4}$ is a Bayer RAW image, we first average the four channels to obtain a grayscale image:
$
    \mathbf{G}_i = \frac{1}{4} \sum_{c=1}^{4} \mathbf{I}_{i,c}^{\text{RAW}}, \quad \mathbf{G}_i \in \mathbb{R}^{H \times W}.
$

To suppress spectral leakage, we subtract the mean and apply a 2D Hann window $\mathbf{W} \in \mathbb{R}^{H \times W}$:
$
    \tilde{\mathbf{G}}_i = \bigl(\mathbf{G}_i - \mu(\mathbf{G}_i)\bigr) \odot \mathbf{W},
$
where $\mu(\mathbf{G}_i) = \frac{1}{HW} \sum_{u,v} \mathbf{G}_{i,u,v}$. We then compute the 2D discrete Fourier transform and shift the zero-frequency component to the center:
\begin{equation}
    \mathbf{S}_i = \operatorname{FFTShift}\bigl(\mathcal{F}_{\text{2D}}(\tilde{\mathbf{G}}_i)\bigr), \quad \mathbf{P}_i = |\mathbf{S}_i|^2, \quad \mathbf{P}_i \in \mathbb{R}^{H \times W}.
\end{equation}

Since real-valued signals yield conjugate-symmetric spectra, we retain only the valid half-plane. A binary mask $\mathbf{B}$ then excludes frequencies near the DC component and beyond a maximum radius:
\begin{equation}
    \mathbf{B}_{u,v} = \begin{cases}
        1, & r_{\min} \leq d_{u,v} \leq r_{\max} \;\land\; \bigl((u-c_u)<0 \lor ((u-c_u)=0 \land (v-c_v)>0)\bigr) \\
        0, & \text{otherwise},
    \end{cases}
\end{equation}
where $d_{u,v} = \sqrt{(u-c_u)^2 + (v-c_v)^2}$ is the Euclidean distance from frequency coordinates $(u,v)$ to the spectrum center $(c_u, c_v) = (\lfloor H/2 \rfloor, \lfloor W/2 \rfloor)$, and $r_{\min}, r_{\max}$ define the valid frequency ring. The masked power spectrum is $\tilde{\mathbf{P}}_i = \mathbf{P}_i \odot \mathbf{B}$.

\begin{figure}[t]
    \centering
    \includegraphics[width=\textwidth]{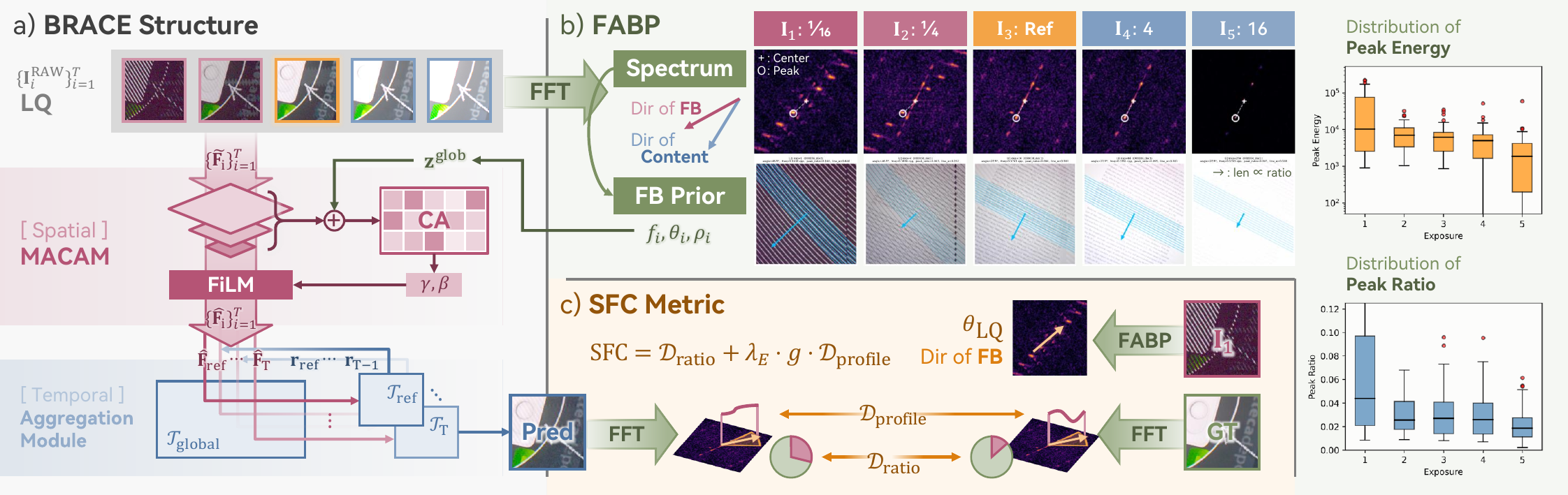}
    \vspace{-4.5mm}
    \caption{Overview of the proposed BRACE methods. \textbf{a)} The overall architecture. \textbf{b)} The banding frequency analysis shows the example of peak frequency shifts from PWM banding to inherent screen periodicity, with peak energy/ratio distribution visualization. \textbf{c)} The SFC metric calculation.}
    \vspace{-5mm}
    \label{fig:model_structure}
\end{figure}

We locate the dominant spectral peak via argmax:
$
    (u_i^*, v_i^*) = \arg\max_{u,v} \tilde{\mathbf{P}}_{i,u,v}.
$ From this peak, we extract three banding descriptors for each frame:
\begin{equation}
\begin{aligned}
    f_{x,i} &= \frac{v_i^* - c_v}{W}, \quad
     f_{y,i} = \frac{u_i^* - c_u}{H}, \quad
     f_i = \sqrt{f_{x,i}^2 + f_{y,i}^2}, \\
    \theta_i &= \Bigl(\operatorname{atan2}(f_{y,i}, f_{x,i}) + \frac{\pi}{2}\Bigr) \bmod \pi, \quad
     \rho_i = \frac{\tilde{\mathbf{P}}_{i,u_i^*,v_i^*}}{\sum_{u,v} \tilde{\mathbf{P}}_{i,u,v}}
    = \frac{E_i^{\text{peak}}}{E_i^{\text{total}}},
\end{aligned}
\end{equation}
where $f_i$ is the stripe frequency in cycles per pixel, $\theta_i \in [0, \pi)$ is the stripe orientation, and $\rho_i$ is the peak energy ratio. A higher $\rho_i$ indicates stronger and more coherent banding at a dominant frequency, most pronounced in underexposed frames, as analyzed in Fig.~\ref{fig:model_structure}.b.

\vspace{-2mm}
\subsubsection{Spatial Cross-Attention Fusion}
\vspace{-1.5mm}

We pack each frame's banding attributes into $\mathbf{s}_i \in \mathbb{R}^{8}$, including frequency, direction, energy, energy ratio, and phase, and compute the $\rho_i$-weighted mean $\boldsymbol{\mu}$ and standard deviation $\boldsymbol{\sigma}$.
These cross-frame statistics are projected through an MLP into a compact global prior token:
\begin{equation}
    \mathbf{z}^{\text{glob}} = \operatorname{MLP}\bigl([\boldsymbol{\mu}; \boldsymbol{\sigma}]\bigr) \in \mathbb{R}^{d}, \quad d = 64.
\end{equation}
The token $\mathbf{z}^{\text{glob}}$ captures global banding frequency, orientation, energy concentration, and cross-frame variability, but lacks frame-level spatial interaction. We therefore design the \textbf{Multi-scale Spatial Cross-Attention Modulator (MSCAM)} to inject this prior into cross-frame fusion.

Let $\{\tilde{\mathbf{F}}_i\}_{i=1}^{T}$ denote features aligned by optical-flow warping and deformable convolution, where $\tilde{\mathbf{F}}_1$ is the reference. MSCAM pools each feature map at three scales using $\operatorname{AvgPool}_{s \times s}$ with $s \in \{1,2,4\}$, producing multi-granularity tokens $\mathbf{h}_i^{(s)} \in \mathbb{R}^{C}$. These tokens are fused with the global prior:
\begin{equation}
    \mathbf{t}_i^{(s)} = \mathbf{W}_f \, \mathbf{h}_i^{(s)} + \mathbf{W}_g \, \mathbf{z}^{\text{glob}} \in \mathbb{R}^{d_a},
\end{equation}
where $\mathbf{W}_f, \mathbf{W}_g \in \mathbb{R}^{d_a \times C}$ and $d_a = 64$. We then apply scaled dot-product cross-attention with the reference frame serving as the query and all frames as keys and values:
\begin{equation}
    \mathbf{c}^{(s)} = \operatorname{Attention}\bigl(Q=\mathbf{t}_3^{(s)},\; K=\{\mathbf{t}_i^{(s)}\}_{i=1}^{T},\; V=\{\mathbf{t}_i^{(s)}\}_{i=1}^{T}\bigr),
\end{equation}
which learns the importance of each frame relative to the reference, conditioned on the banding prior. The resulting multi-scale contexts are then mapped by an MLP to FiLM~\cite{perez2018film} modulation parameters:
\begin{equation}
    \mathbf{c} = \frac{1}{3} \sum_{s \in \{1,2,4\}} \mathbf{c}^{(s)}, \quad
    [\boldsymbol{\gamma}, \boldsymbol{\beta}] = \operatorname{MLP}(\mathbf{c}) \in \mathbb{R}^{2C}.
\end{equation}
These parameters modulate the aligned features channel-wise:
\begin{equation}
    \hat{\mathbf{F}}_i = \tilde{\mathbf{F}}_i \odot \bigl(1 + \alpha \cdot \tanh(\boldsymbol{\gamma})\bigr) + \alpha \cdot \boldsymbol{\beta},
\end{equation}
where $\alpha = 0.1$ controls the modulation strength and $\boldsymbol{\gamma}, \boldsymbol{\beta} \in \mathbb{R}^{C \times 1 \times 1}$ are broadcast across all spatial positions. This FiLM scheme amplifies or suppresses channels according to the global prior and cross-frame similarity without dense spatial attention.

\vspace{-1.5mm}
\subsubsection{Overall Model}
\vspace{-2mm}
Multi-frame methods typically align frames by optical flow and then aggregate features~\cite{wu2023rbsr,wang2023benchmark}. Following this pipeline, FABP-guided MSCAM modulates the recurrent network during spatial cross-frame fusion, and we incorporate the temporal aggregation module from TMRNet~\cite{Zhang2025BracketIRE} to exploit the temporal structure of bracketed exposures. The recurrent state update takes the form
\begin{equation}
    \mathbf{r}_i = \mathcal{T}_i\bigl(\mathcal{T}_{\text{global}}(\hat{\mathbf{F}}_i, \mathbf{r}_{i-1})\bigr),
\end{equation}
where $\mathbf{r}_{i-1}$ is the hidden state from the previous frame, $\hat{\mathbf{F}}_i$ is the MSCAM-modulated feature map, and $\mathcal{T}_{\text{global}}$ and $\mathcal{T}_i$ denote the global aggregation and per-frame refinement modules inherited from TMRNet, respectively. The final restoration result is obtained by feeding the output features of the last frame through a reconstruction network, as illustrated in Fig.~\ref{fig:model_structure}. Training proceeds in two stages: the model is first trained on the synthetic dataset to learn basic banding removal, and then fine-tuned on the real dataset to adapt to the more complex degradations in real-world captures.

\textbf{Stage One: Synthetic Pretraining.}
In addition to the standard L1 loss, we adopt the multi-level wavelet-based frequency loss~\cite{Zhu2026CLEAR}. Since banding artifacts reside predominantly in the low-frequency components of the image (Sec.~\ref{sec:banding_frequency}), this loss enhances restoration of those components by computing a weighted discrepancy over multi-scale, multi-orientation subbands, which stabilizes the suppression of distortions at different levels and orientations. The total loss for stage one is
\begin{equation}
    \mathcal{L}_{\text{stage1}} = \mathcal{L}_{\text{L1}} + \lambda_{\text{wavelet}} \cdot \mathcal{L}_{\text{MWL}}.
\end{equation}

\textbf{Stage Two: Real-World Fine-Tuning.}
The second stage adds a color loss for the color shifts common in real captures. As analyzed in Sec.~\ref{sec:problem}, FB often introduces chromatic deviations that widen the perceptual and color gap to the ground truth, so we adopt MS-SWD~\cite{he2024msswd}, which matches distributions across Gaussian pyramids in CIELAB space. The total loss is
\begin{equation}
    \mathcal{L}_{\text{stage2}} = \mathcal{L}_{\text{stage1}} + \lambda_{\text{color}} \cdot \mathcal{L}_{\text{MSSWD}}.
\end{equation}

\begin{figure}[t]
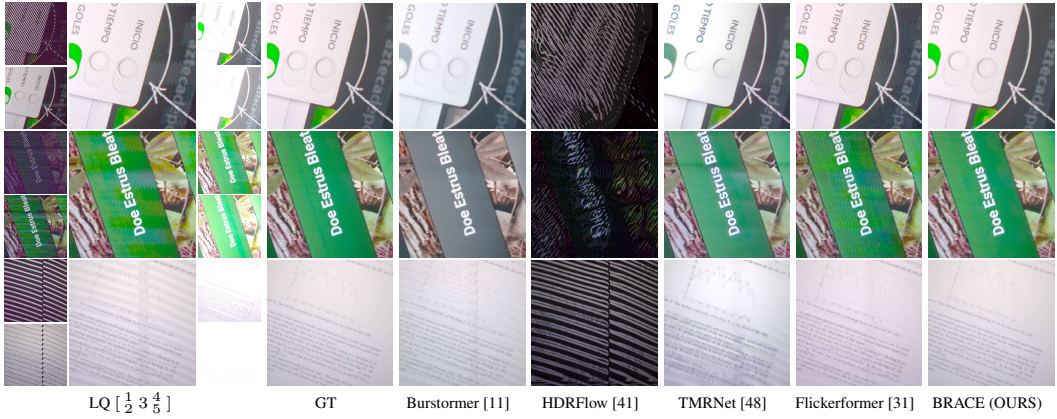

  \centering
    \insertcomparison{000002_UI}
    \insertcomparison{000011_UI}
    \insertcomparison{000016_doc2}
    \vspace{0.5mm}
    \insertmethodlegend
    \caption{Visual comparison on the real dataset. BRACE gains great advantages over other methods.}
    \vspace{-4mm}
    \label{fig:qual_results}
\end{figure}

\vspace{-6mm}
\subsection{Stripe Frequency Consistency (SFC)}
\label{sec:sfc}
\vspace{-2mm}

To quantitatively evaluate banding removal, we propose SFC, which measures how well the predicted output reproduces the banding-direction energy structure of the ground truth. Let $\mathbf{I}_{1}$ be the shortest-exposure frame. We first extract the dominant banding direction $\theta_{\text{LQ}} \in (-\pi, \pi]$ from $\mathbf{I}_{1}$ via FABP, and then compute the power spectra of both the predicted output $\mathbf{I}_{\text{pred}}$ and the ground truth $\mathbf{I}_{\text{GT}}$ to measure their directional energy ratio within a sector centered at $\theta_{\text{LQ}}$:
\vspace{1mm}
\begin{equation}
    \rho_{\text{pred}} = \frac{E_{\text{sector}}(\mathbf{I}_{\text{pred}}; \theta_{\text{LQ}})}{E_{\text{ring}}(\mathbf{I}_{\text{pred}})},\quad
    \rho_{\text{gt}} = \frac{E_{\text{sector}}(\mathbf{I}_{\text{GT}}; \theta_{\text{LQ}})}{E_{\text{ring}}(\mathbf{I}_{\text{GT}})},
\vspace*{1mm}
\end{equation}
where $E_{\text{sector}}$ sums the power spectrum within a sector of half-width $\Delta\theta = 10^\circ$ around $\theta_{\text{LQ}}$, and $E_{\text{ring}}$ is the total energy in the valid frequency ring (excluding DC and excessive frequencies).

We further extract the radial energy profile along this sector by averaging the power over all frequency points that lie at the same radius $k$ from the spectrum center:
\vspace{1mm}
\begin{equation}
    P(k) = \frac{1}{|\mathcal{S}_k|} \sum_{(u,v) \in \mathcal{S}_k} \tilde{\mathbf{P}}_{u,v}, \quad k = r_{\min}, \dots, r_{\max},
\end{equation}
where $\mathcal{S}_k$ denotes the set of frequency coordinates at radius $k$ within the sector. SFC then combines a direction-ratio distance and a radial-profile distance, gated by banding saliency:
\vspace{1mm}
\begin{equation}
    \text{SFC} = \mathcal{D}_{\text{ratio}} + \lambda_E \cdot g \cdot \mathcal{D}_{\text{profile}},
\end{equation}
where
$
    \mathcal{D}_{\text{ratio}} = |\rho_{\text{pred}} - \rho_{\text{gt}}|,\quad
    \mathcal{D}_{\text{profile}} = \frac{1}{K} \sum_{k=1}^{K} |P_{\text{pred}}(k) - P_{\text{gt}}(k)|,
$
and $g = \max(\rho_{\text{pred}}, \rho_{\text{gt}})$ gates the profile distance: when banding-direction energy is low, the profile distance is down-weighted to avoid penalizing predictions where banding is already weak or absent. $\lambda_E = 5.0$ balances the two terms, and SFC is scaled by 100 for readability. Appendix~\ref{sec:sfc_detail} reports a controlled sensitivity study, in which the same scene is measured across ten exposure times, and a decomposition of SFC into its direction-ratio and radial-profile terms for all compared methods.

\vspace{-2mm}
\section{Experiments}
\vspace{-2.5mm}
\subsection{Experimental Setup}
\vspace{-2mm}

\textbf{Data Construction.}
To ensure diversity, we collect screen-content images from multiple sources~\cite{li2020docbank,zhong2019publaynet,LSDIR,taesiri2024videogamebunnyvisionassistantsvideo,deka2017rico,wang2025dscvc,scvqa2025}. The Bricker-Syn dataset contains 1,000 training and 100 test samples, while the Bricker-Real dataset contains 250 training and 40 test samples. Each sample includes a 5-frame bracketed RAW sequence and a corresponding ground-truth frame. Real images are downsampled by 4 to suppress fine screen texture and capture imperfections. The real test split is manually verified to ensure adequate alignment between the reference frame and the ground truth (Appendix~\ref{sec:app_gt_validation}).

\textbf{Implementation Details.}
We use a batch size of 16 and randomly crop input patches to $128 \times 128$ with random horizontal flipping and rotation. The model is optimized with AdamW~\cite{loshchilov2019adamw}. In stage one, the learning rate is set to $1 \times 10^{-4}$, and training runs for 50 epochs. In stage two, the learning rate is lowered to $1 \times 10^{-5}$, and fine-tuning runs for 20 epochs. The loss weights are set to $\lambda_{\text{wavelet}} = 0.5$ and $\lambda_{\text{color}} = 0.05$. All experiments are conducted on an NVIDIA RTX A6000 GPU (Appendix~\ref{sec:app_cost}).

\textbf{Evaluation Metrics.}
We assess restoration quality using PSNR and SSIM~\cite{wang2004ssim} for pixel-level and structural fidelity, LPIPS~\cite{zhang2018lpips} and DISTS~\cite{ding2022dists} for perceptual quality, and MS-SWD~\cite{he2024msswd} to measure color differences between the restored output and the ground truth. Additionally, we report our proposed SFC metric to quantify banding removal in the frequency domain.

\textbf{Compared Methods.}
We compare BRACE with representative multi-frame and bracketed exposure restoration methods, including Burstormer~\cite{dudhane2023burstormer}, HDRFlow~\cite{xu2024hdrflow}, TMRNet~\cite{Zhang2025BracketIRE}, and FlickerFormer~\cite{qu2026ittakestwo}. For fairness, each method is retrained on Bricker following its own training protocol. More implementation details are provided in Appendix~\ref{sec:app_implementation_details}.

\begin{table*}[t]
\centering
\caption{Quantitative comparison on Bricker-Syn and Bricker-Real. The best and second best results are colored with \best{red} and \secondbest{blue}. BRACE achieves consistent improvements across all metrics.}
\vspace{-6pt}
\scriptsize
\renewcommand{\arraystretch}{1.10}
\label{tab:quant_results}
\vspace{1mm}
\begin{tabularx}{\textwidth}{
    l
    *{11}{>{\centering\arraybackslash}X}
}
\toprule
\multirow{2}{*}{\textbf{Method}} & \multicolumn{5}{c}{\textbf{Syn}} & \multicolumn{6}{c}{\textbf{Real}} \\
\cmidrule(lr){2-6} \cmidrule(lr){7-12}
& PSNR$\uparrow$ & SSIM$\uparrow$ & LPIPS$\downarrow$ & DISTS$\downarrow$ & \tiny{MSSWD}$\downarrow$
& PSNR$\uparrow$ & SSIM$\uparrow$ & LPIPS$\downarrow$ & DISTS$\downarrow$ & \tiny{MSSWD}$\downarrow$ & SFC$\downarrow$ \\
\midrule
Burstormer~\cite{dudhane2023burstormer} & 24.59 & 0.8156 & 0.3240 & 0.2012 & 1.2731
& 24.92 & 0.7480 & 0.4374 & 0.2398 & 1.3537 & 0.5153 \\

HDRFlow~\cite{xu2024hdrflow} & 6.62 & 0.1160 & 0.8224 & 0.5924 & 7.4281 
& 6.75 & 0.1287 & 0.7883 & 0.4565 & 6.2348 & 17.6967 \\

TMRNet~\cite{Zhang2025BracketIRE} & \secondbest{26.79} & \secondbest{0.8708} & \secondbest{0.2560} & 0.2419 & 0.7040
& 23.63 & \secondbest{0.7704} & 0.2820 & 0.1797 & 1.0473 & 1.1135 \\

FlickerFormer~\cite{qu2026ittakestwo} & 24.86 & 0.5771 & 0.4159 & \secondbest{0.2048} & \secondbest{0.5779}
& \secondbest{25.49} & 0.6690 & \secondbest{0.1977} & \secondbest{0.1361} & \secondbest{0.6615} & \secondbest{0.4036} \\

\textbf{BRACE} (ours) & \best{28.44} & \best{0.8823} & \best{0.2050} & \best{0.1606} & \best{0.5102}
& \best{28.15} & \best{0.8007} & \best{0.1925} & \best{0.1113} & \best{0.2052} & \best{0.1847} \\

\bottomrule
\end{tabularx}
\vspace{-2mm}
\end{table*}

\vspace{-2mm}
\subsection{Main Results}
\vspace{-2mm}

\textbf{Quantitative Results.}
Table~\ref{tab:quant_results} reports quantitative results on synthetic and real test sets. BRACE achieves the best performance across all metrics, showing strong banding removal while preserving image quality and color fidelity. Its SFC improvement further confirms effective frequency-domain banding suppression, consistent with the visual results.

\textbf{Qualitative Results.}
Figure~\ref{fig:qual_results} shows visual comparisons on Bricker-Real. BRACE removes banding artifacts while restoring color and texture details, including in challenging cases with strong banding and color shifts. In contrast, other methods either fail to fully suppress banding (e.g., FlickerFormer) or leave noticeable color distortions (e.g., Burstormer, TMRNet). HDRFlow's design for video HDR reconstruction failed to handle the severe changes of FB artifacts across the bracketed sequence.

\vspace{-2mm}
\subsection{Ablation Study}
\vspace{-2mm}

\begin{figure*}[t]
\centering
\begin{minipage}[c]{0.685\textwidth}
\centering
\captionof{table}{Ablation study on different components. The best and second best results are colored with \best{red} and \secondbest{blue}.}
\vspace{-6pt}
\label{tab:ablation}
\scriptsize
\vspace{1mm}
\begin{tabularx}{\linewidth}{
    *{1}{>{\centering\arraybackslash}p{0.35cm}}
    *{2}{>{\centering\arraybackslash}p{0.45cm}}
    *{2}{>{\centering\arraybackslash}p{0.35cm}}
    *{6}{>{\centering\arraybackslash}X}
}
\toprule
\multirow{2}{*}{\makebox[0pt][c]{\textbf{Expr.}}}
& \multicolumn{2}{c}{\textbf{Stage 1}} & \multicolumn{2}{c}{\textbf{Stage 2}}
& \multicolumn{6}{c}{\textbf{Metrics}} \\
\cmidrule(lr){2-3} \cmidrule(lr){4-5} \cmidrule(lr){6-11}
& $\mathcal{L}_\text{MWL}$ & \makebox[0pt][c]{\tiny MSCAM} & $\mathcal{L}_\text{L1}$ & \makebox[0pt][c]{$\mathcal{L}_\text{Color}$}
& PSNR$\uparrow$ & SSIM$\uparrow$ & LPIPS$\downarrow$ & DISTS$\downarrow$ & \tiny{MSSWD}$\downarrow$ & \makebox[0pt][c]{\hspace{5pt} SFC$\downarrow$} \\
\midrule
1 & & & & & 22.53 & 0.7678 & 0.2720 & 0.1642 & 1.1487 & 1.0612 \\
2 & & $\checkmark$ & & & 23.47 & 0.7726 & 0.2706 & 0.1558 & 1.0393 & 0.8202 \\
3 & $\checkmark$ & & & & 23.22 & 0.7714 & 0.2561 & 0.1510 & 0.9110 & 0.7071 \\
4 & $\checkmark$ & $\checkmark$ & & & 24.24 & 0.7742 & 0.2534 & 0.1551 & 0.7870 & 0.7984 \\
\arrayrulecolor{gray}\midrule\arrayrulecolor{black}
5 & $\checkmark$ & $\checkmark$ & & $\checkmark$ & 27.04 & 0.7810 & \secondbest{0.2094} & \secondbest{0.1266} & 0.3023 & \secondbest{0.2458} \\
6 & $\checkmark$ & $\checkmark$ & $\checkmark$ & & \best{28.23} & \best{0.8069} & 0.2116 & 0.1345 & \secondbest{0.2510} & 0.2629 \\
7 & $\checkmark$ & $\checkmark$ & $\checkmark$ & $\checkmark$ & \secondbest{28.15} & \secondbest{0.8007} & \best{0.1925} & \best{0.1113} & \best{0.2052} & \best{0.1847} \\
\bottomrule
\end{tabularx}
\vspace{-1mm}
\end{minipage}%
\hfill
\begin{minipage}[c]{0.29\textwidth}
\centering
\vspace{2pt}
\scriptsize\setlength{\tabcolsep}{1.0pt}\renewcommand{\arraystretch}{0.6}
\begin{tabular}{@{}ccc@{}}
\includegraphics[width=0.32\linewidth]{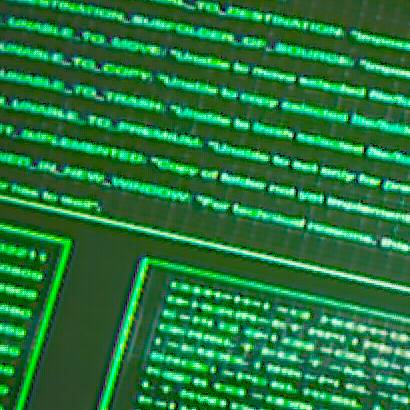} &
\includegraphics[width=0.32\linewidth]{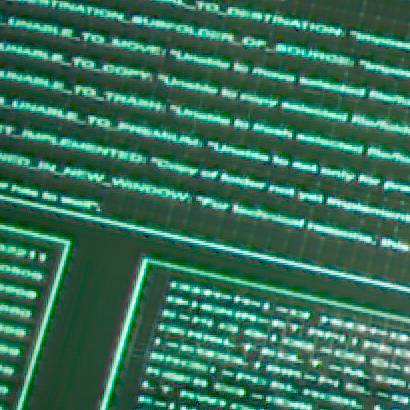} &
\includegraphics[width=0.32\linewidth]{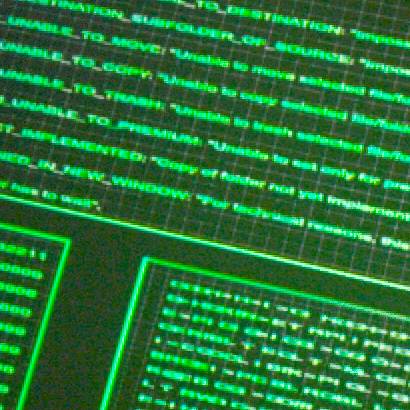} \\
\tiny w/ $\mathcal{L}_\text{Color}$ & \tiny w/o $\mathcal{L}_\text{Color}$ & \tiny GT \\[4pt]
\includegraphics[width=0.32\linewidth]{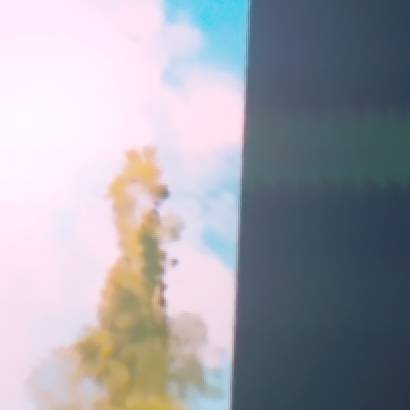} &
\includegraphics[width=0.32\linewidth]{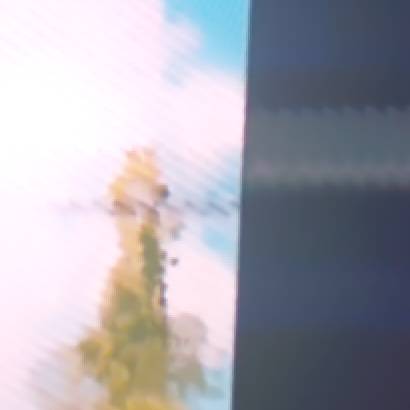} &
\includegraphics[width=0.32\linewidth]{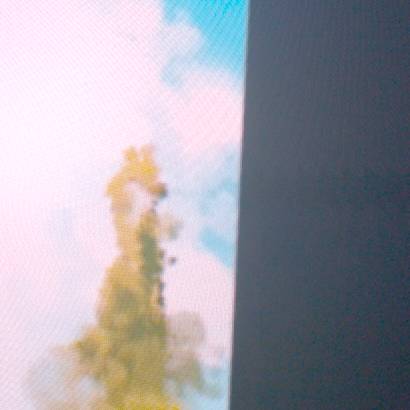} \\
\tiny w/ GT $\mathcal{L}_\text{L1}$ & \tiny w/o GT $\mathcal{L}_\text{L1}$ & \tiny GT \\[-2pt]
\end{tabular}
\caption{Visual comparison of ablation settings.}
\label{fig:ablation_visual}
\end{minipage}
\vspace{-4mm}
\end{figure*}


Table~\ref{tab:ablation} and Fig.~\ref{fig:ablation_visual} report the ablation study, with all metrics measured on Bricker-Real. The Stage 1 columns mark whether the model uses MSCAM and the wavelet loss, and the Stage 2 columns mark the losses added in real fine-tuning, where $\mathcal{L}_{\text{L1}}$ is the L1 loss to the real ground truth. MSCAM improves DISTS/SFC by injecting the banding prior into cross-frame fusion. The wavelet loss strengthens frequency-domain supervision, while synthetic-only training already transfers to real captures. During real fine-tuning, $\mathcal{L}_{1}$ improves fidelity-oriented metrics and $\mathcal{L}_{\mathrm{Color}}$ improves perceptual quality and color consistency. Additional ablations on the number of input frames and on uniform-exposure input are reported in Appendix~\ref{sec:app_additional}.

\vspace{-2mm}
\section{Conclusion}
\vspace{-2.5mm}
We presented a systematic analysis of complex Flicker-Banding, and introduced Bricker, the first bracketed RAW dataset for FB, and BRACE, a multi-frame model combining frequency-aware priors with spatial cross-attention fusion. We also proposed the SFC metric for banding removal evaluation. Extensive experiments validate our approach. We believe this dataset and framework will support future work on screen-capture restoration under diverse real-world capture conditions.

\newpage

\begin{ack}
\vspace{-2mm}
This work was supported by Huawei Explore X funds and the CAAI-Tencent Rhino-Bird Open Research Fund. This work was also sponsored by the AI Hundred Schools Program. The authors declare no competing financial interests related to this work.
\end{ack}

\bibliographystyle{plain}
{\small
\bibliography{neurips_2026}
}

\newpage
\appendix

\section{Detailed Causes of Complex FB Degradations}
\label{sec:app_complex_banding}

\medskip
\begin{figure}[ht]
    \centering
    \includegraphics[width=0.48\linewidth]{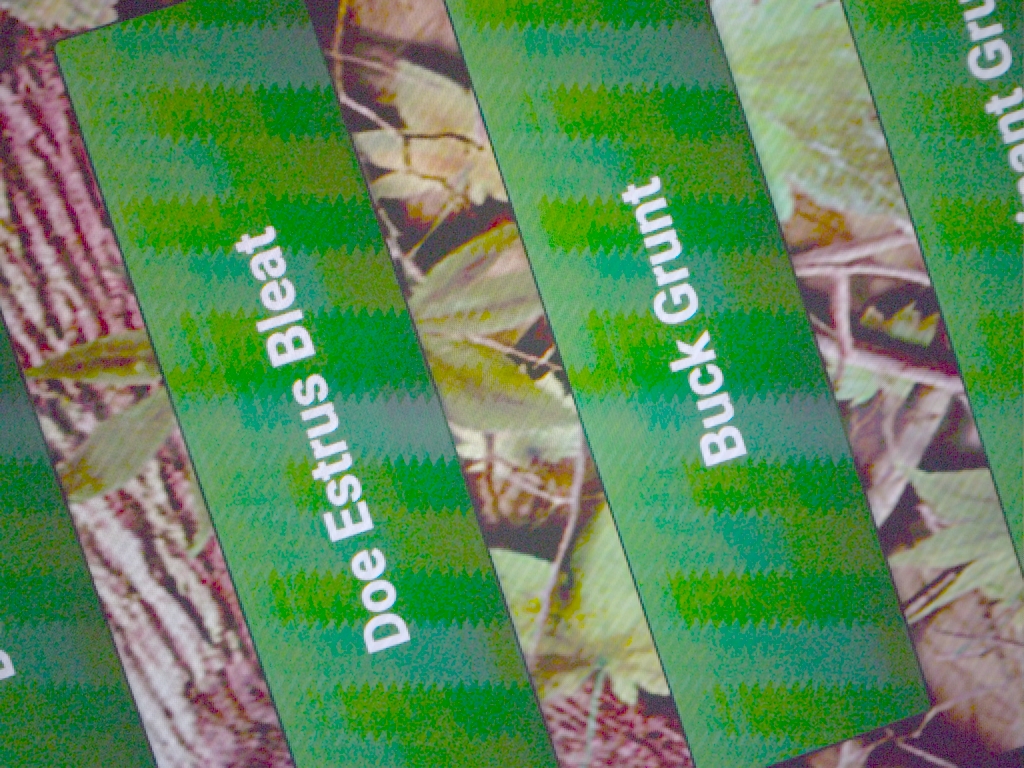}\hfill
    \includegraphics[width=0.48\linewidth]{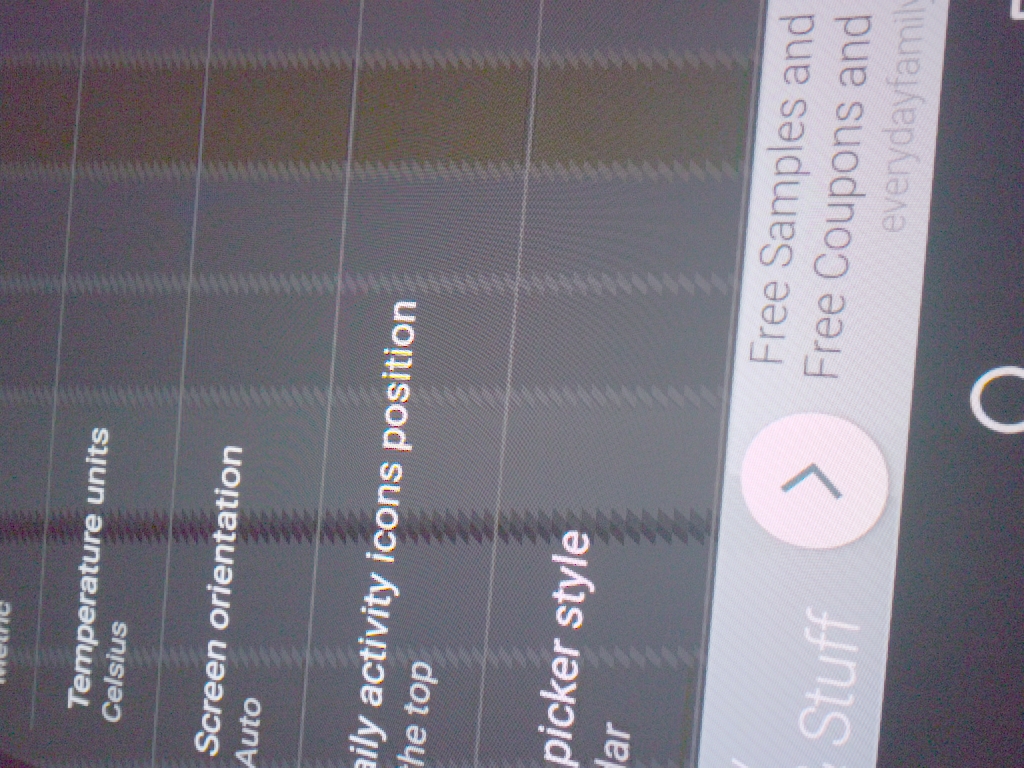}
    \caption{Examples of complex FB degradations. \textbf{Left:} Jagged FB with color shift. \textbf{Right:} Jagged FB with compound periodicity. Both degradations are caused by the brightness dimming mechanism.}
    \label{fig:complex_banding_examples}
\end{figure}
\medskip

We identify several complex FB degradations that commonly occur in Sec.~\ref{sec:problem}, including color shift, compound FB, and jagged FB, and analyze their underlying causes as follows.

\textbf{Color Shift.}
RGB subpixels achieve their target luminance through independent PWM patterns. Since different subpixels require different brightness levels, their temporal on/off sequences are inherently misaligned. Consequently, at any given instant, the instantaneous color of a pixel deviates from the intended hue: for instance, a region meant to appear orange may present as red, yellow, or even green depending on which combination of subpixels is active at that moment (Fig.~\ref{fig:complex_banding}.c, Fig.~\ref{fig:complex_banding_examples}). The rolling shutter samples different temporal phases at different rows, causing this color inconsistency to vary spatially across the frame and introducing pronounced chromatic shifts on top of luminance banding.

\textbf{Compound FB.}
Many display drivers embed nested temporal cycles: a longer outer cycle with period $T_O$ contains several shorter inner cycles ($T_I \ll T_O$), often separated by a blanking margin interval used for frame refresh or data processing~\cite{lin2021active}. The resulting brightness modulation is a superposition of periodic components at multiple scales, producing composite banding patterns that combine coarse and fine periodic structures (Fig.~\ref{fig:complex_banding}.d, Fig.~\ref{fig:complex_banding_examples}).

\textbf{Jagged FB.}
Modular LED displays are assembled from independently driven tiles, each operating with a fixed phase offset in its refresh timing. As the rolling shutter scans across tile boundaries, the sampled banding phase undergoes a discontinuous jump, causing the observed FB to appear as stair-shaped, block-like, or diamond-shaped geometric patterns rather than continuous parallel stripes (Fig.~\ref{fig:complex_banding}.e, Fig.~\ref{fig:complex_banding_examples}). When superimposed with color shift, these fragmented patterns produce particularly severe and visually disruptive FB artifacts in darker image regions.

\section{Simulation Pipeline and Reverse ISP}
\label{sec:app_simulation}

The simulator is a self-contained Python/NumPy forward imaging model built from the ground up. Each screen image is placed at a spatial position relative to a virtual camera, and the camera--screen geometry is modeled explicitly: camera pixels are mapped to screen coordinates by ray intersection, and the observed brightness follows from integrating the screen emission over each pixel's exposure window (Sec.~\ref{sec:dataset}). Since both the screen content and its temporal modulation enter the integral, the resulting mask depends on content, geometry, and exposure time, and differs across the bracketed frames.

For the reverse ISP, the source content is nonlinearly encoded sRGB. We apply the inverse gamma transfer function to obtain linear RGB, undo the white-balance gains and the color correction matrix, mosaic the result into a Bayer pattern, and add sensor noise. White-balance gains and CCMs are aggregated from the DNG files analyzed in Appendix~\ref{sec:app_exif}, with a moderate range expansion when sampling. Motion blur uses near-linear kernels of up to 20 px and glow uses Gaussian blur, with the blur direction following the inter-frame perturbation of the scene construction.

\section{RAW dng EXIF Metadata}
\label{sec:app_exif}

As the post-processing of RAW images is heavily dependent on specific device characteristics, we provide the analysis of the EXIF metadata on multiple devices to inform the design of our simulation pipeline and to facilitate future research in this area.

The DNG (Digital Negative) format is a widely used RAW image format that contains both the raw sensor data and metadata describing the capture conditions and device characteristics~\cite{adobe2023dngspec}. It contains a wealth of information that can be crucial for accurately simulating the post-processing process and for developing an effective reverse-ISP pipeline. Key EXIF metadata fields relevant to our work include channel gains (e.g., `AsShotNeutral') and color correction matrices (CCMs, e.g., `ColorMatrix1', `ColorMatrix2'), which provide insights into the color response and white balance settings of the camera.

\begin{figure}[ht]
    \centering
    \begin{subfigure}{0.48\linewidth}
        \centering
        \includegraphics[width=\linewidth]{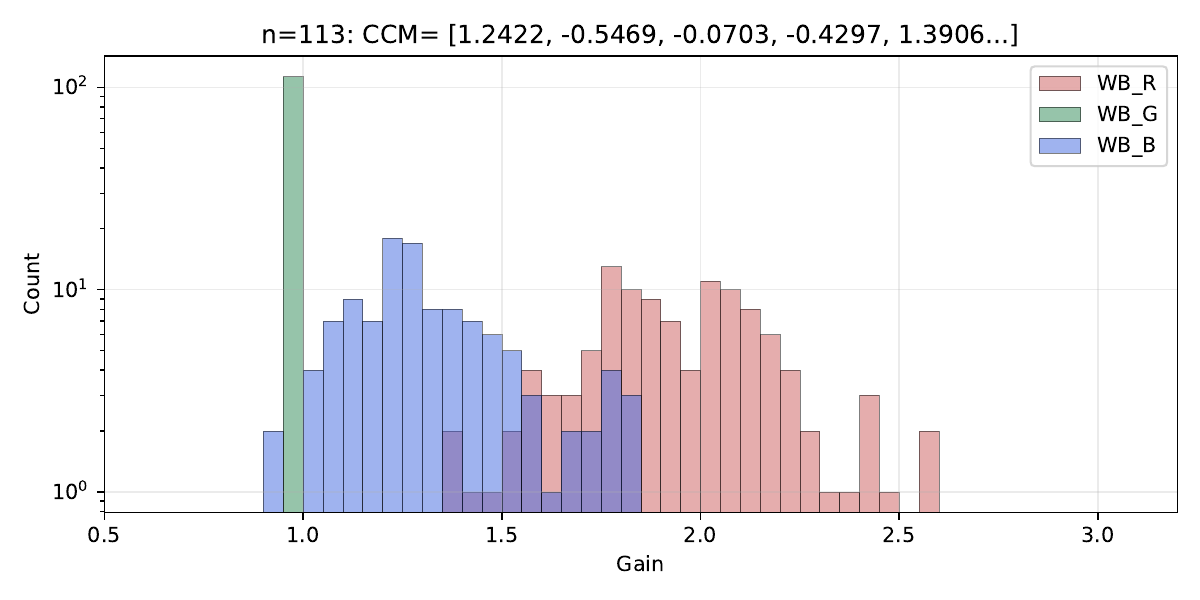}
        \vspace{-15pt}
        \caption{Xiaomi 17 Pro, Lens 1}
        \vspace{5pt}
    \end{subfigure}
    \hfill
    \begin{subfigure}{0.48\linewidth}
        \centering
        \includegraphics[width=\linewidth]{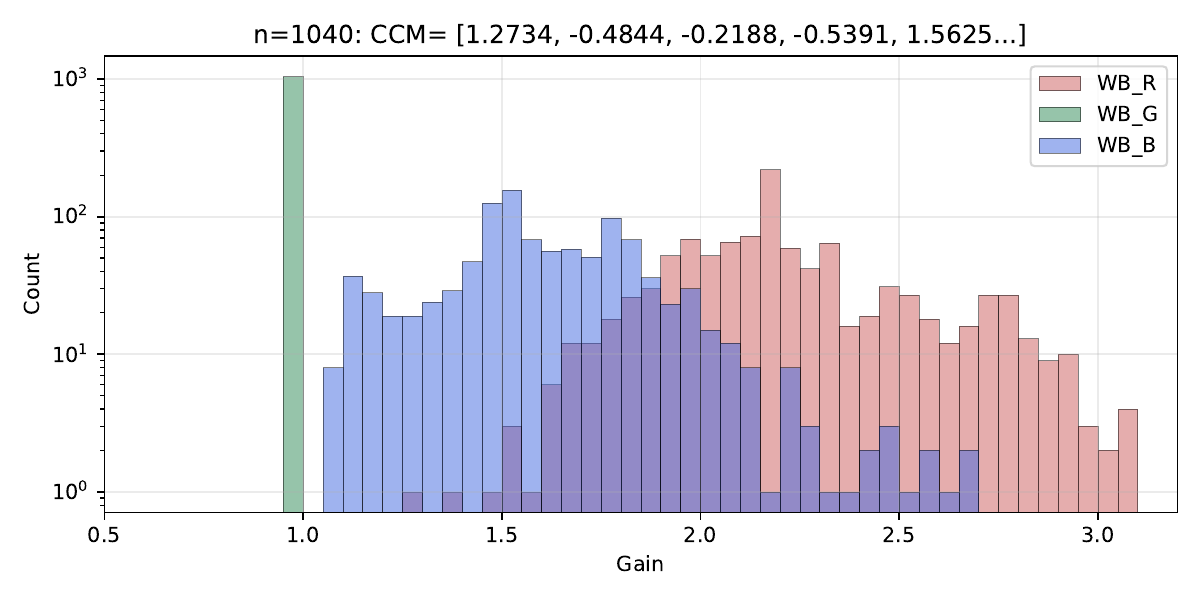}
        \vspace{-15pt}
        \caption{Xiaomi 17 Pro, Lens 2}
        \vspace{5pt}
    \end{subfigure}

    \vfill
    
    \begin{subfigure}{0.48\linewidth}
        \centering
        \includegraphics[width=\linewidth]{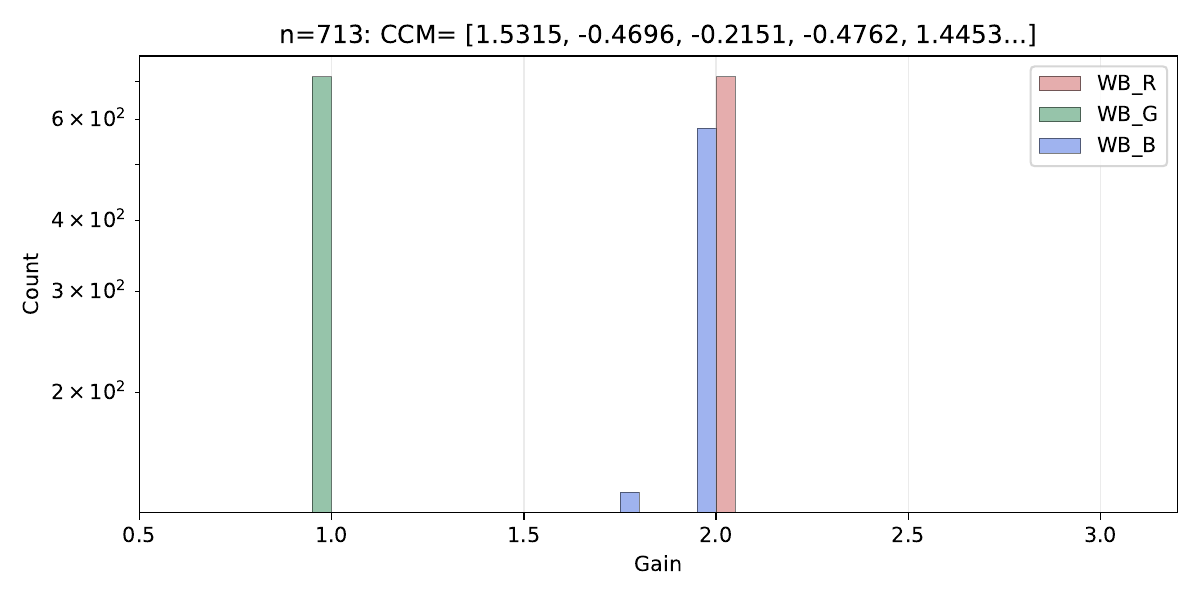}
        \vspace{-15pt}
        \caption{Vivo X200 Pro}
    \end{subfigure}
    \hfill
    \begin{subfigure}{0.48\linewidth}
        \centering
        \includegraphics[width=\linewidth]{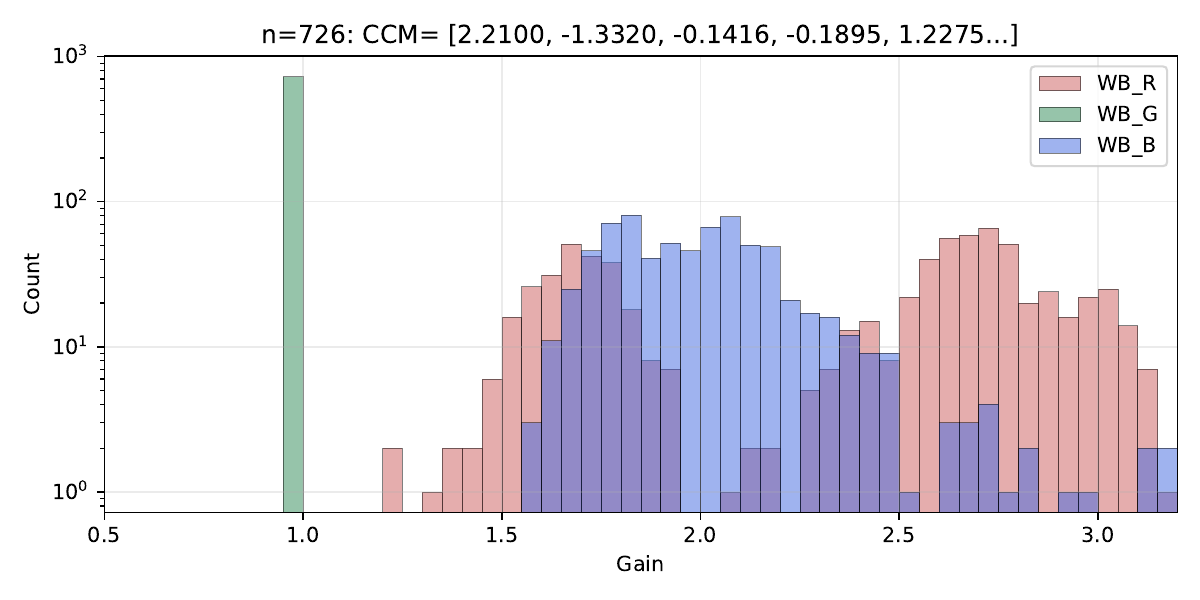}
        \vspace{-15pt}
        \caption{Huawei Mate 60 Pro}
    \end{subfigure}
    \caption{WB gains distribution of the DNG files captured by different devices.}
    \label{fig:exif_wb_gains}
\end{figure}

Therefore, we extract and analyze the gains and CCMs from the DNG files captured by multiple devices under various exposure settings, as visualized in Fig.~\ref{fig:exif_wb_gains}. We observe that the gains and CCMs exhibit significant variation across different devices and exposure conditions, reflecting the diversity in sensor characteristics and ISP processing pipelines. This tells us that a one-size-fits-all simulation approach may not be sufficient to capture the complex interactions between the sensor data and the ISP, and that device-specific modeling may be necessary for accurate simulation and effective reverse-ISP design.

\section{Capture Protocol and Automated Tool}
\label{sec:app_capture_tool}

Our automated capture tool is developed based on the open-source Android application Open Camera~\cite{opencamera}, which provides access to the Android Camera2 API~\cite{camera2_api} for fine-grained camera control.

\begin{figure}[ht]
    \centering
    \includegraphics[width=0.75\linewidth]{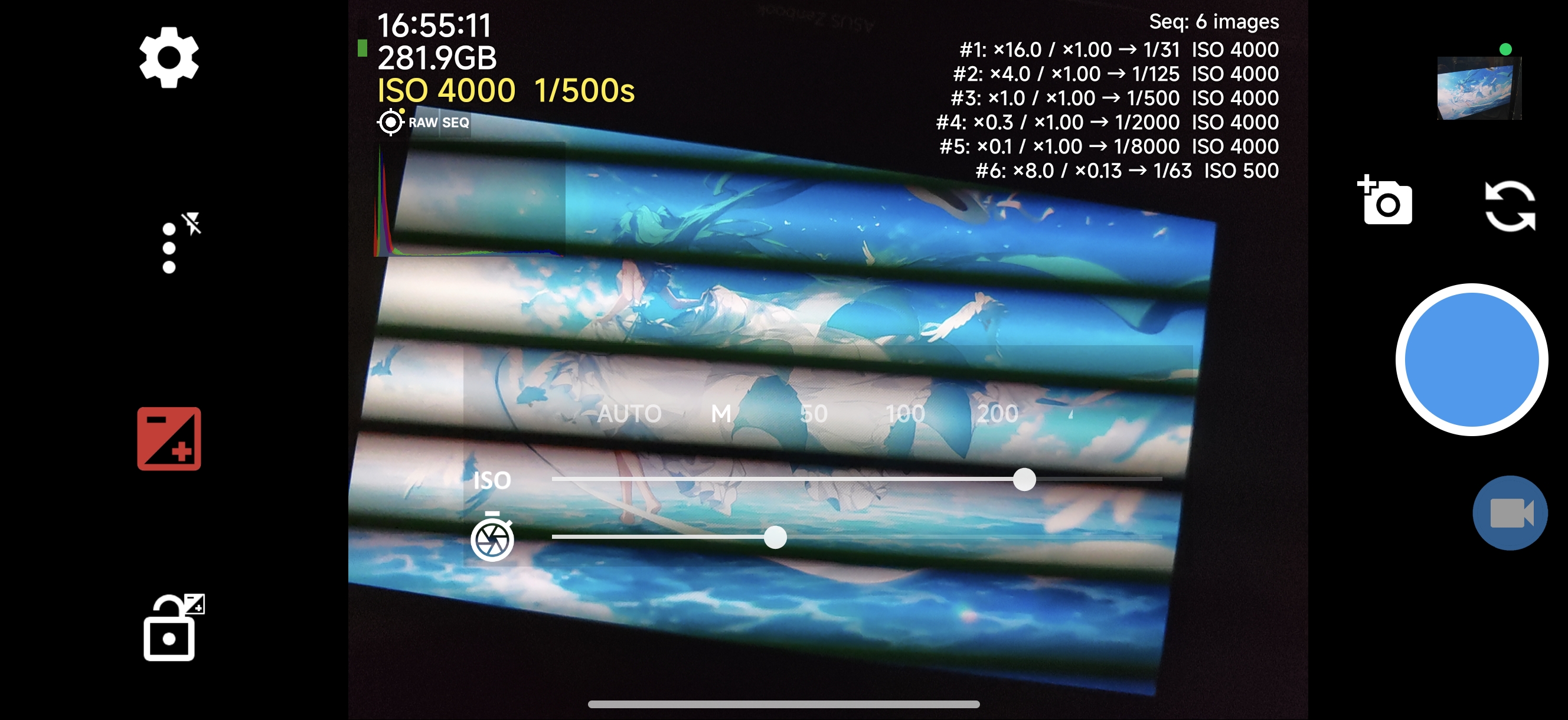}
    \caption{Automated capture tool interface.}
    \label{fig:capture_tool}
\end{figure}

As illustrated in Fig.~\ref{fig:capture_tool}, the tool allows users to define a sequence of capture settings, including exposure time and ISO (as displayed in the top right corner), which are then executed automatically to capture a burst of RAW images with varying exposure parameters. The tool also reports the capture status in real time and lets the sequence be configured before a run. This automation significantly reduces the manual effort required to capture the diverse set of images needed for training and evaluating FB removal.

\section{SFC Sensitivity and Decomposition}
\label{sec:sfc_detail}

We introduce the SFC metric to quantitatively evaluate banding removal in the frequency domain in Sec.~\ref{sec:sfc}.
Recall that SFC combines a direction-ratio loss $\mathcal{D}_{\text{ratio}}$ and a gated radial-profile loss $\mathcal{D}_{\text{profile}}$:
$\text{SFC} = [\,\mathcal{D}_{\text{ratio}} + \lambda_E \cdot g \cdot \mathcal{D}_{\text{profile}}\,] \times 100$.
To provide a more fine-grained diagnosis, we report the decomposed results of $\mathcal{D}_{\text{ratio}}$ and $\mathcal{D}_{\text{profile}}$, with $\Delta_{\text{ratio}}=\rho_{\text{pred}}-\rho_{\text{gt}}$ and $\Delta E_{\text{dir}}=100\,(E_{\text{dir}}(\mathbf{I}_{\text{pred}})/E_{\text{dir}}(\mathbf{I}_{\text{GT}})-1)$, where $E_{\text{dir}}$ is the sector energy. Both are reported as mean $\pm$ standard deviation across the test set.
Intuitively, $\mathcal{D}_{\text{ratio}}$ captures how well the directional energy concentration is preserved, while $\mathcal{D}_{\text{profile}}$ measures the fidelity of the radial energy profile along the banding direction.

To check that SFC tracks banding strength and is insensitive to image content, we capture a real sequence and render its synthetic counterpart at ten exposure times, keeping $\text{ISO} \times T^{\text{expo}}$ constant so that the brightness stays nearly unchanged. Table~\ref{tab:sfc_sweep} lists the crops and SFC values. In both sequences SFC decreases with longer exposure and reaches zero at the FB-free target, following the banding residual. The real sequence has one local step between $1/2{,}000$ s and $1/1{,}525$ s: the former shows wider but shallower FB and the latter shows narrower but deeper stripes, so the two frames are hard to rank even by eye and their scores are effectively tied. The overall decreasing trend remains clear.

Table~\ref{tab:sfc_main} reports the decomposed results for all compared methods, and Table~\ref{tab:sfc_ablation} reports the corresponding ablation variants (see Table~\ref{tab:ablation} in the main text for the configuration of each expression).

\begin{table}[ht]
\centering
\caption{SFC sensitivity to exposure time. The same scene is captured (Real) and rendered (Syn) at ten exposure times with $\text{ISO} \times T^{\text{expo}}$ held constant; column 10 is the FB-free target. SFC follows the banding strength in both sequences.}
\label{tab:sfc_sweep}
\vspace{4pt}
\scriptsize
\setlength{\tabcolsep}{1.2pt}
\renewcommand{\arraystretch}{1.0}
\begin{tabularx}{\textwidth}{@{}>{\raggedright\arraybackslash}m{0.85cm} *{10}{>{\centering\arraybackslash}X}@{}}
\toprule
ID & 1 & 2 & 3 & 4 & 5 & 6 & 7 & 8 & 9 & 10 (GT) \\
\midrule
$T^{\text{expo}}$ & 1/10,000 & 1/8,400 & 1/6,800 & 1/5,200 & 1/3,600 & 1/2,000 & 1/1,525 & 1/1,050 & 1/575 & 1/100 \\
\midrule
Syn & \includegraphics[width=\linewidth]{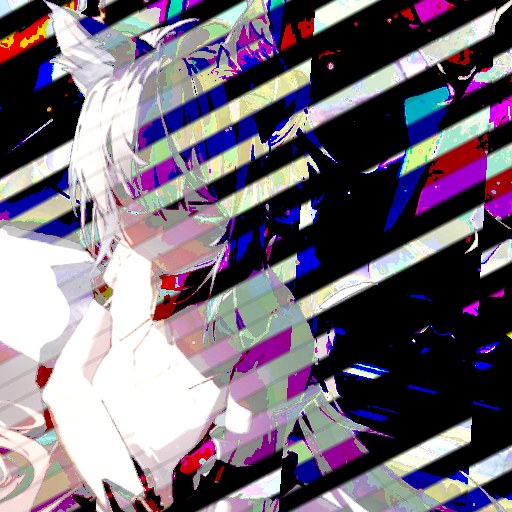} & \includegraphics[width=\linewidth]{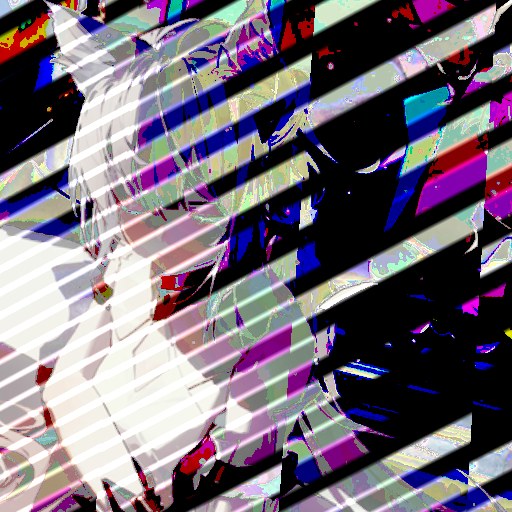} & \includegraphics[width=\linewidth]{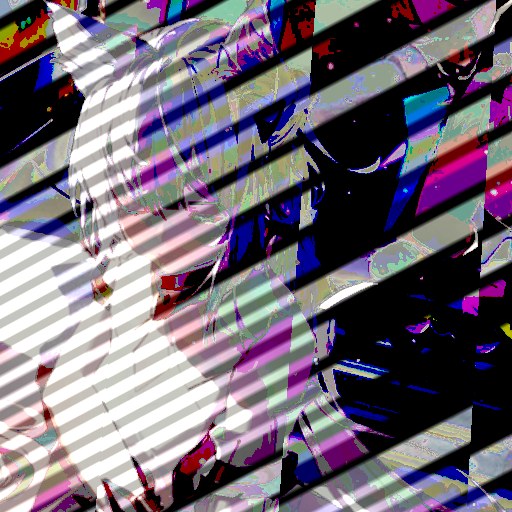} & \includegraphics[width=\linewidth]{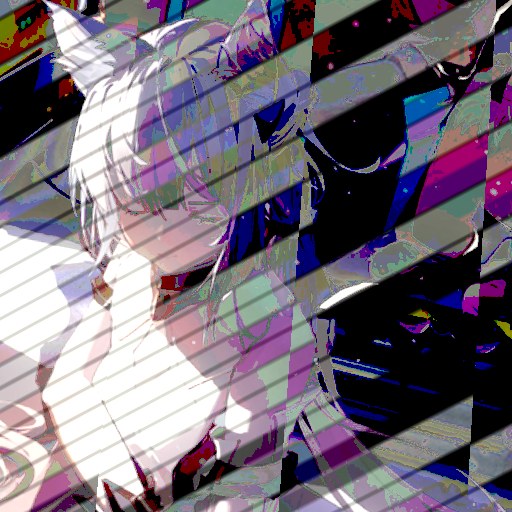} & \includegraphics[width=\linewidth]{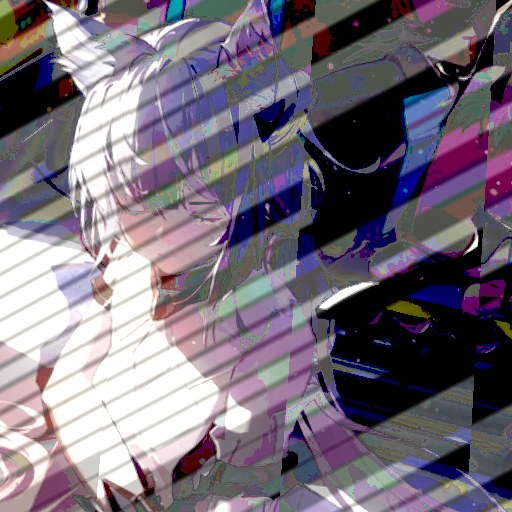} & \includegraphics[width=\linewidth]{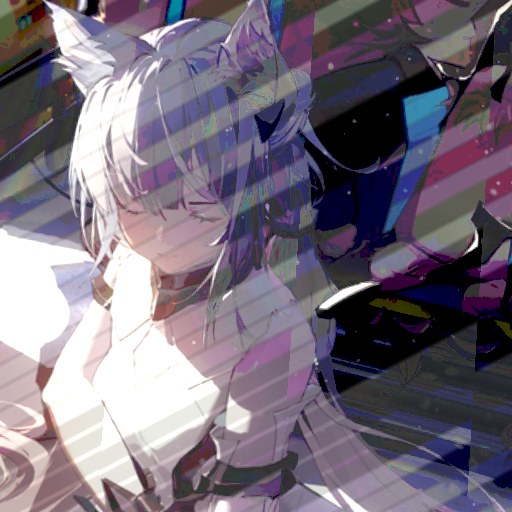} & \includegraphics[width=\linewidth]{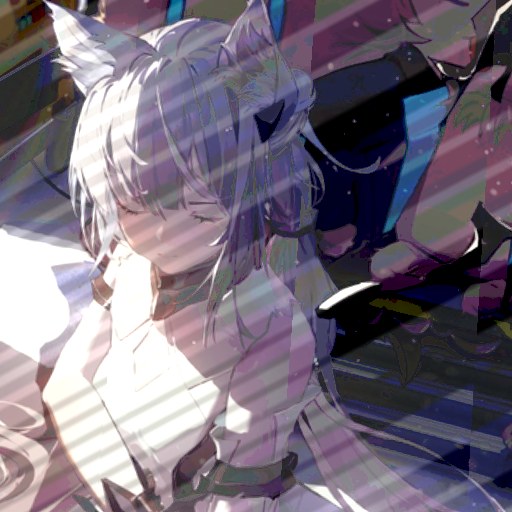} & \includegraphics[width=\linewidth]{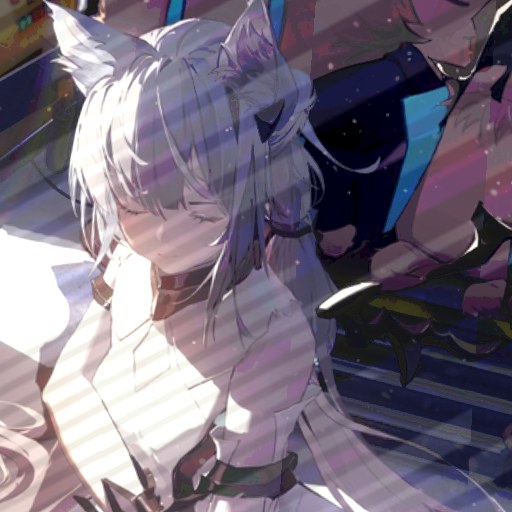} & \includegraphics[width=\linewidth]{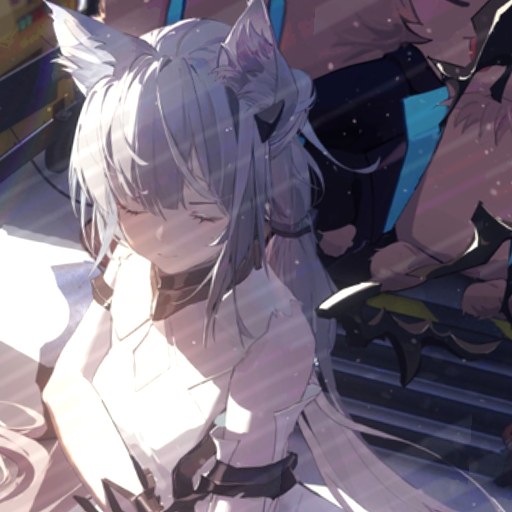} & \includegraphics[width=\linewidth]{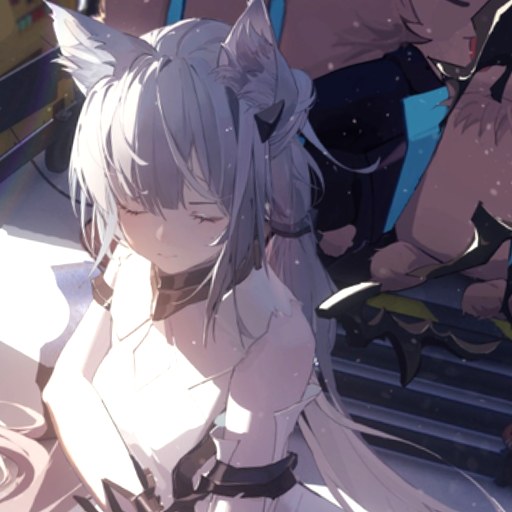} \\
\addlinespace[1.5pt]
SFC & 5.91 & 5.03 & 3.44 & 1.32 & 1.14 & 0.69 & 0.24 & 0.14 & 0.01 & 0.00 \\
\midrule
Real & \includegraphics[width=\linewidth]{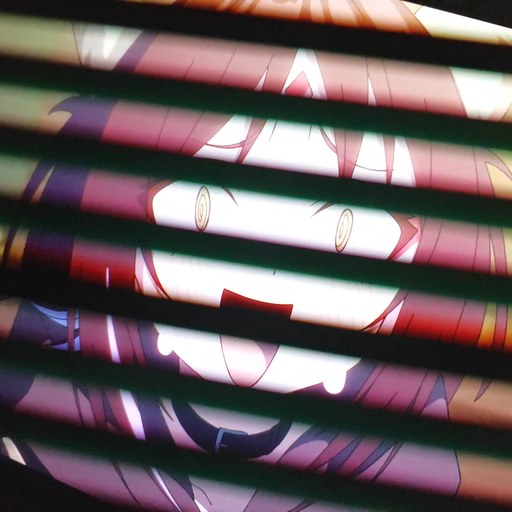} & \includegraphics[width=\linewidth]{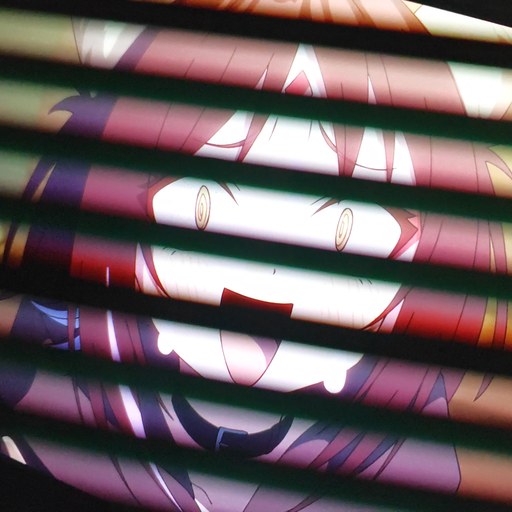} & \includegraphics[width=\linewidth]{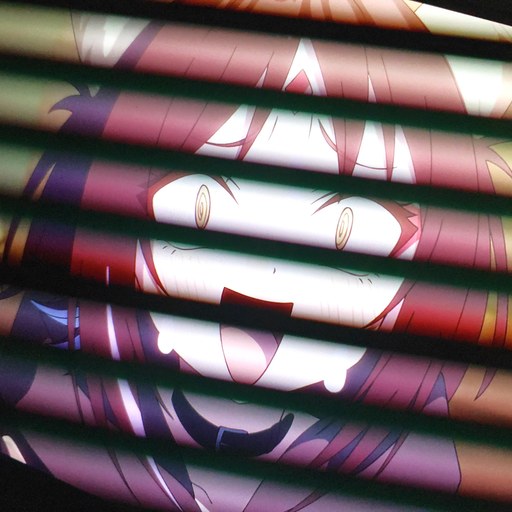} & \includegraphics[width=\linewidth]{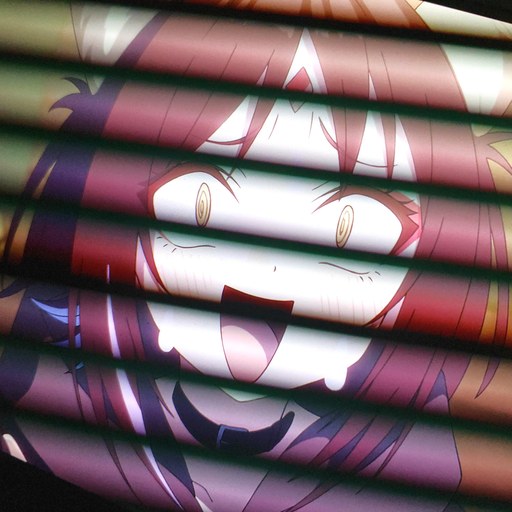} & \includegraphics[width=\linewidth]{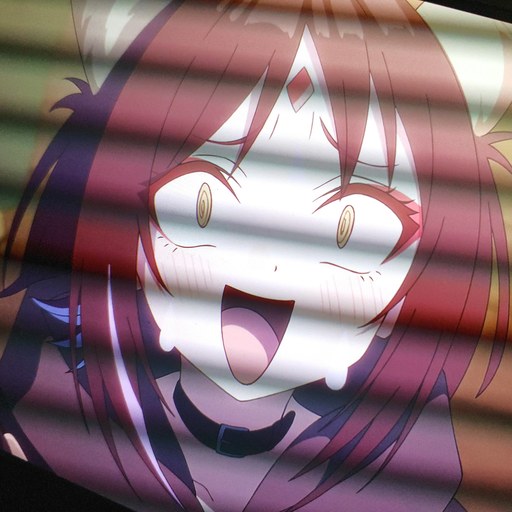} & \includegraphics[width=\linewidth]{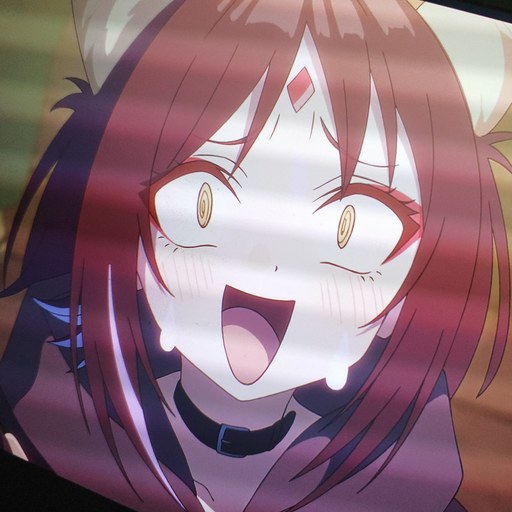} & \includegraphics[width=\linewidth]{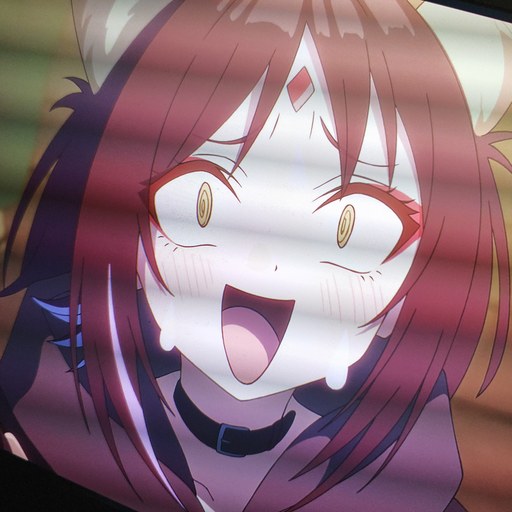} & \includegraphics[width=\linewidth]{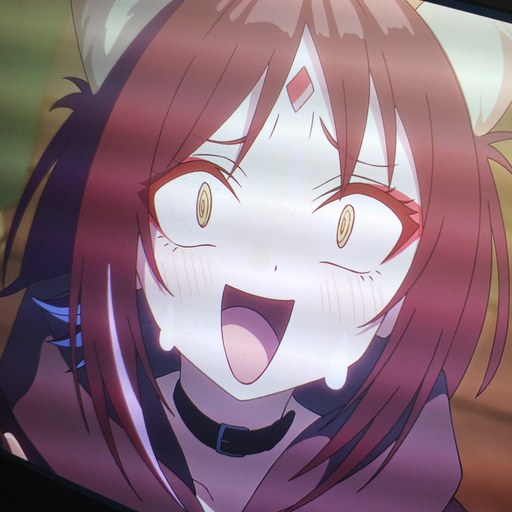} & \includegraphics[width=\linewidth]{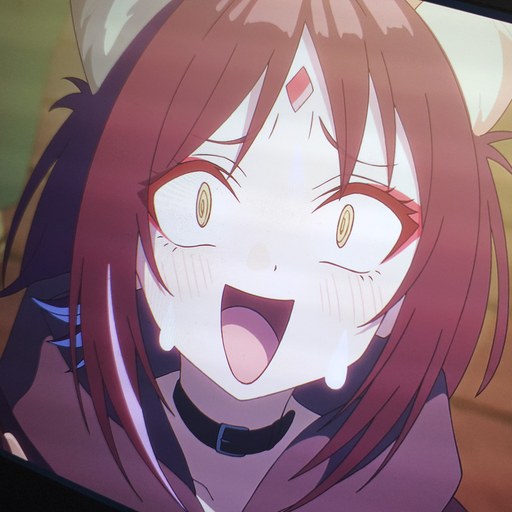} & \includegraphics[width=\linewidth]{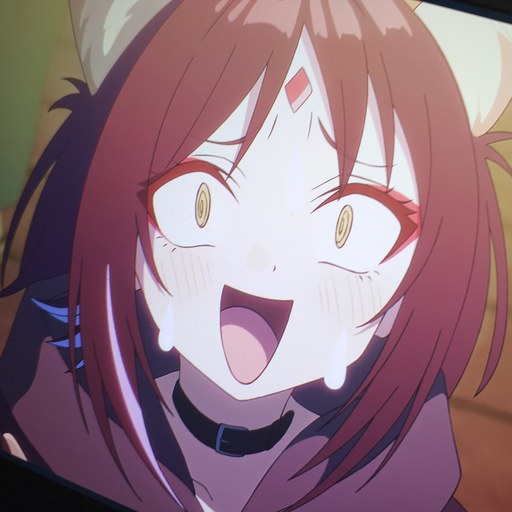} \\
\addlinespace[1.5pt]
SFC & 29.36 & 28.86 & 27.05 & 22.88 & 8.01 & 1.54 & 1.58 & 0.90 & 0.38 & 0.00 \\
\bottomrule
\end{tabularx}
\vspace{-2mm}
\end{table}

\begin{table}[ht]
\centering
\caption{Decomposed SFC results of all compared methods on the Bricker-Real test set. The best and second best results are colored with \best{red} and \secondbest{blue}.}
\vspace{3pt}
\label{tab:sfc_main}
\scriptsize
\begin{tabularx}{\linewidth}{
    *{1}{>{\centering\arraybackslash}p{2.5cm}}
    *{3}{>{\centering\arraybackslash}p{1.0cm}}
    *{4}{>{\centering\arraybackslash}X}
}
\toprule
\multirow{2}{*}{\textbf{Method}}
& \multirow{2}{*}{\textbf{SFC}$\downarrow$}
& \multirow{2}{*}{$\mathcal{D}_{\text{ratio}}$}
& \multirow{2}{*}{$\mathcal{D}_{\text{profile}}$}
& \multicolumn{2}{c}{$\Delta_{\text{ratio}}$}
& \multicolumn{2}{c}{$\Delta E_{\text{dir}}$} \\
\cmidrule(lr){5-6} \cmidrule(lr){7-8}
& & & & {\vspace{-8pt} \tiny $\mu$} & {\vspace{-8pt} \tiny $\sigma$} & {\vspace{-8pt} \tiny $\mu$} & {\vspace{-8pt} \tiny $\sigma$} \\ [-3pt]
\midrule
BRACE (ours)          & \best{0.1847}    & \best{0.1760}    & \best{0.0604}    & $+$0.0915 & 0.267  & $-$0.2436 & 2.411 \\
FlickerFormer~\cite{qu2026ittakestwo} & \secondbest{0.4036} & \secondbest{0.3623} & 0.1724 & $+$0.2339 & 0.941  & $+$4.8366 & 21.309 \\
Burstormer~\cite{dudhane2023burstormer}    & 0.5153           & 0.4949           & \secondbest{0.1203} & $+$0.4108 & 0.776  & $-$0.4643 & 5.910 \\
TMRNet~\cite{Zhang2025BracketIRE}    & 1.1135           & 0.9415           & 0.5899 & $+$0.7518 & 1.509  & $+$23.3748 & 92.957 \\
HDRFlow~\cite{xu2024hdrflow}       & 17.6967          & 3.7265           & 14.9219 & $+$0.4853 & 5.844  & $+$678.6443 & 2593.946 \\
\bottomrule
\end{tabularx}
\vspace{-1mm}
\end{table}

\begin{table}[ht]
\centering
\caption{Decomposed SFC results of ablation variants on the Bricker-Real test set (40 scenes). The best and second best results are colored with \best{red} and \secondbest{blue}.}
\vspace{3pt}
\label{tab:sfc_ablation}
\scriptsize
\begin{tabularx}{\linewidth}{
    *{1}{>{\centering\arraybackslash}p{2.5cm}}
    *{3}{>{\centering\arraybackslash}p{1.0cm}}
    *{4}{>{\centering\arraybackslash}X}
}
\toprule
\multirow{2}{*}{\textbf{Expr.}}
& \multirow{2}{*}{\textbf{SFC}$\downarrow$}
& \multirow{2}{*}{$\mathcal{D}_{\text{ratio}}$}
& \multirow{2}{*}{$\mathcal{D}_{\text{profile}}$}
& \multicolumn{2}{c}{$\Delta_{\text{ratio}}$}
& \multicolumn{2}{c}{$\Delta E_{\text{dir}}$} \\
\cmidrule(lr){5-6} \cmidrule(lr){7-8}
& & & & {\vspace{-8pt} \tiny $\mu$} & {\vspace{-8pt} \tiny $\sigma$} & {\vspace{-8pt} \tiny $\mu$} & {\vspace{-8pt} \tiny $\sigma$} \\ [-3pt]
\midrule
7 & \best{0.1847}    & \best{0.1760}    & \best{0.0604}    & $+$0.0915  & 0.267   & $-$0.2436  & 2.411 \\
5 & \secondbest{0.2458} & \secondbest{0.2339} & 0.0973  & $+$0.1433  & 0.371   & $+$1.2199  & 9.005 \\
6 & 0.2629           & 0.2523           & \secondbest{0.0701} & $+$0.0793  & 0.445   & $-$0.2478  & 2.914 \\
3 & 0.7071           & 0.6248           & 0.3413  & $+$0.4883  & 1.045   & $+$13.3978 & 51.244 \\
4 & 0.7984           & 0.7125           & 0.3628  & $+$0.6628  & 1.032   & $+$14.2394 & 55.438 \\
2 & 0.8202           & 0.7621           & 0.3036  & $+$0.5416  & 1.127   & $+$9.4589  & 28.858 \\
1 & 1.0612           & 0.8298           & 0.6530  & $+$0.6757  & 1.437   & $+$27.5407 & 133.254 \\
\bottomrule
\end{tabularx}
\vspace{-1mm}
\end{table}

\section{Bricker: Statistics and Content Sources}
\label{sec:app_statistics}

Table~\ref{tab:app_stats} summarizes the two subsets. Bricker-Real is deliberately small because paired LQ--GT captures require stringent conditions, whereas Bricker-Syn covers a wider range of camera exposure parameters, spatial configurations, display types, and PWM settings. RAW parameters are strongly device-dependent, so the synthetic subset is simulated for one target device and the real subset uses a single phone and two LED matrix displays; the experiments therefore evaluate within-device adaptation rather than cross-device generalization.

The content spans six categories taken from eight public sources: natural scenes (LSDIR~\cite{LSDIR}), documents (DocBank~\cite{li2020docbank} and PubLayNet~\cite{zhong2019publaynet}), anime illustrations (Danbooru), video games (VideoGameBunny~\cite{taesiri2024videogamebunnyvisionassistantsvideo}), mobile interfaces (Rico~\cite{deka2017rico}), and webpages (DSCVC~\cite{wang2025dscvc} and screen-content-dataset~\cite{scvqa2025}). An IQA-based filter removes low-quality content, and the displayed content of the training and test splits does not overlap.

Because the GT frame needs a longer exposure and a lower ISO, the lighting is controlled: there is no overexposure from the prolonged exposure and no external reflection or glare. The real training and test sets share the same capture device and display, which is the setting of interest for practical fine-tuning.

\begin{table}[ht]
\centering
\vspace{-2pt}
\caption{Statistics of the two subsets of Bricker. A sequence contains five bracketed frames and one ground-truth frame.}
\label{tab:app_stats}
\vspace{3pt}
\small
\begin{tabular}{@{}lcc@{}}
\toprule
 & Syn & Real \\
\midrule
Camera & global / rolling shutter, varied direction and speed & Xiaomi 17 Pro \\
Display & OLED / LED, varied PWM frequency and scanning direction & 2 LED matrix displays \\
Train & $1000 \times (5+1)$ & $250 \times (5+1)$ \\
Test & $100 \times (5+1)$ & $40 \times (5+1)$ \\
Resolution & $1024 \times 1024$ & $1024 \times 768$ \\
\bottomrule
\end{tabular}
\end{table}

\section{Ground-Truth Validation on Bricker-Real}
\label{sec:app_gt_validation}

During capture we minimize the time gap between the reference frame and the GT by adjusting the sequence settings and holding the camera steady (Fig.~\ref{fig:real_capture}.a), so that the two frames are aligned in position and overall brightness.

FB cannot be measured directly on a pair that has no FB-free reference, so we follow the observation of RIFLE~\cite{Zhu2025RIFLE} that gray FB stripes reside mainly in the Y channel. After converting the reference frame and the GT to YCbCr, the Y residual therefore reflects residual banding, while the Cb and Cr residuals mainly reflect misalignment and chromatic inconsistency. On the 16 test pairs with the mildest color shift, all channel means are close to zero (Table~\ref{tab:app_gt_mean}) while the Y residual deviation clearly dominates (Table~\ref{tab:app_gt}), consistent with FB being a luminance phenomenon.

Table~\ref{tab:app_gt} repeats the deviations on the remaining pairs and on the full test set. Replacing the Y channel of the LQ image with that of the GT gives PSNR 35.04 and SSIM 0.9199 on the selected pairs, which indicates that they are close to pixel-aligned and chromatically consistent. On the remaining pairs, where the color shift is stronger, the same replacement only reaches 25.72 dB, so the residual discrepancy there is chromatically dominated and is not removed by the luminance substitution.

\begin{table}[ht]
\centering
\vspace{-5pt}
\begin{minipage}[t]{0.36\textwidth}
\centering
\captionof{table}{Residual means on the 16 selected pairs (YCbCr, 0--255): mean and std of the per-pair channel means.}
\vspace{3pt}
\label{tab:app_gt_mean}
\small
\begin{tabular}{@{}lcc@{}}
\toprule
 & Mean & Std \\
\midrule
$\Delta$Y & $-0.48$ & $1.00$ \\
$\Delta$Cb & $0.24$ & $1.52$ \\
$\Delta$Cr & $-0.27$ & $1.38$ \\
\bottomrule
\end{tabular}
\end{minipage}\hfill
\begin{minipage}[t]{0.60\textwidth}
\centering
\captionof{table}{Residual standard deviations on Bricker-Real, and the RGB PSNR obtained after replacing the Y channel of the LQ image with that of the GT. Values are mean $\pm$~std over the pairs.}
\label{tab:app_gt}
\vspace{3pt}
\small
\begin{tabular}{@{}lccc@{}}
\toprule
 & Selected (16) & Remaining (24) & All (40) \\
\midrule
$\Delta$Y std & $9.52 \pm 4.18$ & $14.18 \pm 5.70$ & $12.32 \pm 5.59$ \\
$\Delta$Cb std & $3.12 \pm 1.96$ & $9.19 \pm 4.32$ & $6.76 \pm 4.64$ \\
$\Delta$Cr std & $4.02 \pm 3.15$ & $10.03 \pm 4.61$ & $7.62 \pm 5.02$ \\
$\text{PSNR}_\text{Y$\leftarrow$GT}$ & $35.04 \pm 5.07$ & $25.72 \pm 3.92$ & $29.45 \pm 6.35$ \\
\bottomrule
\end{tabular}
\end{minipage}
\end{table}

\vspace{-2mm}
\section{Frame Count and Input Settings}
\label{sec:app_additional}

\textbf{Number of frames.} Without changing the architecture, we train variants that take one frame (the reference only) or three frames ($1/4\times$, $1\times$, $4\times$) and compare them with the full five-frame model in Table~\ref{tab:app_frames}. Accuracy improves consistently with the number of exposures, which indicates that the gain comes from the complementary observations rather than from the reference frame alone.

\begin{table}[ht]
\centering
\vspace{-5pt}
\caption{Effect of the number of input frames under the same architecture.}
\label{tab:app_frames}
\vspace{3pt}
\small
\setlength{\tabcolsep}{2.4pt}
\begin{tabular}{@{}lcccccccccc@{}}
\toprule
& \multicolumn{5}{c}{Syn} & \multicolumn{5}{c}{Real} \\
\cmidrule(lr){2-6} \cmidrule(lr){7-11}
Frames & PSNR$\uparrow$ & SSIM$\uparrow$ & LPIPS$\downarrow$ & DISTS$\downarrow$ & MSSWD$\downarrow$
& PSNR$\uparrow$ & SSIM$\uparrow$ & LPIPS$\downarrow$ & DISTS$\downarrow$ & MSSWD$\downarrow$ \\
\midrule
1 & 26.10 & 0.8286 & 0.2630 & 0.1940 & 0.7832 & 26.38 & 0.7718 & 0.2273 & 0.1286 & 0.3815 \\
3 & 27.50 & 0.8699 & 0.2180 & 0.1692 & 0.6320 & 27.48 & 0.7913 & 0.2094 & 0.1194 & 0.4127 \\
5 & \textbf{28.44} & \textbf{0.8823} & \textbf{0.2050} & \textbf{0.1606} & \textbf{0.5102} & \textbf{28.15} & \textbf{0.8007} & \textbf{0.1925} & \textbf{0.1113} & \textbf{0.2052} \\
\bottomrule
\end{tabular}
\end{table}

\textbf{Uniform exposure.} Since FB intensity barely varies between frames of the same exposure, a uniform-exposure burst carries little complementary information. Retraining on uniform-exposure synthetic data gives 26.37 dB PSNR, slightly above the single-frame input and well below the bracketed five-frame model (Table~\ref{tab:app_uniform}). SFC is not reported for this variant because the metric is defined through the exposure-dependent banding direction.

\begin{table}[ht]
\centering
\vspace{-5pt}
\caption{Uniform-exposure input on Bricker-Syn, with the same five-frame budget as BRACE. SFC is not applicable to the uniform-exposure variant.}
\label{tab:app_uniform}
\vspace{3pt}
\small
\begin{tabular}{@{}lccccc@{}}
\toprule
Input & PSNR$\uparrow$ & SSIM$\uparrow$ & LPIPS$\downarrow$ & DISTS$\downarrow$ & MSSWD$\downarrow$ \\
\midrule
1 frame (reference only) & 26.10 & 0.8286 & 0.2630 & 0.1940 & 0.7832 \\
Uniform exposure (5 frames) & 26.37 & 0.8443 & 0.2440 & 0.1814 & 0.6289 \\
Bracketed (5 frames, BRACE) & \textbf{28.44} & \textbf{0.8823} & \textbf{0.2050} & \textbf{0.1606} & \textbf{0.5102} \\
\bottomrule
\end{tabular}
\end{table}

\textbf{Frame count of FlickerFormer.} FlickerFormer hard-codes a three-frame input. After adapting it to accept all five frames and retraining, its PSNR changes from 24.86 to 24.77 dB on Bricker-Syn and from 25.49 to 25.93 dB on Bricker-Real, so the comparison with BRACE is not driven by the number of input frames.

\section{Computational Cost}
\label{sec:app_cost}

We measure the cost on an NVIDIA RTX A6000 in fp16 with an input of five frames of size $512 \times 512 \times 4$, corresponding to $1024 \times 1024$ after demosaicing (Table~\ref{tab:app_cost}). The dense residual computation dominates the FLOPs, and the memory footprint has not been optimized; operator fusion, pruning, or distillation would be needed for on-device use. Further work is needed for practical mobile deployment.

\begin{table}[ht]
\centering
\caption{Computational cost of BRACE on an RTX A6000 (fp16, five frames of $512\times512\times4$).}
\label{tab:app_cost}
\vspace{3pt}
\small
\begin{tabular}{@{}lc@{}}
\toprule
Metric & Value \\
\midrule
Parameters & 13.32 M \\
Mean latency & 448.03 ms \\
Inference memory & 3267.6 MiB \\
FLOPs & 5897.9 G \\
\bottomrule
\end{tabular}
\end{table}

\section{Implementation Details of Compared Methods}
\label{sec:app_implementation_details}

\textbf{Burstormer~\cite{dudhane2023burstormer}.} We adopt the burst denoising variant released by the authors, which matches our restoration task more closely than the super-resolution variant. 5 RAW LQ frames are fed into the model to predict the GT RGB image. For the synthetic test set we train it on the Bricker-Syn training set with the default hyperparameters, and for the real test set we use the Bricker-Syn and Bricker-Real training sets and train from scratch with the same settings.

\textbf{HDRFlow~\cite{xu2024hdrflow}.} 5 post-processed RGB LQ frames are treated as a video sequence and fed into the model to predict the GT image. 
For the synthetic test set, we train the model with the Bricker-Syn training set for 100 epochs with a batch size of 32, lr of 0.001. For the real test set, we use the Bricker-Syn and Bricker-Real training sets and train from scratch with the same hyperparameters.

\textbf{TMRNet~\cite{Zhang2025BracketIRE}.} 5 RAW LQ frames are directly fed into the model to predict the GT RAW image, which is then post-processed with the same ISP pipeline as the input frames to obtain the final RGB output. For the synthetic test set, we train the model with the Bricker-Syn training set for 50 epochs with a batch size of 16, lr of 0.0001. For the real test set, we apply TMRNet's self-supervised domain adaptation method on the Bricker-Real training set for 20 epochs.

\textbf{FlickerFormer~\cite{qu2026ittakestwo}.} The middle 3 post-processed RGB LQ frames (idx 2, 3, 4) are treated as the input sequence and fed into the model. Input image size is 256x256. For the synthetic test set, we train the model with the Bricker-Syn training set for 200000 iterations using its default hyperparameters. For the real test set, we use the Bricker-Syn and Bricker-Real training sets and train from scratch with the same hyperparameters.

\begin{figure}[h]
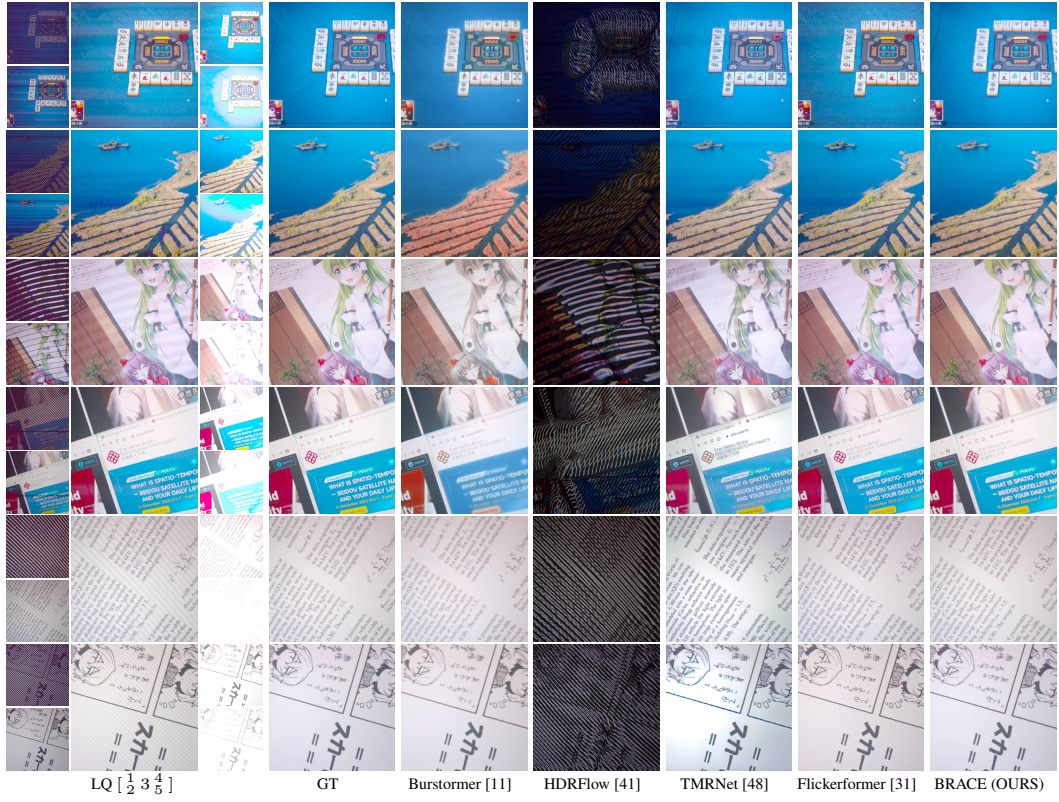

    \centering
    \insertcomparison{000007_web2}
    \insertcomparison{000009_static}
    \insertcomparison{000010_anime}
    \insertcomparison{000012_web2}
    \insertcomparison{000014_doc2}
    \insertcomparison{000019_anime}
    \insertmethodlegend
    \caption{Additional visual comparison on Bricker-Real.}
    \label{fig:app_qual_results}
\end{figure}

\begin{figure}[h]
    \centering
    \insertcomparison{syn/000002_UI}
    \insertcomparison{syn/000015_anime}
    \insertcomparison{syn/000020_anime}
    \insertcomparison{syn/000025_web2}
    \insertmethodlegend
    \caption{Additional visual comparison on Bricker-Syn.}
    \label{fig:app_qual_results_syn}
\end{figure}

\section{Additional Visual Comparison}
\label{sec:app_visual_comparison}

Fig.~\ref{fig:app_qual_results} presents additional visual comparisons on the Bricker-Real test set, and Fig.~\ref{fig:app_qual_results_syn} presents additional visual comparisons on the Bricker-Syn test set.

\section{Bricker Dataset Visualization}
\label{sec:app_bricker_visualization}

\begin{figure}[h]
    \centering
    \begin{subfigure}{0.155\linewidth}
        \centering
        \includegraphics[width=\linewidth]{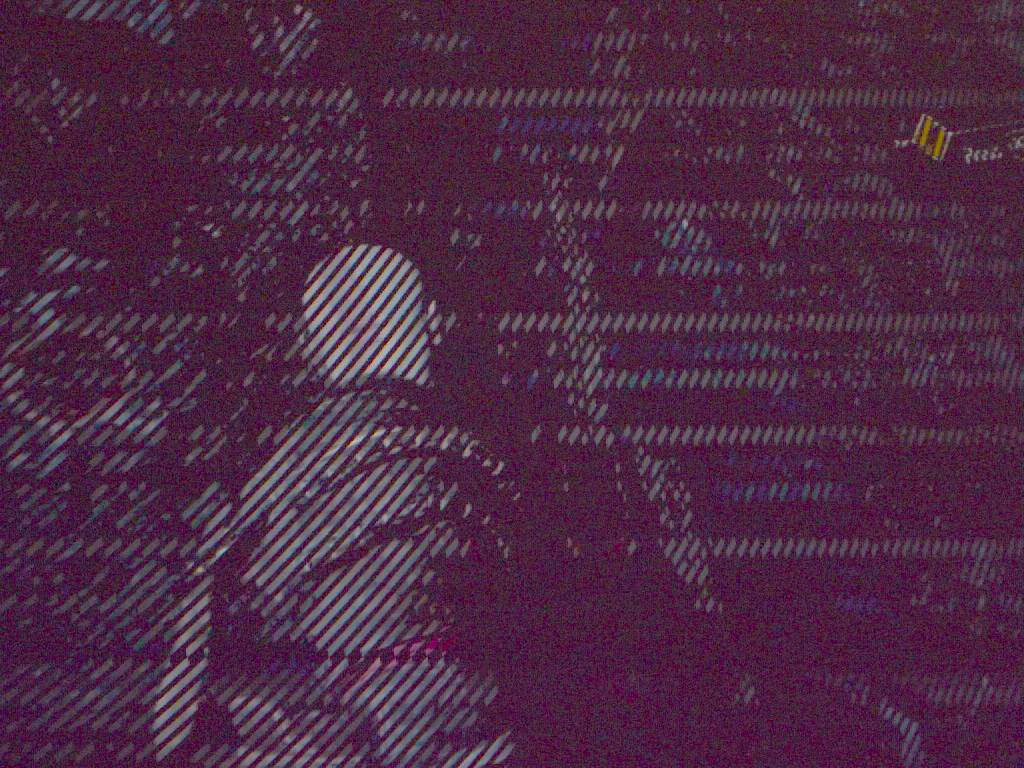}
    \end{subfigure}
    \begin{subfigure}{0.155\linewidth}
        \centering
        \includegraphics[width=\linewidth]{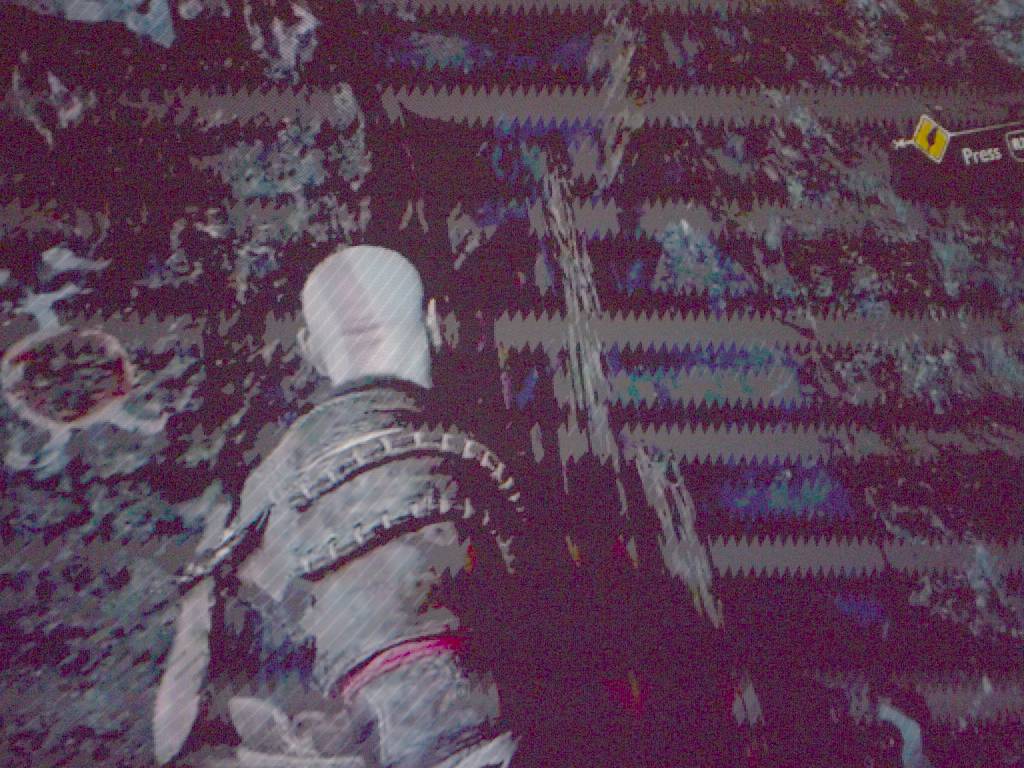}
    \end{subfigure}
    \begin{subfigure}{0.155\linewidth}
        \centering
        \includegraphics[width=\linewidth]{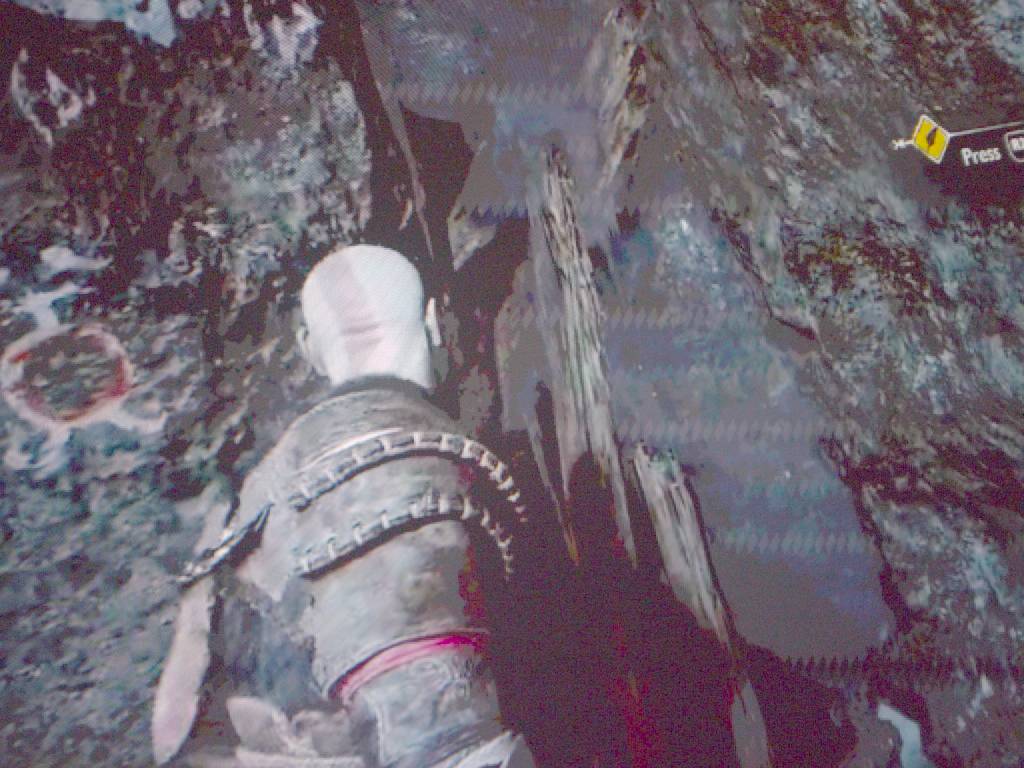}
    \end{subfigure}
    \begin{subfigure}{0.155\linewidth}
        \centering
        \includegraphics[width=\linewidth]{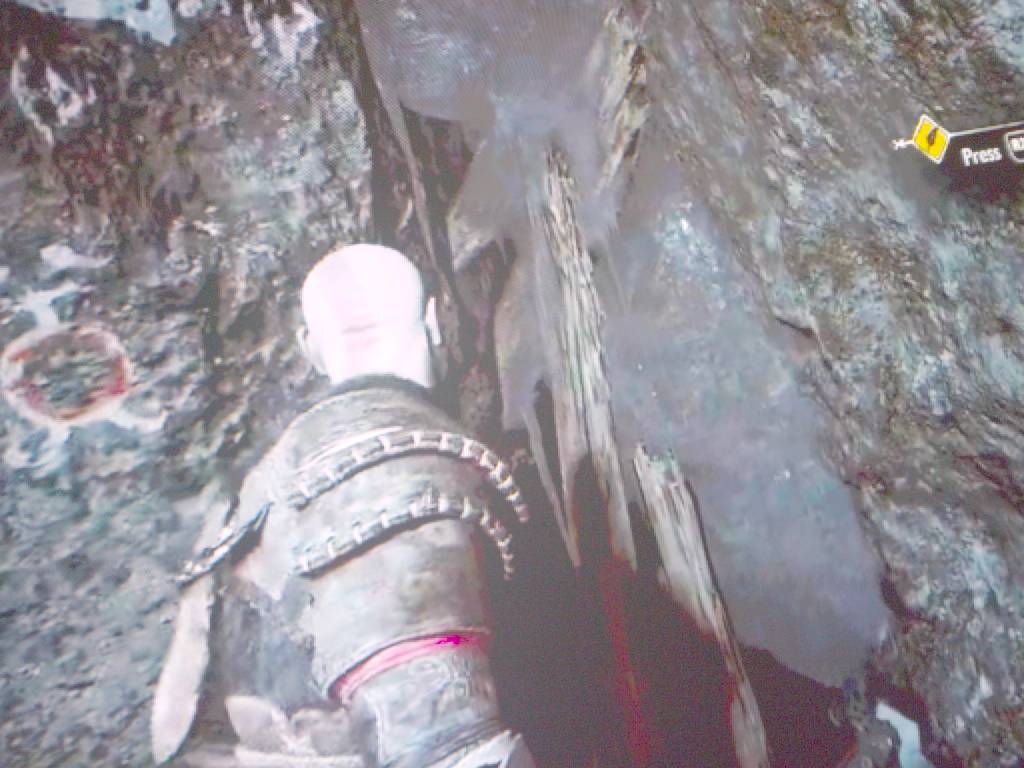}
    \end{subfigure}
    \begin{subfigure}{0.155\linewidth}
        \centering
        \includegraphics[width=\linewidth]{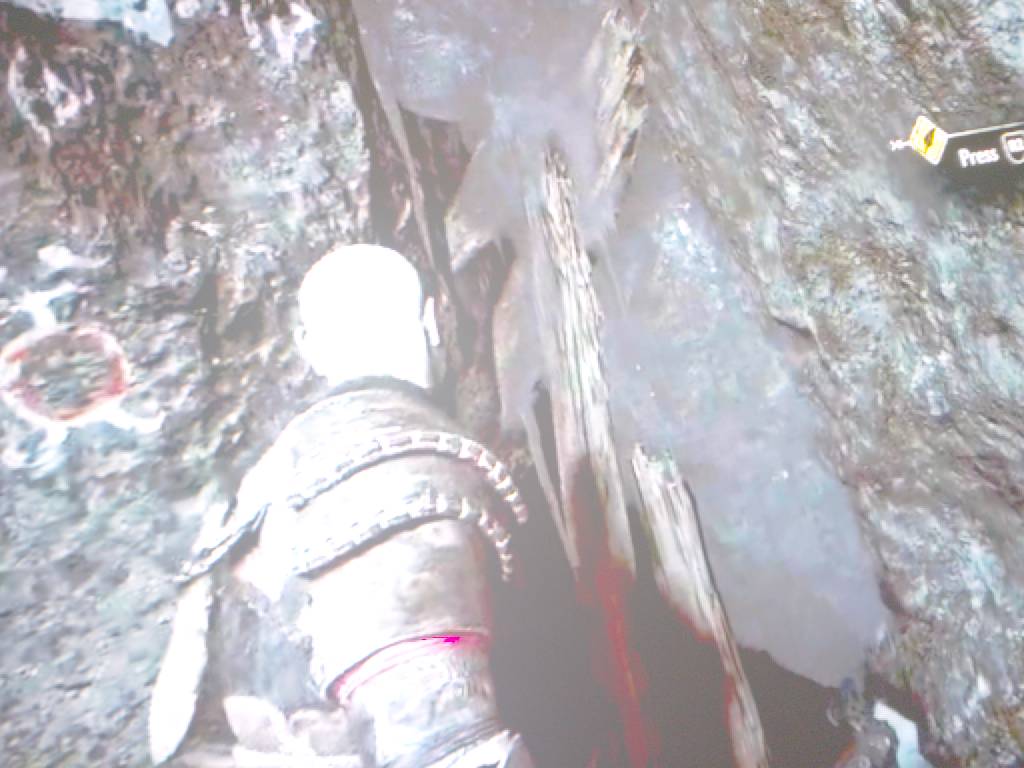}
    \end{subfigure}
    \begin{subfigure}{0.155\linewidth}
        \centering
        \includegraphics[width=\linewidth]{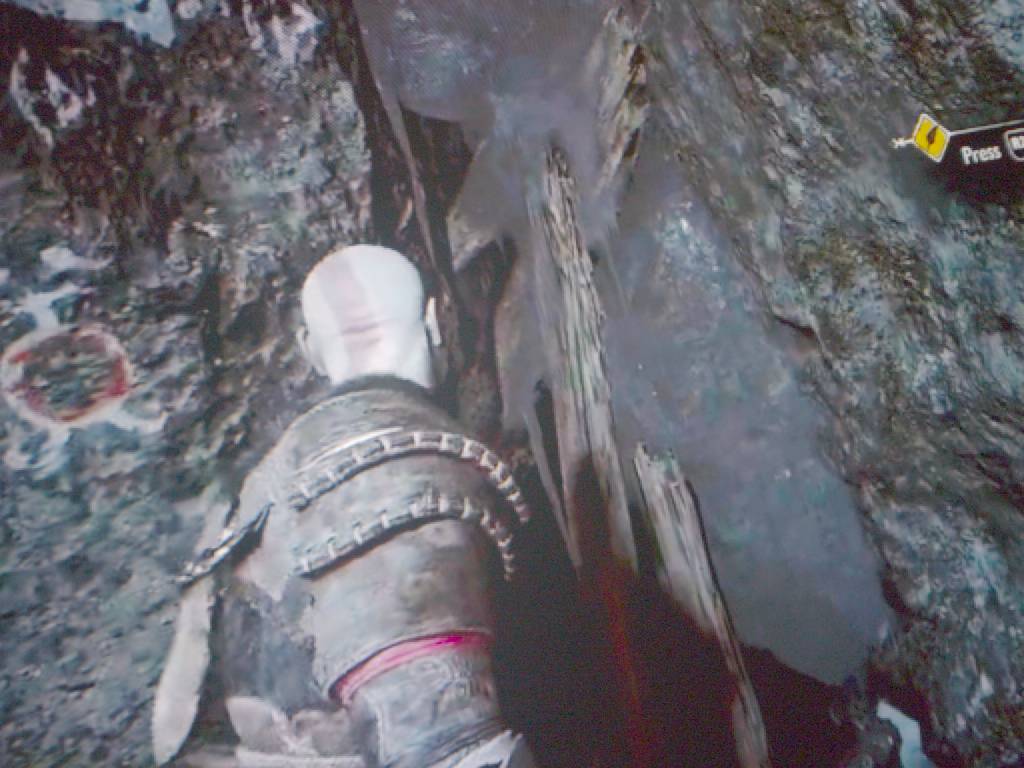}
    \end{subfigure}

    \begin{subfigure}{0.155\linewidth}
        \centering
        \includegraphics[width=\linewidth]{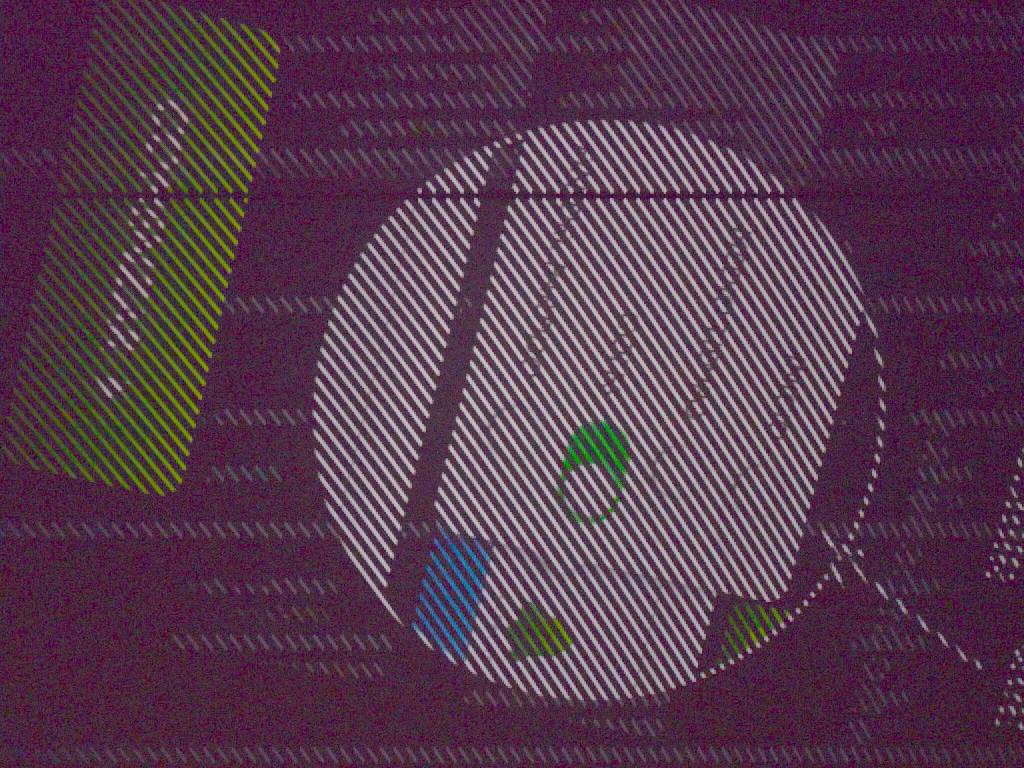}
    \end{subfigure}
    \begin{subfigure}{0.155\linewidth}
        \centering
        \includegraphics[width=\linewidth]{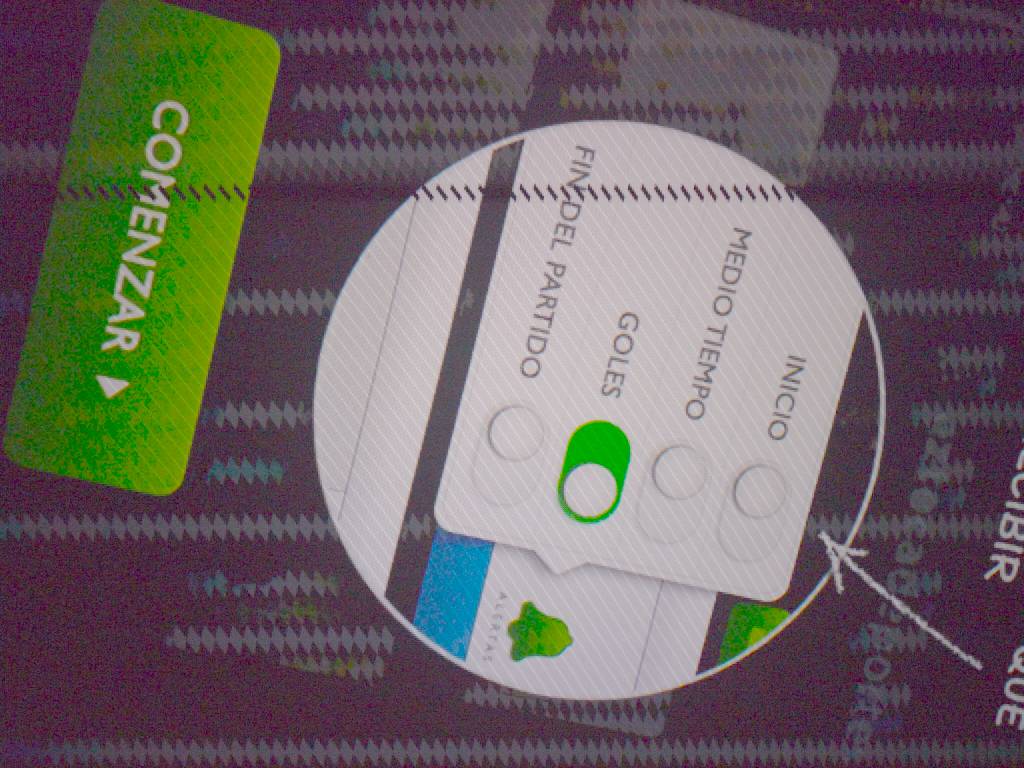}
    \end{subfigure}
    \begin{subfigure}{0.155\linewidth}
        \centering
        \includegraphics[width=\linewidth]{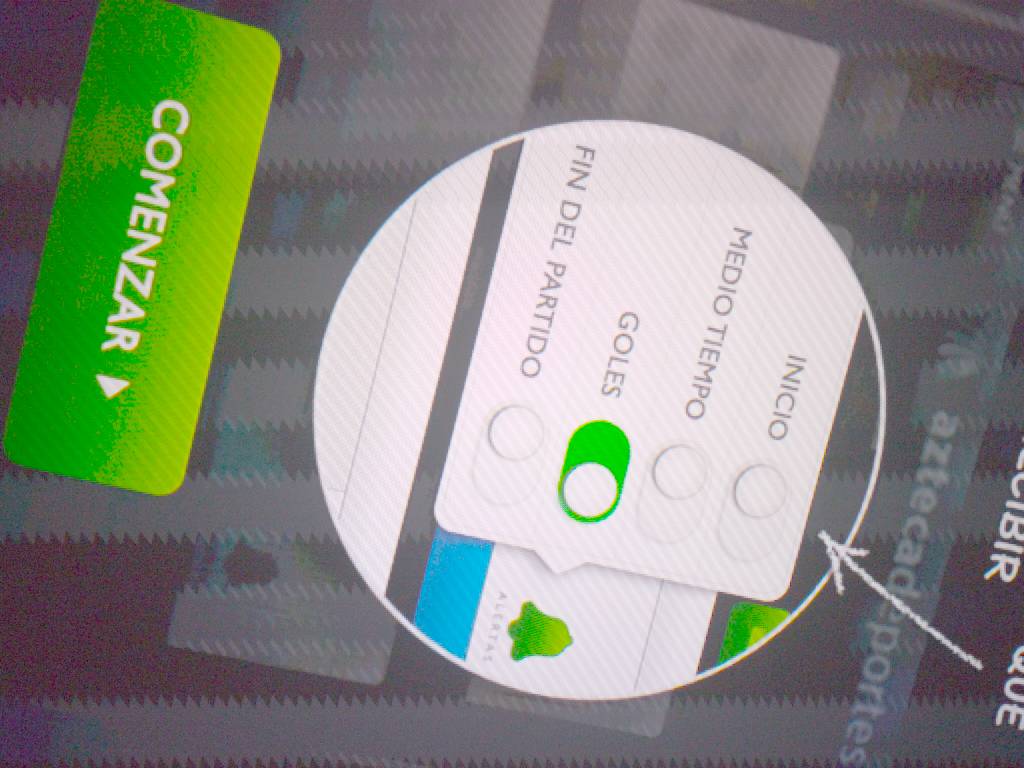}
    \end{subfigure}
    \begin{subfigure}{0.155\linewidth}
        \centering
        \includegraphics[width=\linewidth]{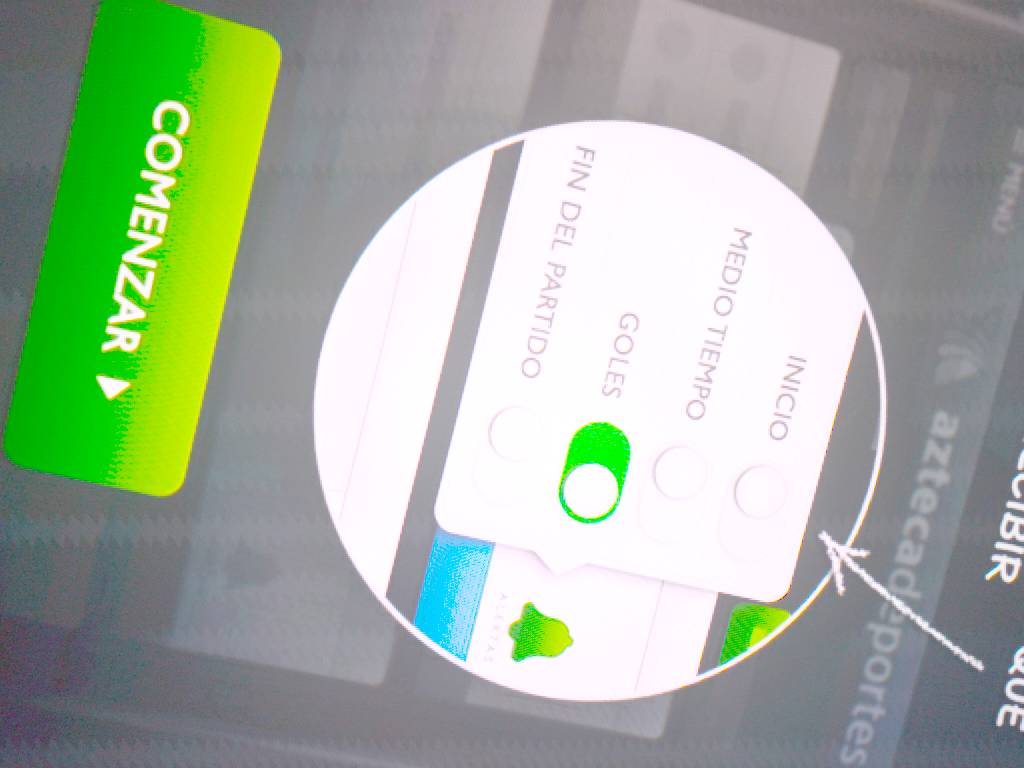}
    \end{subfigure}
    \begin{subfigure}{0.155\linewidth}
        \centering
        \includegraphics[width=\linewidth]{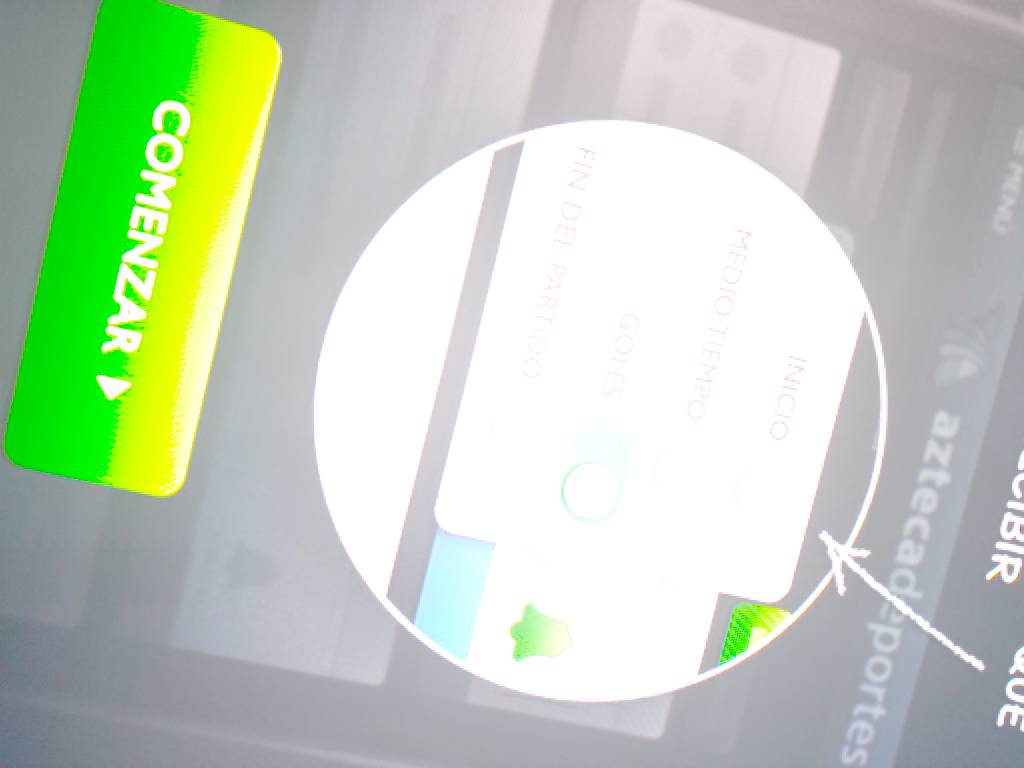}
    \end{subfigure}
    \begin{subfigure}{0.155\linewidth}
        \centering
        \includegraphics[width=\linewidth]{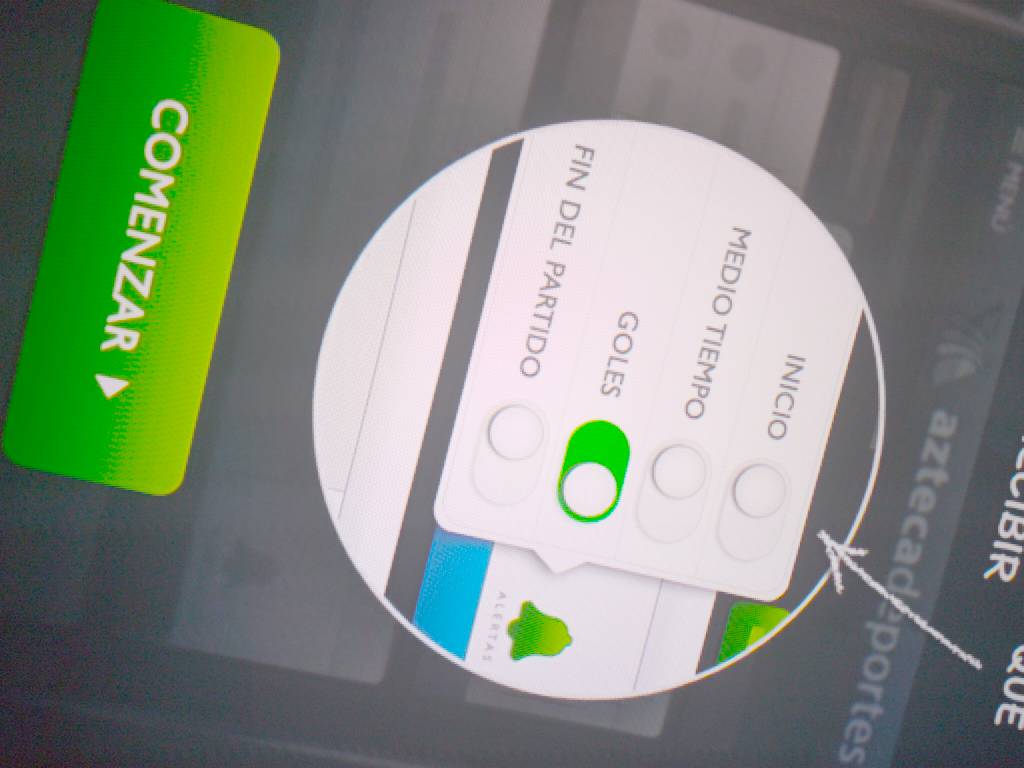}
    \end{subfigure}

    \begin{subfigure}{0.155\linewidth}
        \centering
        \includegraphics[width=\linewidth]{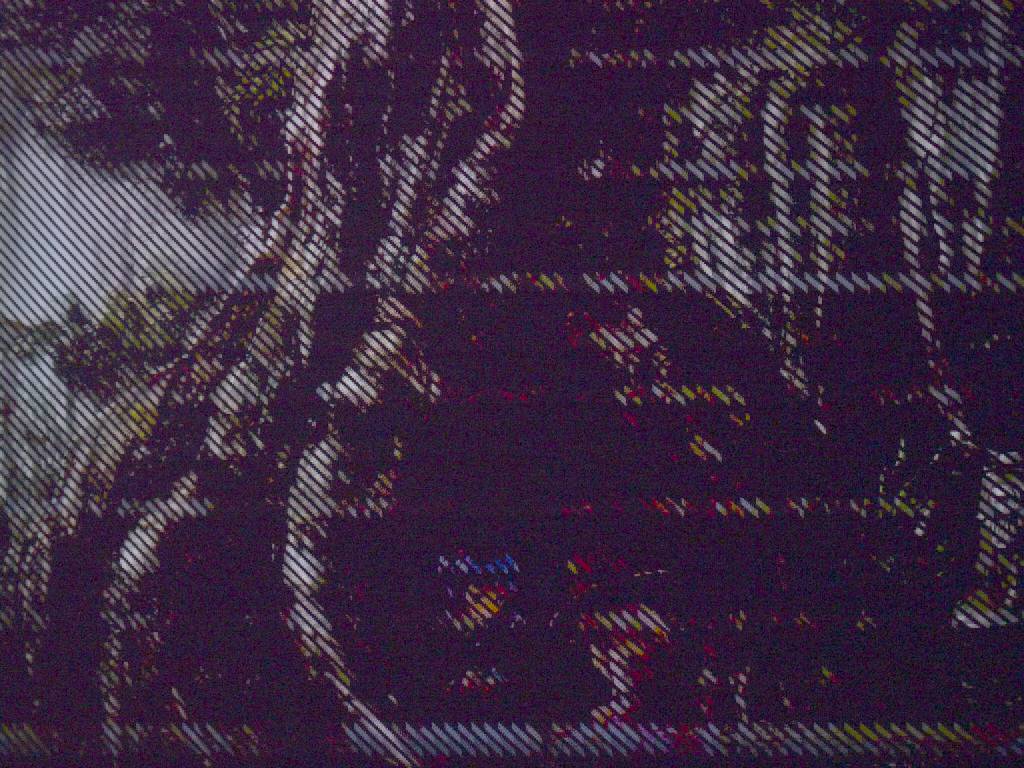}
    \end{subfigure}
    \begin{subfigure}{0.155\linewidth}
        \centering
        \includegraphics[width=\linewidth]{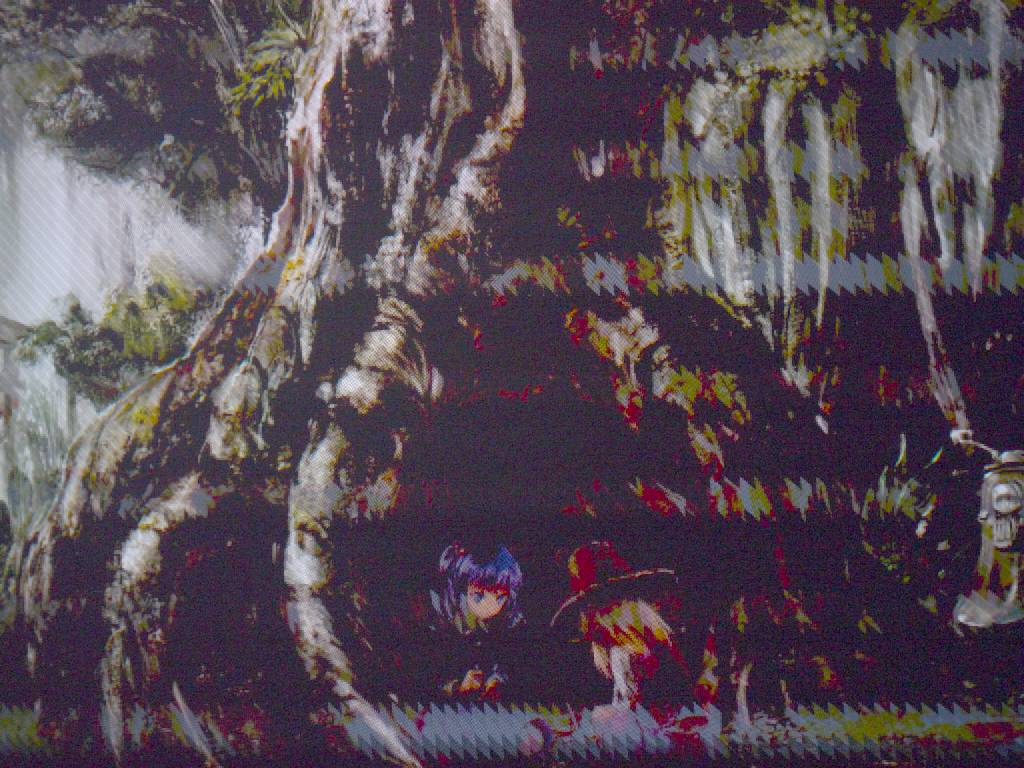}
    \end{subfigure}
    \begin{subfigure}{0.155\linewidth}
        \centering
        \includegraphics[width=\linewidth]{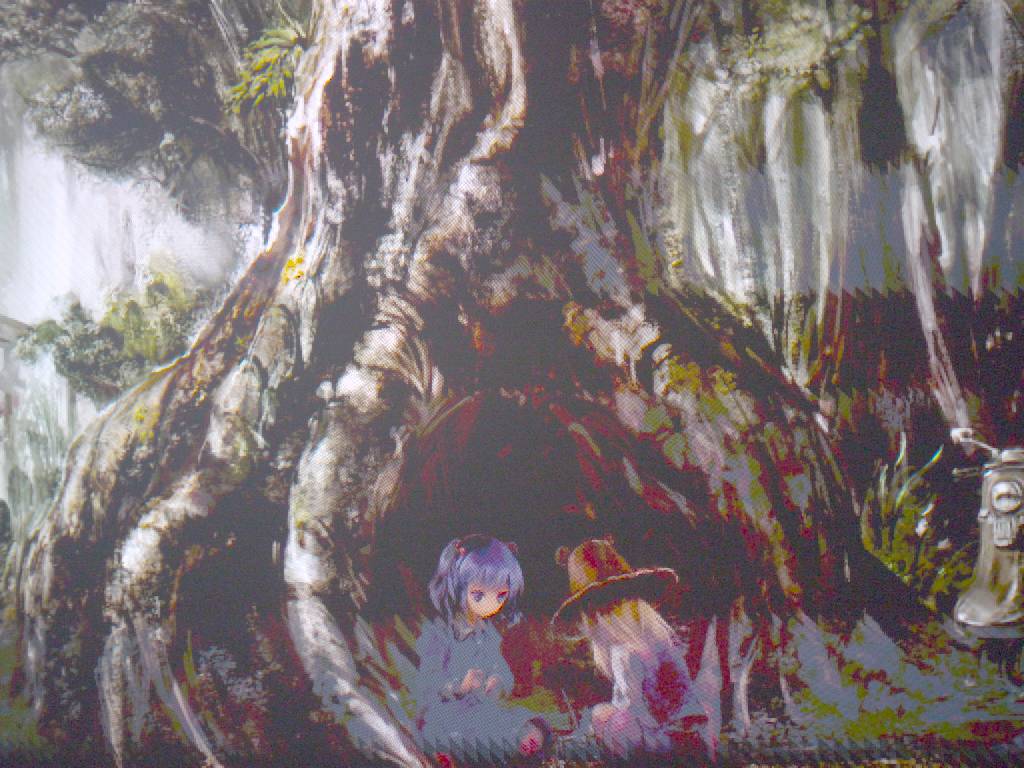}
    \end{subfigure}
    \begin{subfigure}{0.155\linewidth}
        \centering
        \includegraphics[width=\linewidth]{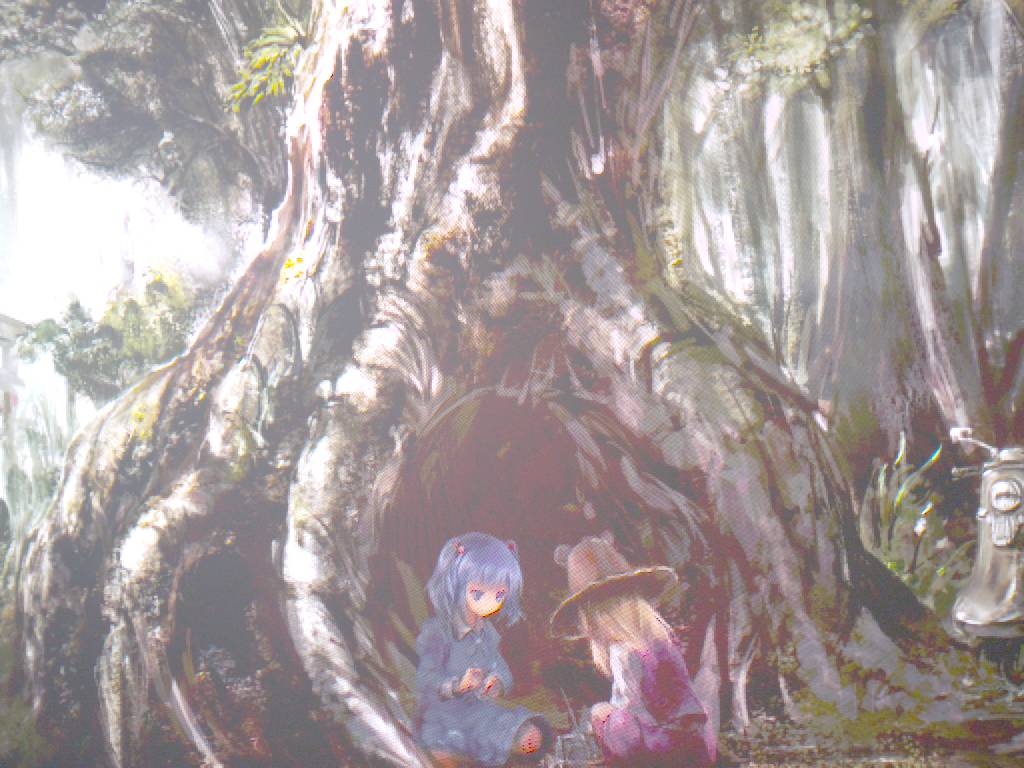}
    \end{subfigure}
    \begin{subfigure}{0.155\linewidth}
        \centering
        \includegraphics[width=\linewidth]{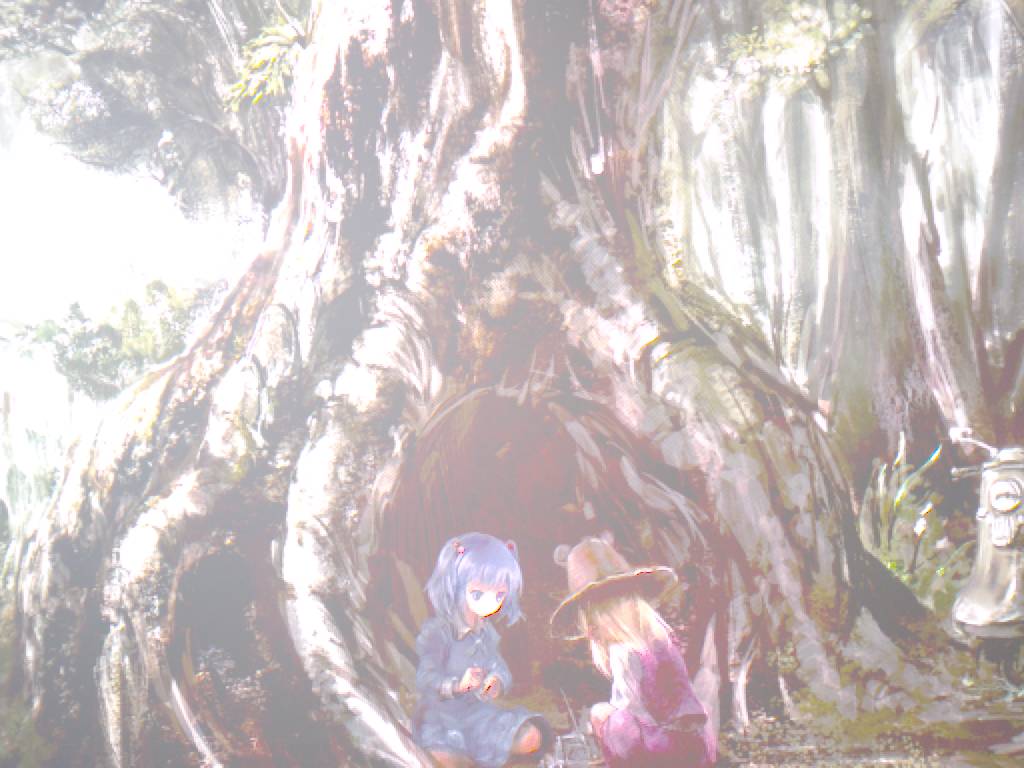}
    \end{subfigure}
    \begin{subfigure}{0.155\linewidth}
        \centering
        \includegraphics[width=\linewidth]{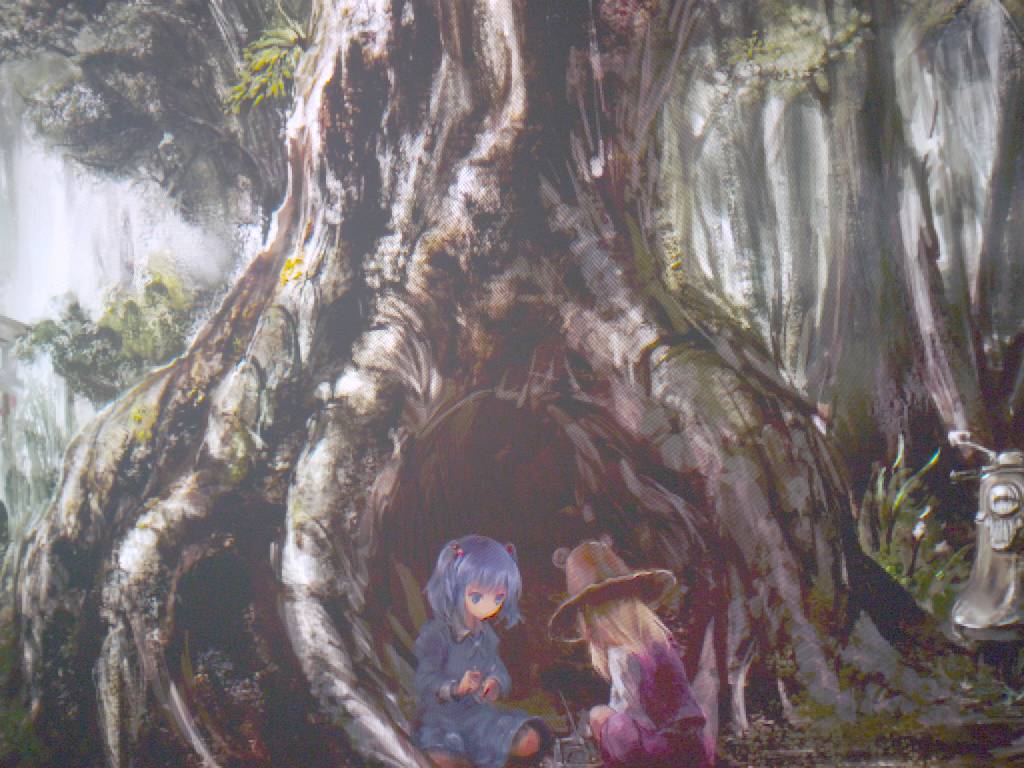}
    \end{subfigure}
    \caption{Bricker-Real dataset visualization. Each row shows a different scene captured at five exposure levels ($1/16\times$, $1/4\times$, $1\times$, $4\times$, $16\times$ relative to the reference), followed by the ground truth.}
    \label{fig:bricker_dataset_vis}
\end{figure}

\begin{figure}[h]
    \centering
    \begin{subfigure}{0.155\linewidth}
        \centering
        \includegraphics[width=\linewidth]{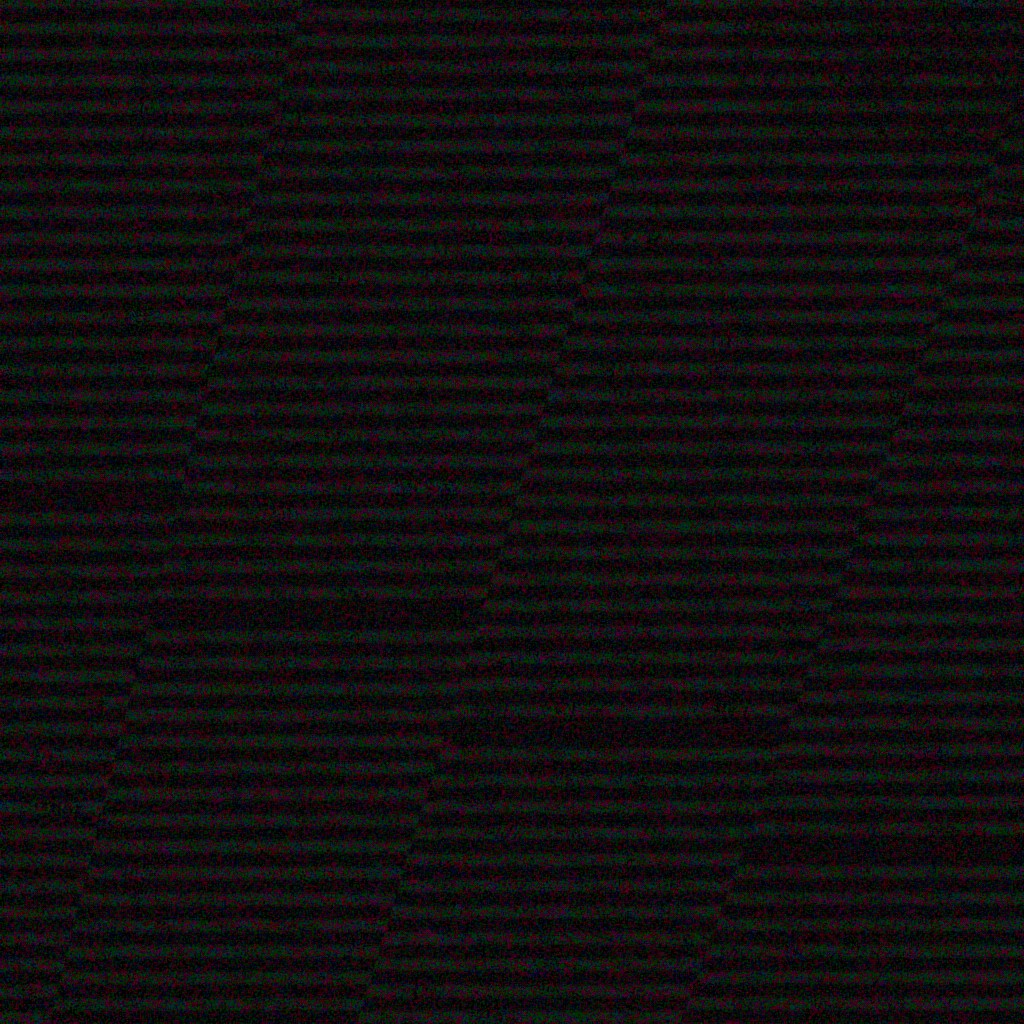}
    \end{subfigure}
    \begin{subfigure}{0.155\linewidth}
        \centering
        \includegraphics[width=\linewidth]{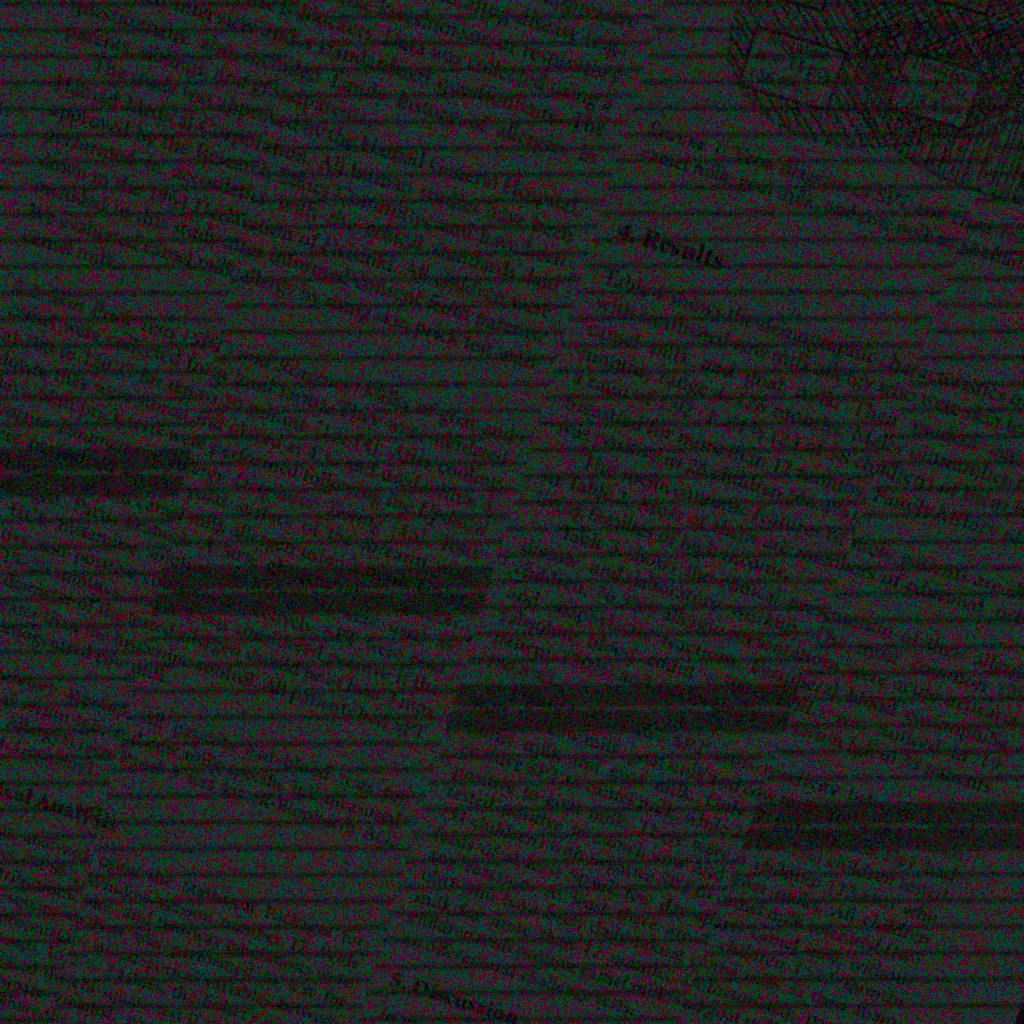}
    \end{subfigure}
    \begin{subfigure}{0.155\linewidth}
        \centering
        \includegraphics[width=\linewidth]{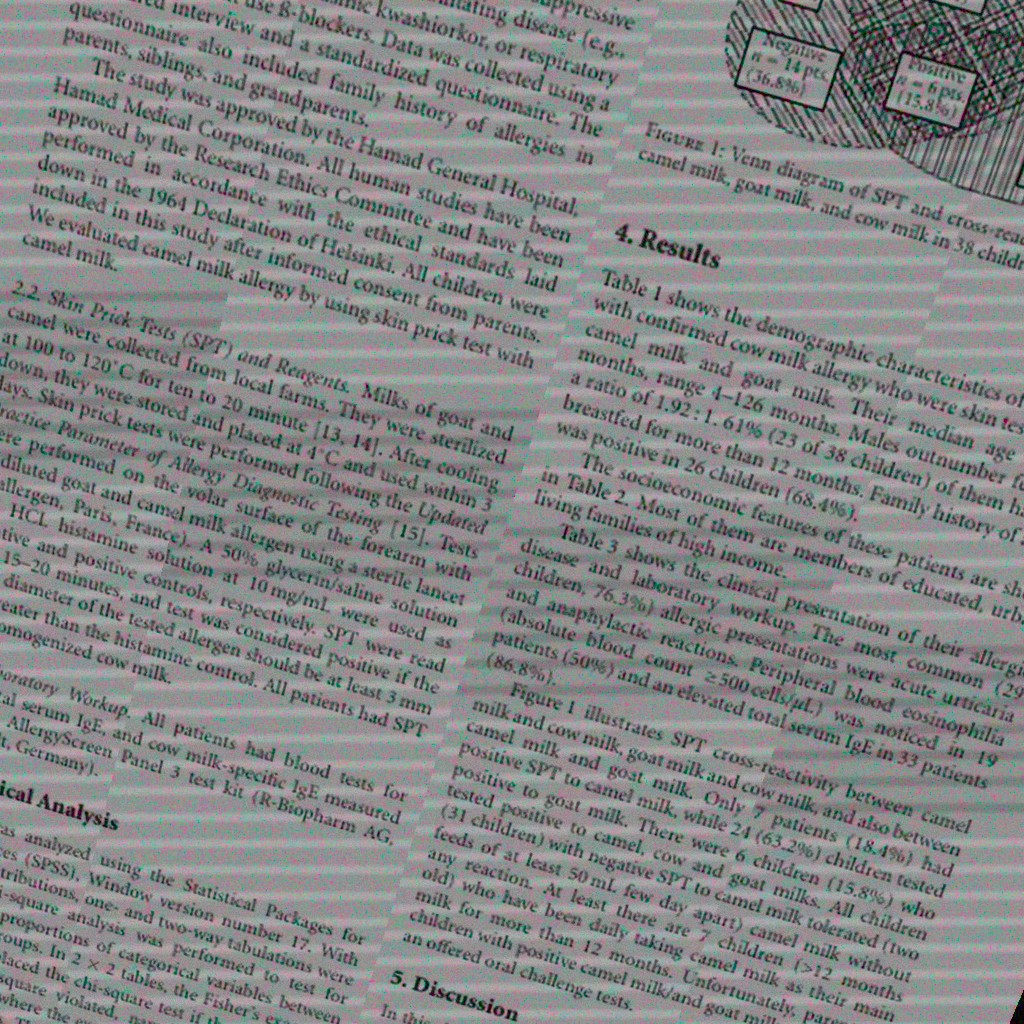}
    \end{subfigure}
    \begin{subfigure}{0.155\linewidth}
        \centering
        \includegraphics[width=\linewidth]{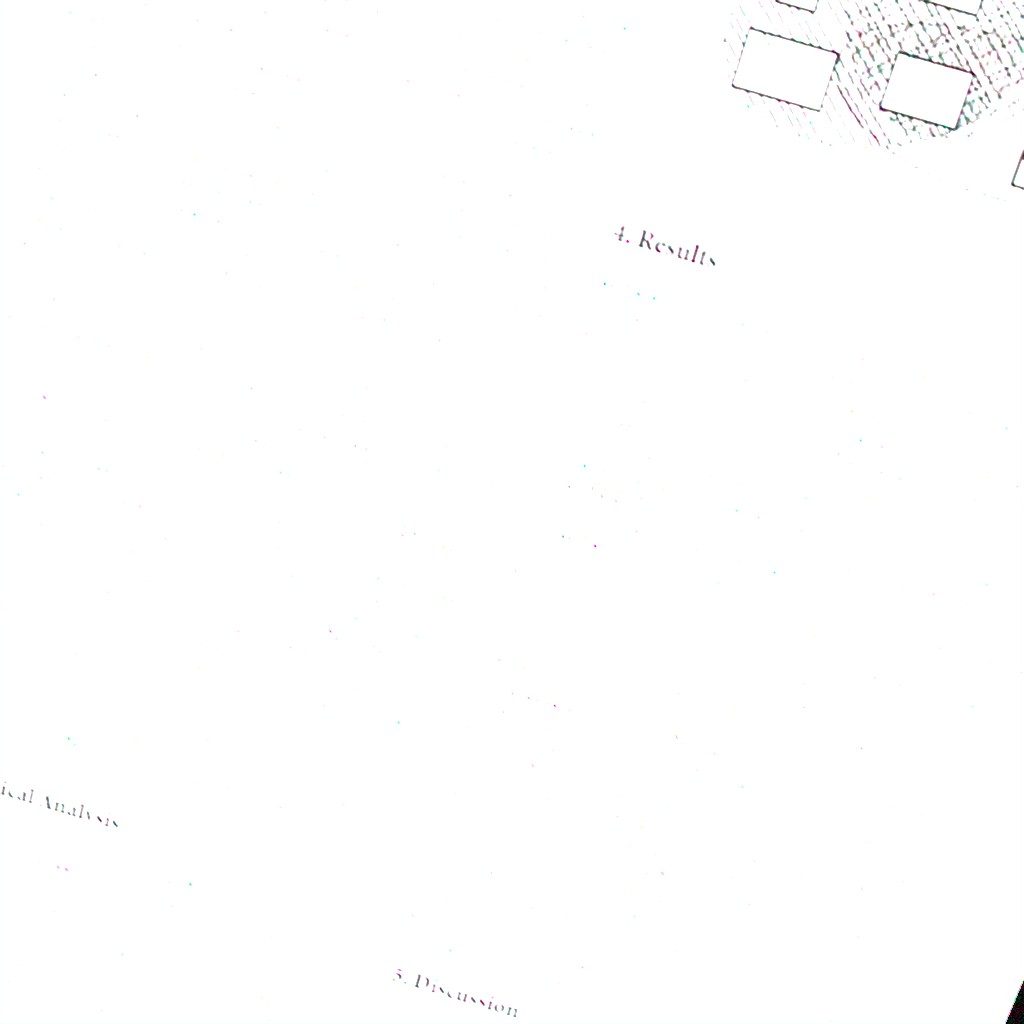}
    \end{subfigure}
    \begin{subfigure}{0.155\linewidth}
        \centering
        \includegraphics[width=\linewidth]{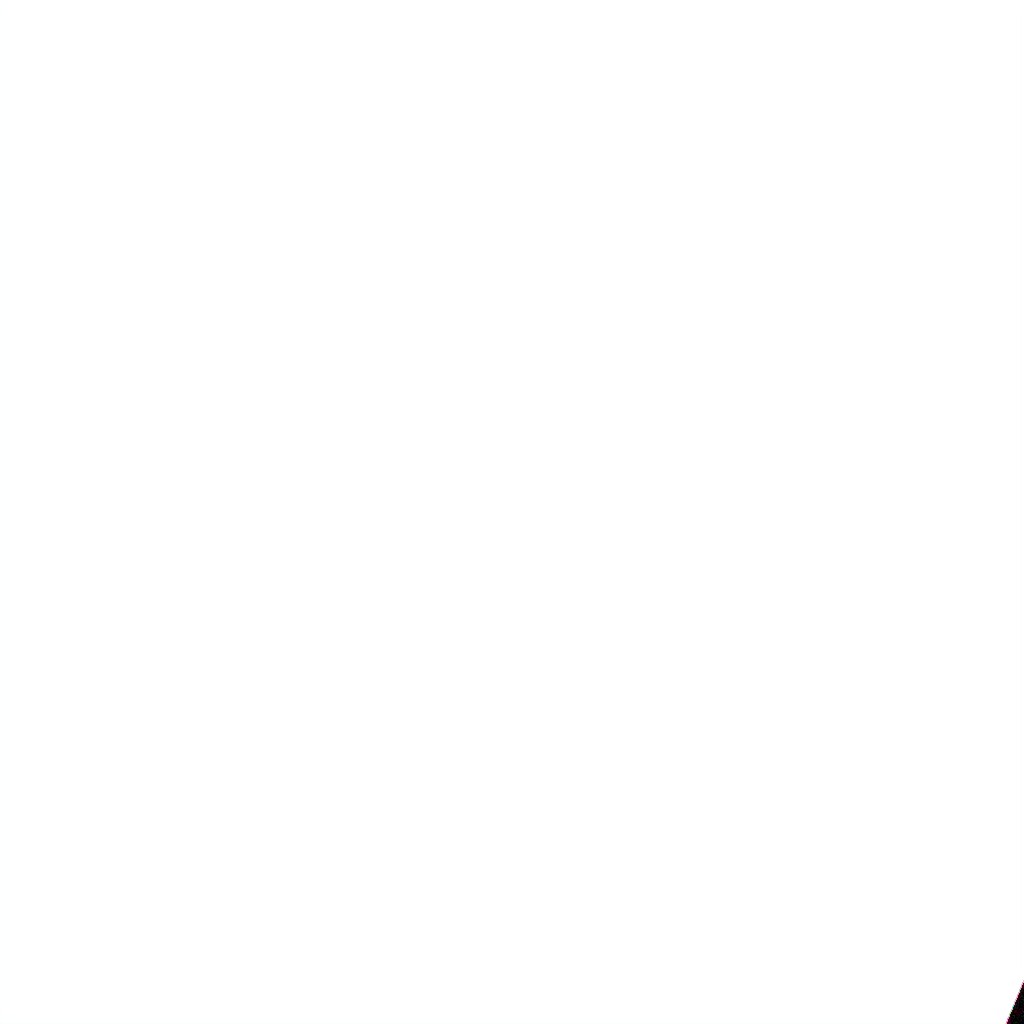}
    \end{subfigure}
    \begin{subfigure}{0.155\linewidth}
        \centering
        \includegraphics[width=\linewidth]{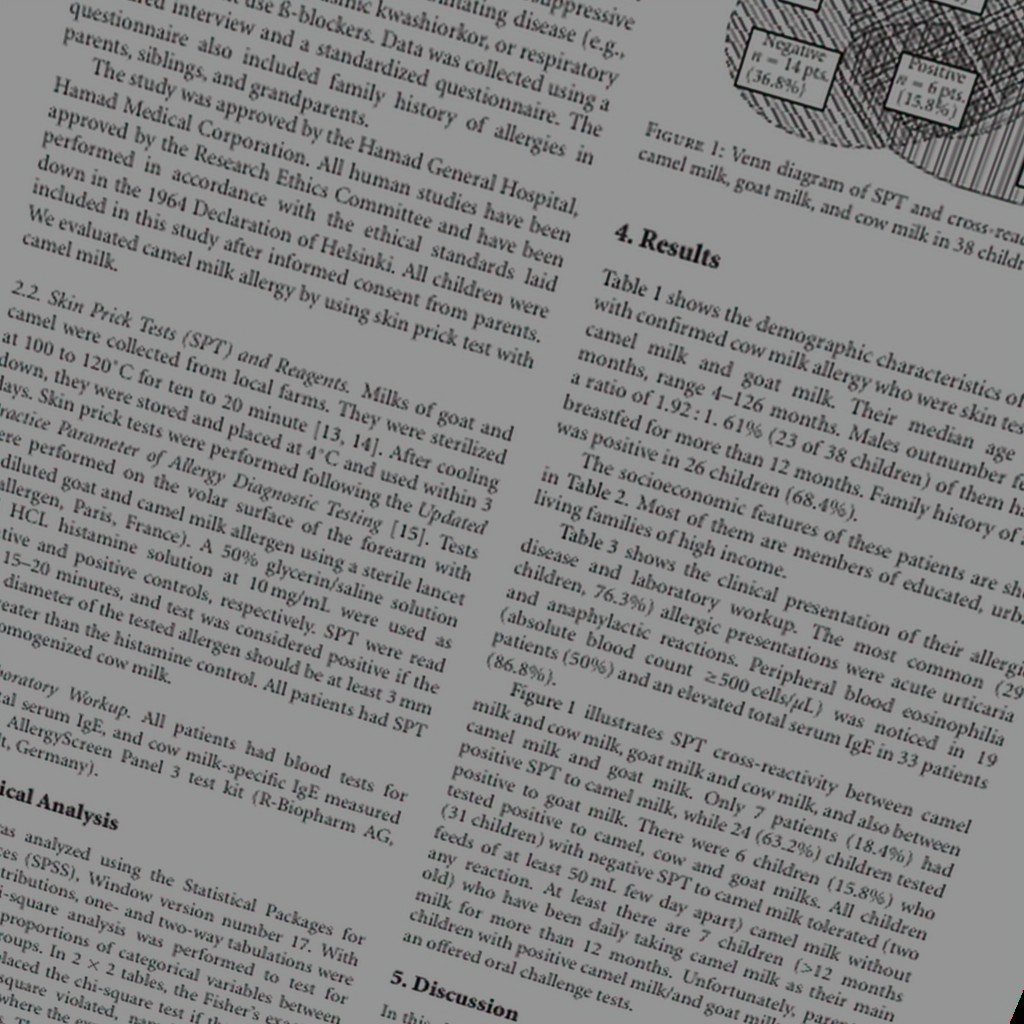}
    \end{subfigure}

    \begin{subfigure}{0.155\linewidth}
        \centering
        \includegraphics[width=\linewidth]{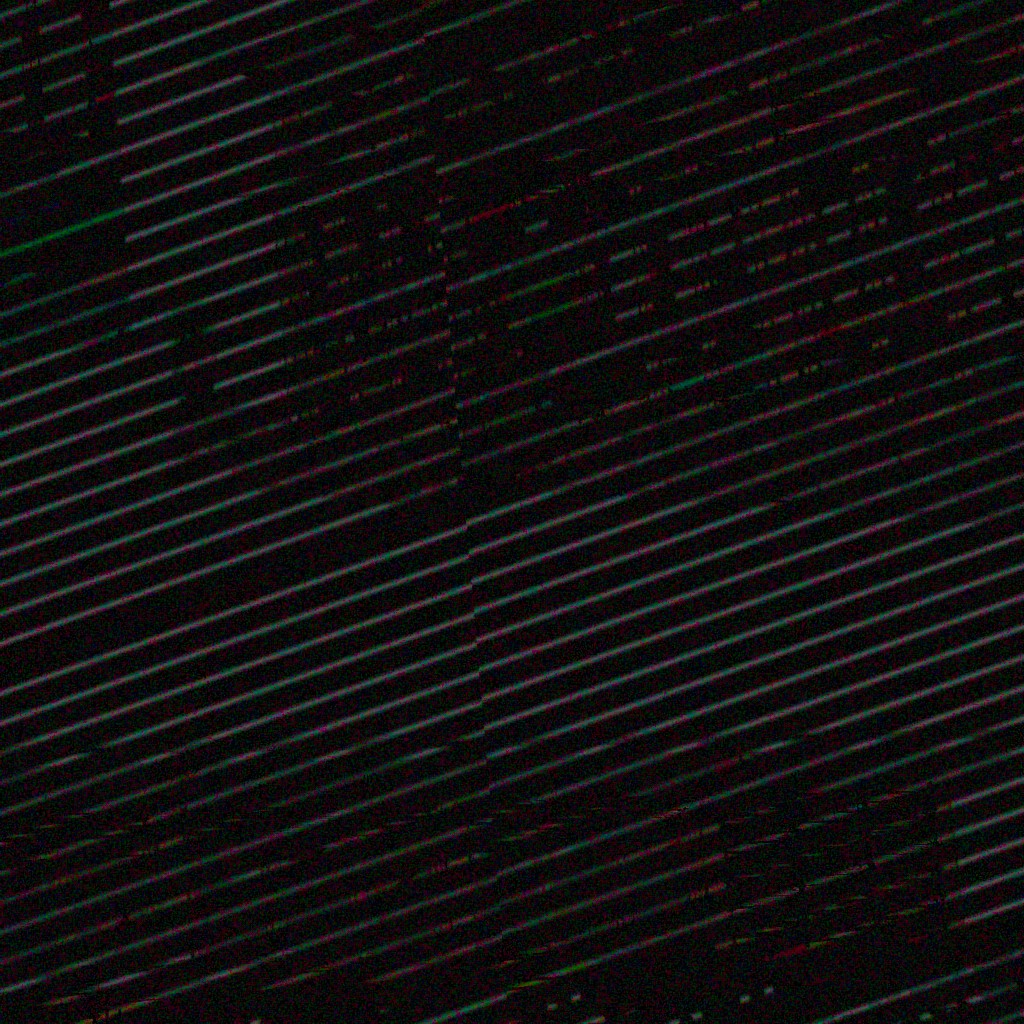}
    \end{subfigure}
    \begin{subfigure}{0.155\linewidth}
        \centering
        \includegraphics[width=\linewidth]{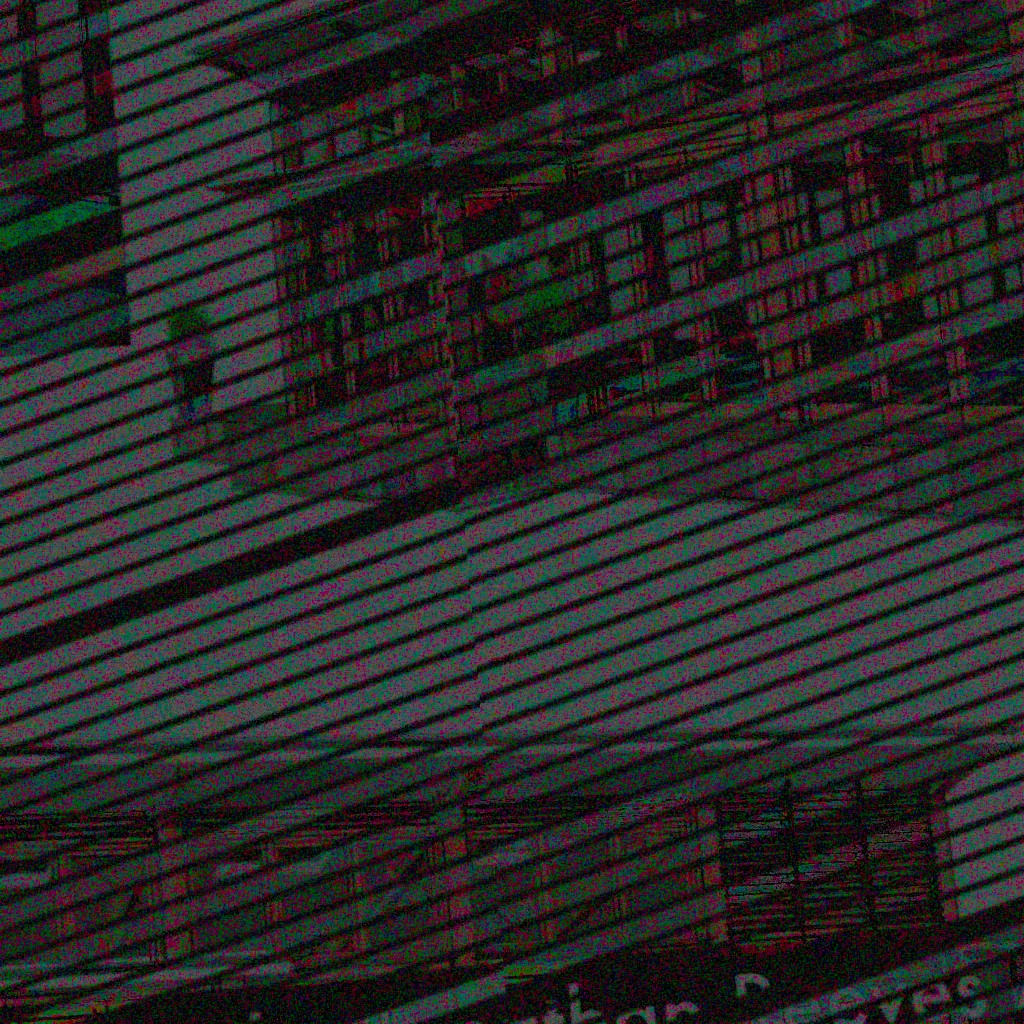}
    \end{subfigure}
    \begin{subfigure}{0.155\linewidth}
        \centering
        \includegraphics[width=\linewidth]{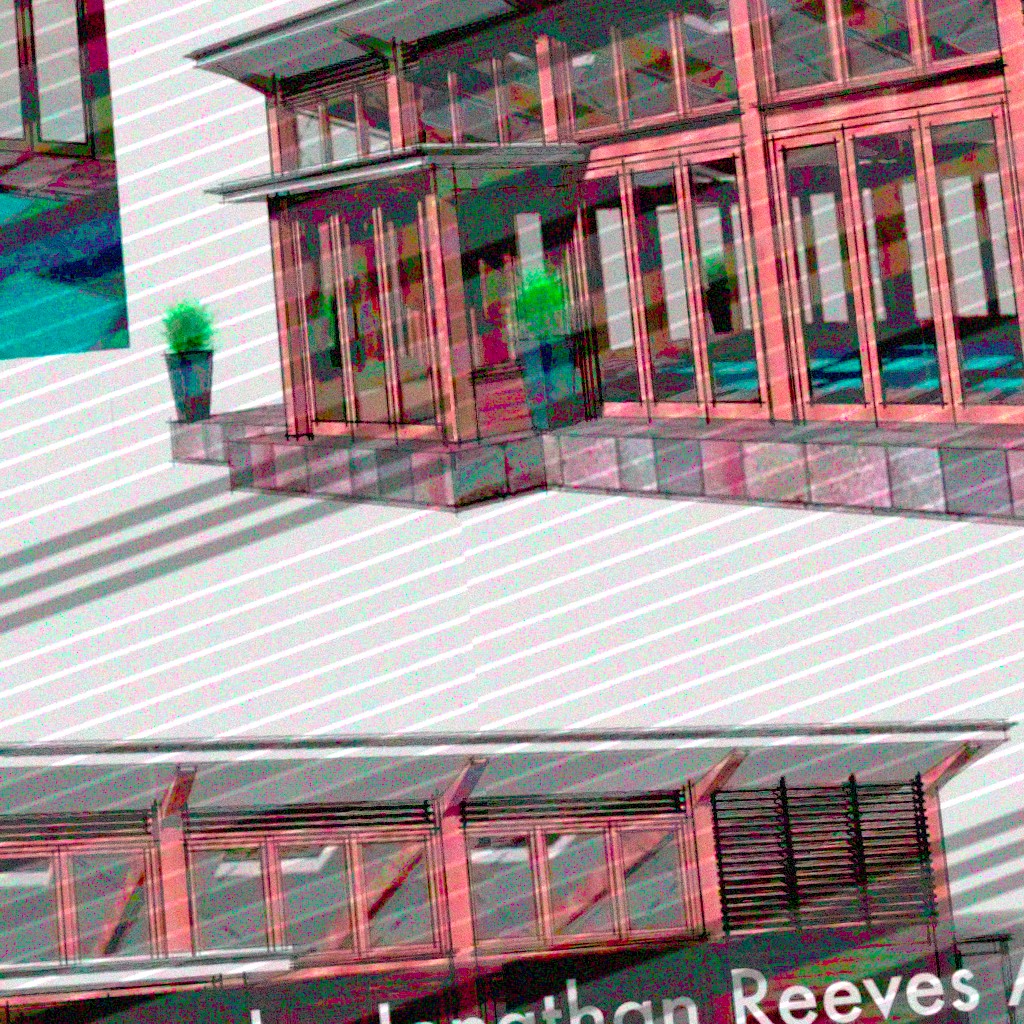}
    \end{subfigure}
    \begin{subfigure}{0.155\linewidth}
        \centering
        \includegraphics[width=\linewidth]{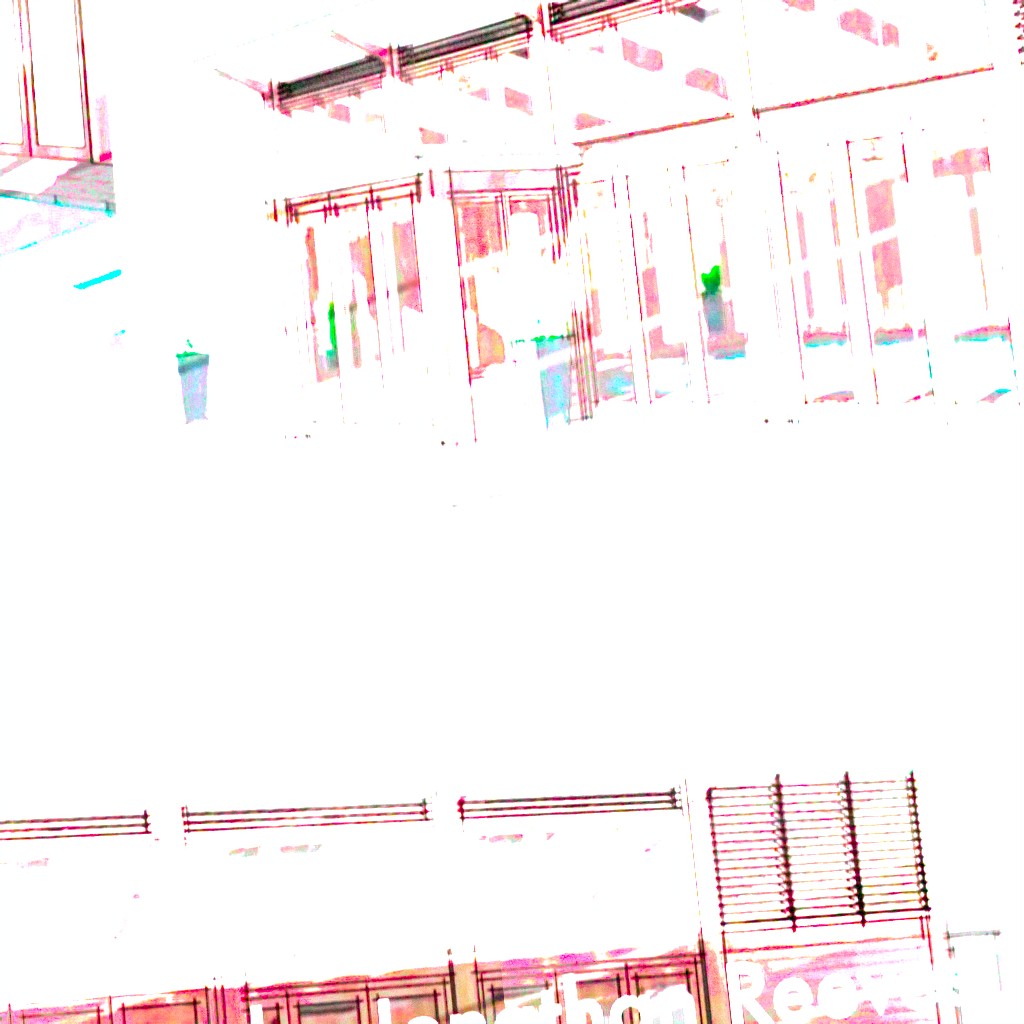}
    \end{subfigure}
    \begin{subfigure}{0.155\linewidth}
        \centering
        \includegraphics[width=\linewidth]{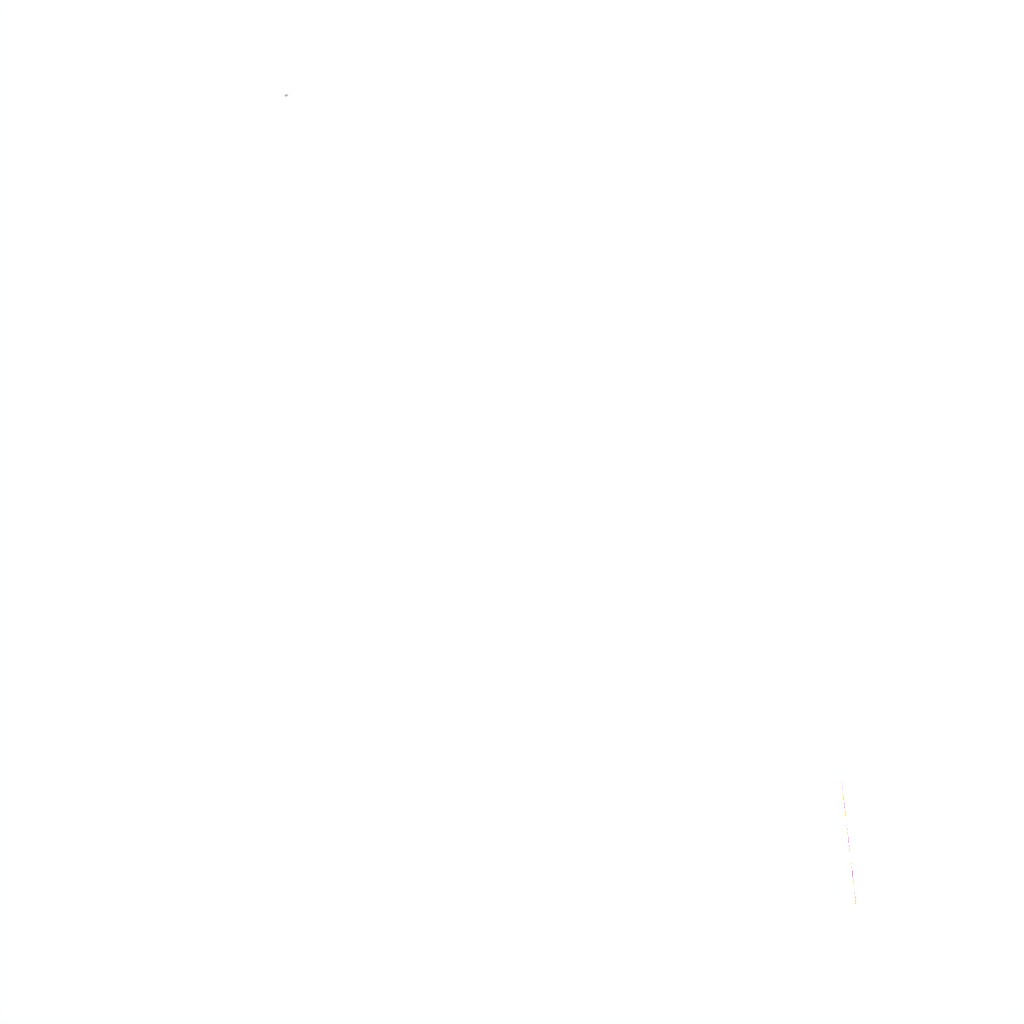}
    \end{subfigure}
    \begin{subfigure}{0.155\linewidth}
        \centering
        \includegraphics[width=\linewidth]{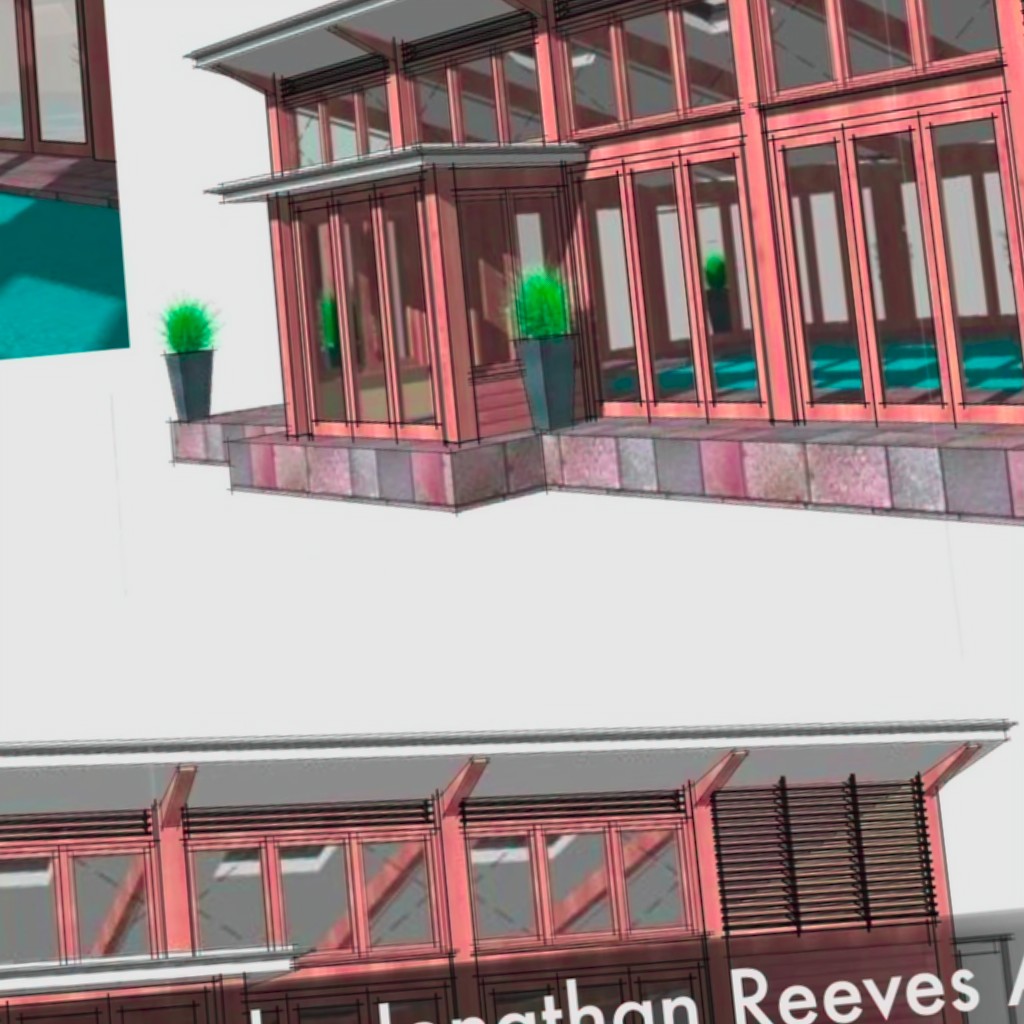}
    \end{subfigure}

    \begin{subfigure}{0.155\linewidth}
        \centering
        \includegraphics[width=\linewidth]{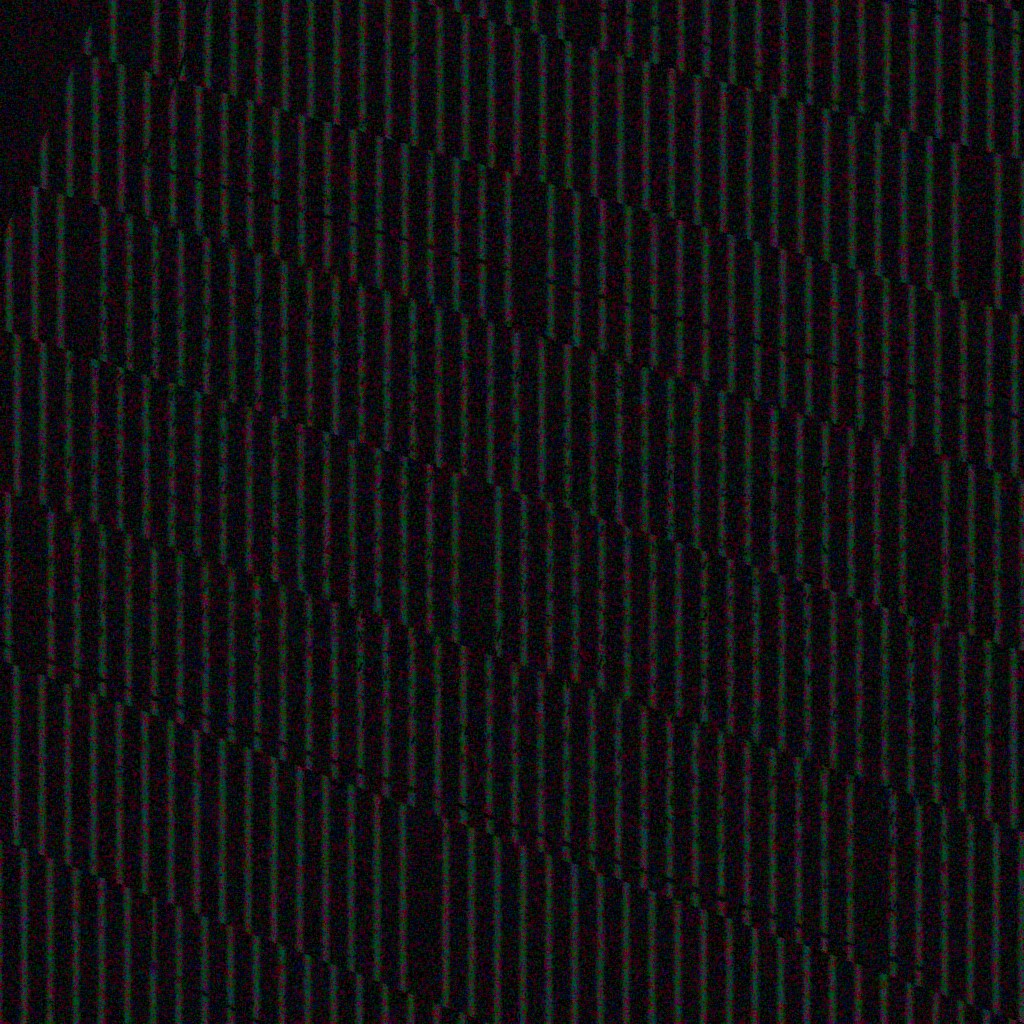}
    \end{subfigure}
    \begin{subfigure}{0.155\linewidth}
        \centering
        \includegraphics[width=\linewidth]{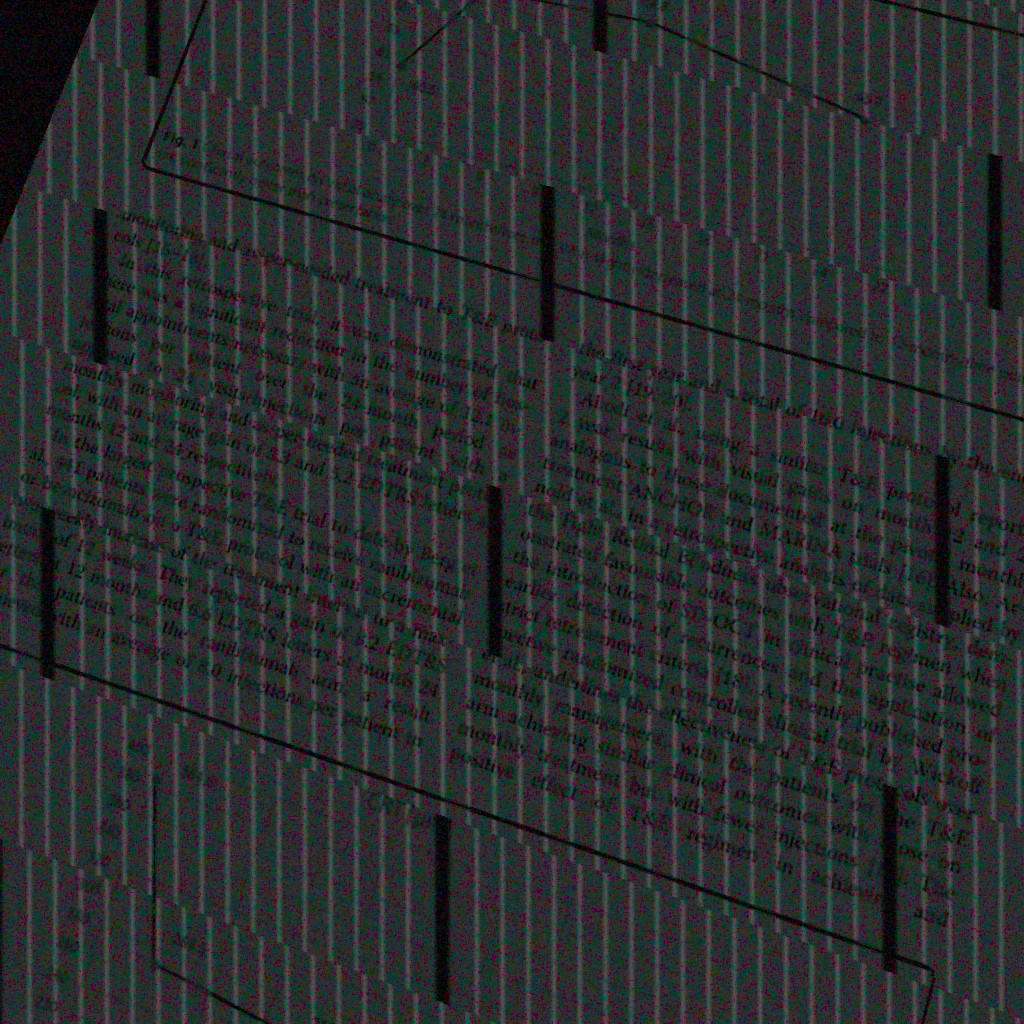}
    \end{subfigure}
    \begin{subfigure}{0.155\linewidth}
        \centering
        \includegraphics[width=\linewidth]{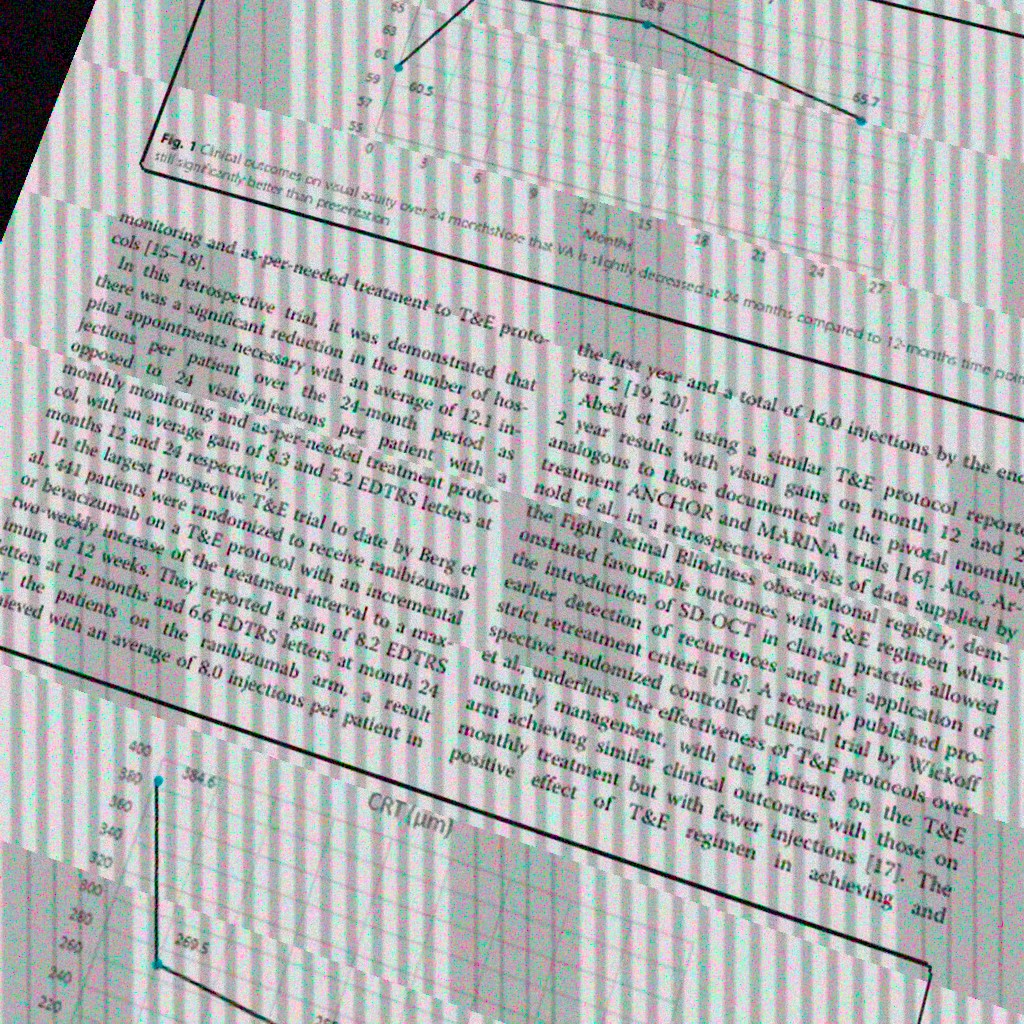}
    \end{subfigure}
    \begin{subfigure}{0.155\linewidth}
        \centering
        \includegraphics[width=\linewidth]{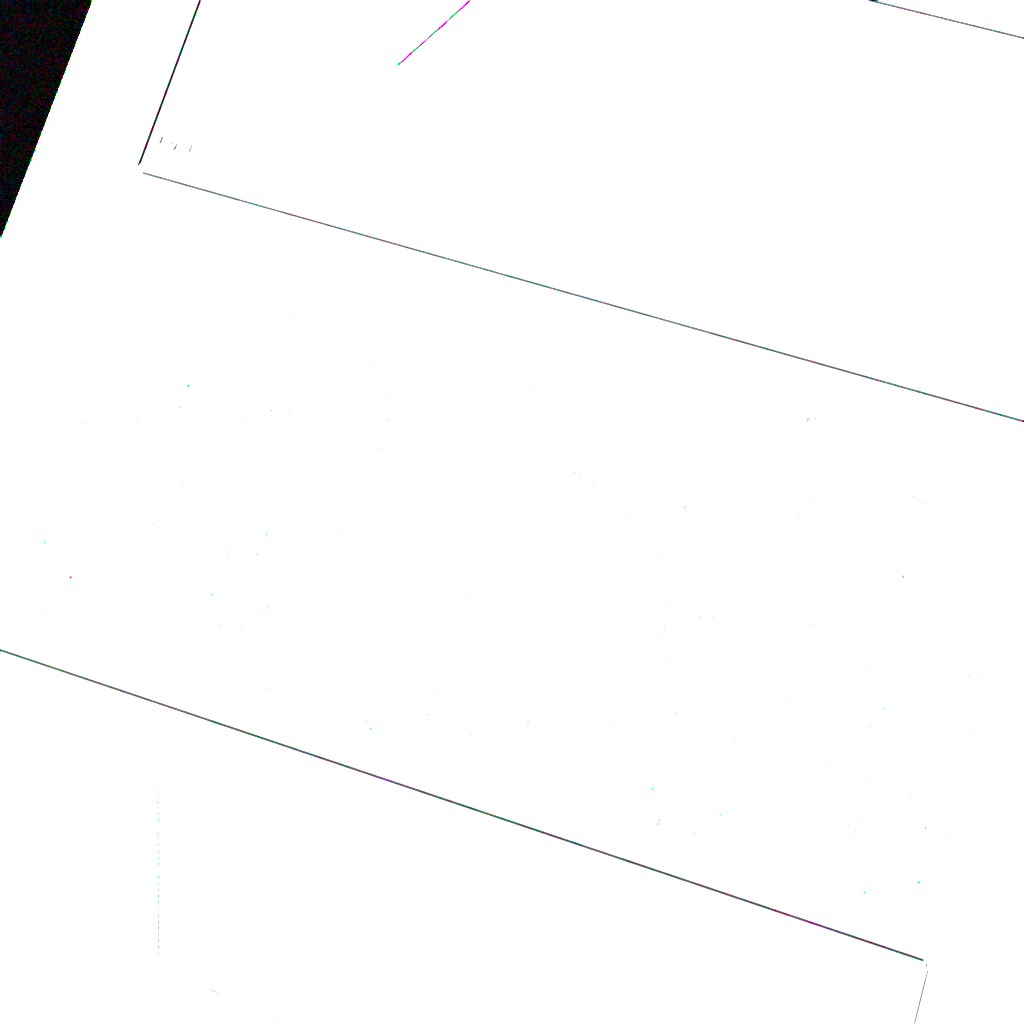}
    \end{subfigure}
    \begin{subfigure}{0.155\linewidth}
        \centering
        \includegraphics[width=\linewidth]{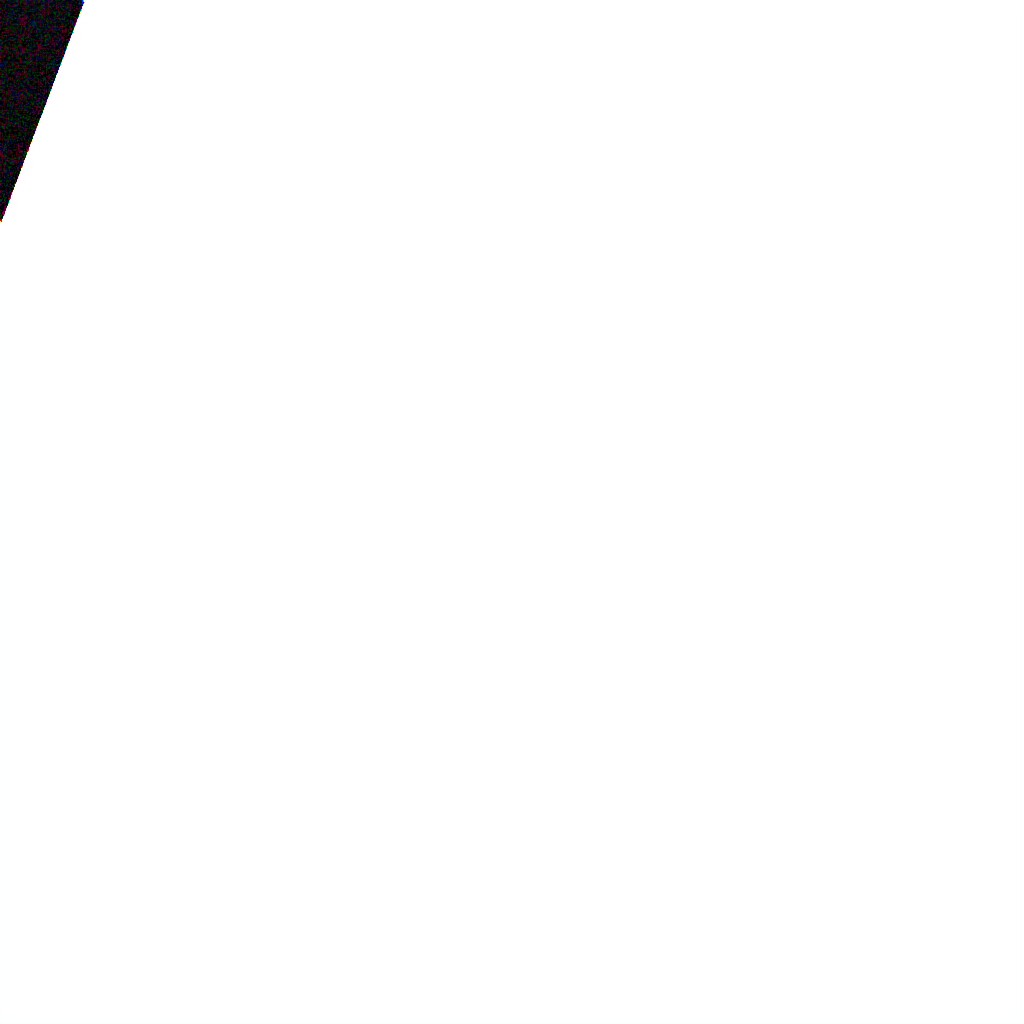}
    \end{subfigure}
    \begin{subfigure}{0.155\linewidth}
        \centering
        \includegraphics[width=\linewidth]{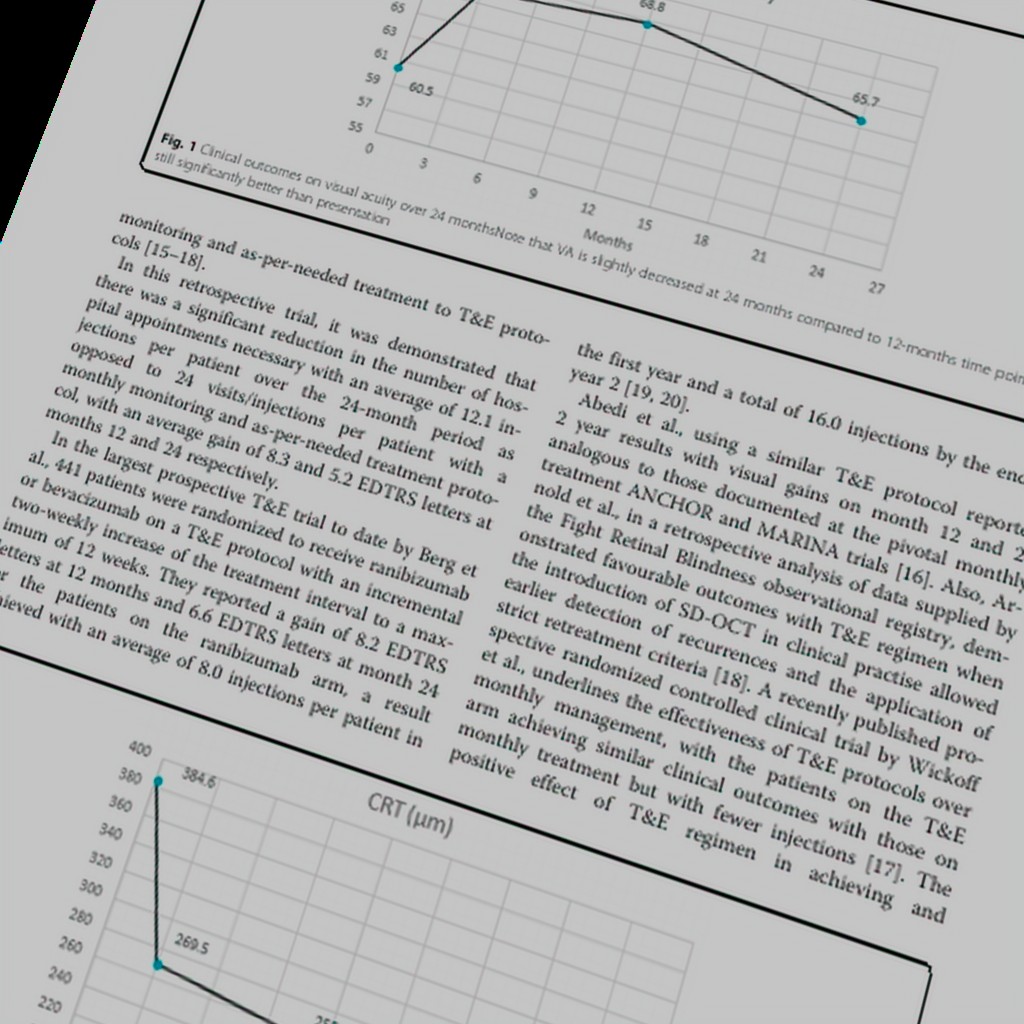}
    \end{subfigure}
    \caption{Bricker-Syn dataset visualization. Each row shows a different scene rendered at five exposure levels ($1/16\times$, $1/4\times$, $1\times$, $4\times$, $16\times$ relative to the reference), followed by the ground truth.}
    \label{fig:bricker_syn_vis}
\end{figure}

Figures~\ref{fig:bricker_dataset_vis} and~\ref{fig:bricker_syn_vis} visualize our Bricker dataset.

\section{Limitations and Areas for Improvement}
\label{sec:app_limitations}

Despite the promising results achieved by BRACE, several limitations remain open for future investigation.

\textbf{Handling of Extremely Complex FB Patterns.}
As analyzed in Sec.~\ref{sec:problem}, FB can exhibit highly diverse morphologies including color shift, compound periodicities, and jagged geometries, often appearing simultaneously in real captures. While our model effectively suppresses most of these artifacts, cases with severe compound-jagged coupling or extreme color distortions across large brightness ranges can still lead to residual banding or local color inaccuracies. These challenging scenarios warrant further exploration, such as incorporating explicit banding phase estimation or stronger temporal reasoning across a longer exposure sequence.

\textbf{Device-Dependent Color Bias.}
Our reverse ISP pipeline is parameterized by device-specific white-balance gains, color correction matrices, and other EXIF metadata (Appendix~\ref{sec:app_exif}). When the model is applied to a camera device with substantially different sensor characteristics or ISP parameters without fine-tuning, residual color shifts may appear in the restored output. This limitation is particularly evident when comparing results across different smartphone models. Future work could explore device-agnostic RAW representations, self-supervised domain adaptation~\cite{Zhang2025BracketIRE}, or incorporating explicit ISP parameter estimation to improve cross-device generalization.

\textbf{Computational Cost and Real-Time Deployment.}
Although BRACE is more efficient than diffusion-based alternatives~\cite{Zhu2025RIFLE,Zhu2026CLEAR}, the multi-frame processing pipeline still incurs non-negligible computational overhead. Optimizing the model architecture for on-device deployment, e.g., through model pruning, knowledge distillation, or efficient attention mechanisms, is a practical direction for future work.

We hope to address these limitations in future research by extending our dataset with more diverse device coverages, developing more robust banding representations, and exploring lightweight architectures suitable for mobile deployment.

\section{Broader Impacts}\label{sec:broader}

Our FB restoration model brings clear societal benefits by enhancing image quality in both everyday photography and professional workflows. It improves visual clarity under challenging conditions, producing more faithful and pleasing images. We mainly focus on the technical contribution of our work, and we do not foresee any direct societal impact of our work.

We encourage transparency about the technology’s capabilities and limitations to prevent unintended harms. Overall, we believe the positive applications outweigh potential negatives when appropriate safeguards are in place.

\newpage
\clearpage
\section*{NeurIPS Paper Checklist}

\begin{enumerate}

\item {\bf Claims}
    \item[] Question: Do the main claims made in the abstract and introduction accurately reflect the paper's contributions and scope?
    \item[] Answer: \answerYes{} 
    \item[] Justification: All the claims in the abstract and the introduction precisely reflect the paper's contribution and are supported by the experiments.
    \item[] Guidelines:
    \begin{itemize}
        \item The answer \answerNA{} means that the abstract and introduction do not include the claims made in the paper.
        \item The abstract and/or introduction should clearly state the claims made, including the contributions made in the paper and important assumptions and limitations. A \answerNo{} or \answerNA{} answer to this question will not be perceived well by the reviewers. 
        \item The claims made should match theoretical and experimental results, and reflect how much the results can be expected to generalize to other settings. 
        \item It is fine to include aspirational goals as motivation as long as it is clear that these goals are not attained by the paper. 
    \end{itemize}

\item {\bf Limitations}
    \item[] Question: Does the paper discuss the limitations of the work performed by the authors?
    \item[] Answer: \answerYes{} 
    \item[] Justification: We discuss the limitations of our work in the Appendix, Section~\ref{sec:app_limitations}.
    \item[] Guidelines:
    \begin{itemize}
        \item The answer \answerNA{} means that the paper has no limitation while the answer \answerNo{} means that the paper has limitations, but those are not discussed in the paper. 
        \item The authors are encouraged to create a separate ``Limitations'' section in their paper.
        \item The paper should point out any strong assumptions and how robust the results are to violations of these assumptions (e.g., independence assumptions, noiseless settings, model well-specification, asymptotic approximations only holding locally). The authors should reflect on how these assumptions might be violated in practice and what the implications would be.
        \item The authors should reflect on the scope of the claims made, e.g., if the approach was only tested on a few datasets or with a few runs. In general, empirical results often depend on implicit assumptions, which should be articulated.
        \item The authors should reflect on the factors that influence the performance of the approach. For example, a facial recognition algorithm may perform poorly when image resolution is low or images are taken in low lighting. Or a speech-to-text system might not be used reliably to provide closed captions for online lectures because it fails to handle technical jargon.
        \item The authors should discuss the computational efficiency of the proposed algorithms and how they scale with dataset size.
        \item If applicable, the authors should discuss possible limitations of their approach to address problems of privacy and fairness.
        \item While the authors might fear that complete honesty about limitations might be used by reviewers as grounds for rejection, a worse outcome might be that reviewers discover limitations that aren't acknowledged in the paper. The authors should use their best judgment and recognize that individual actions in favor of transparency play an important role in developing norms that preserve the integrity of the community. Reviewers will be specifically instructed to not penalize honesty concerning limitations.
    \end{itemize}

\item {\bf Theory assumptions and proofs}
    \item[] Question: For each theoretical result, does the paper provide the full set of assumptions and a complete (and correct) proof?
    \item[] Answer: \answerNA{} 
    \item[] Justification: The paper does not include theoretical results; the analysis of FB formation is descriptive and supported by references.
    \item[] Guidelines:
    \begin{itemize}
        \item The answer \answerNA{} means that the paper does not include theoretical results. 
        \item All the theorems, formulas, and proofs in the paper should be numbered and cross-referenced.
        \item All assumptions should be clearly stated or referenced in the statement of any theorems.
        \item The proofs can either appear in the main paper or the supplemental material, but if they appear in the supplemental material, the authors are encouraged to provide a short proof sketch to provide intuition. 
        \item Inversely, any informal proof provided in the core of the paper should be complemented by formal proofs provided in appendix or supplemental material.
        \item Theorems and Lemmas that the proof relies upon should be properly referenced. 
    \end{itemize}

    \item {\bf Experimental result reproducibility}
    \item[] Question: Does the paper fully disclose all the information needed to reproduce the main experimental results of the paper to the extent that it affects the main claims and/or conclusions of the paper (regardless of whether the code and data are provided or not)?
    \item[] Answer: \answerYes{} 
    \item[] Justification: We have provided experiment details in the implementation details section and appendix.
    \item[] Guidelines:
    \begin{itemize}
        \item The answer \answerNA{} means that the paper does not include experiments.
        \item If the paper includes experiments, a \answerNo{} answer to this question will not be perceived well by the reviewers: Making the paper reproducible is important, regardless of whether the code and data are provided or not.
        \item If the contribution is a dataset and\slash or model, the authors should describe the steps taken to make their results reproducible or verifiable. 
        \item Depending on the contribution, reproducibility can be accomplished in various ways. For example, if the contribution is a novel architecture, describing the architecture fully might suffice, or if the contribution is a specific model and empirical evaluation, it may be necessary to either make it possible for others to replicate the model with the same dataset, or provide access to the model. In general. releasing code and data is often one good way to accomplish this, but reproducibility can also be provided via detailed instructions for how to replicate the results, access to a hosted model (e.g., in the case of a large language model), releasing of a model checkpoint, or other means that are appropriate to the research performed.
        \item While NeurIPS does not require releasing code, the conference does require all submissions to provide some reasonable avenue for reproducibility, which may depend on the nature of the contribution. For example
        \begin{enumerate}
            \item If the contribution is primarily a new algorithm, the paper should make it clear how to reproduce that algorithm.
            \item If the contribution is primarily a new model architecture, the paper should describe the architecture clearly and fully.
            \item If the contribution is a new model (e.g., a large language model), then there should either be a way to access this model for reproducing the results or a way to reproduce the model (e.g., with an open-source dataset or instructions for how to construct the dataset).
            \item We recognize that reproducibility may be tricky in some cases, in which case authors are welcome to describe the particular way they provide for reproducibility. In the case of closed-source models, it may be that access to the model is limited in some way (e.g., to registered users), but it should be possible for other researchers to have some path to reproducing or verifying the results.
        \end{enumerate}
    \end{itemize}

\item {\bf Open access to data and code}
    \item[] Question: Does the paper provide open access to the data and code, with sufficient instructions to faithfully reproduce the main experimental results, as described in supplemental material?
    \item[] Answer: \answerYes{} 
    \item[] Justification: Our code and dataset are available at \url{https://github.com/ZZH-qwq/BRACE}.
    \item[] Guidelines:
    \begin{itemize}
        \item The answer \answerNA{} means that paper does not include experiments requiring code.
        \item Please see the NeurIPS code and data submission guidelines (\url{https://neurips.cc/public/guides/CodeSubmissionPolicy}) for more details.
        \item While we encourage the release of code and data, we understand that this might not be possible, so \answerNo{} is an acceptable answer. Papers cannot be rejected simply for not including code, unless this is central to the contribution (e.g., for a new open-source benchmark).
        \item The instructions should contain the exact command and environment needed to run to reproduce the results. See the NeurIPS code and data submission guidelines (\url{https://neurips.cc/public/guides/CodeSubmissionPolicy}) for more details.
        \item The authors should provide instructions on data access and preparation, including how to access the raw data, preprocessed data, intermediate data, and generated data, etc.
        \item The authors should provide scripts to reproduce all experimental results for the new proposed method and baselines. If only a subset of experiments are reproducible, they should state which ones are omitted from the script and why.
        \item At submission time, to preserve anonymity, the authors should release anonymized versions (if applicable).
        \item Providing as much information as possible in supplemental material (appended to the paper) is recommended, but including URLs to data and code is permitted.
    \end{itemize}

\item {\bf Experimental setting/details}
    \item[] Question: Does the paper specify all the training and test details (e.g., data splits, hyperparameters, how they were chosen, type of optimizer) necessary to understand the results?
    \item[] Answer: \answerYes{} 
    \item[] Justification: All training and test details (splits, patch size, optimizer, learning rates, epochs, loss weights) are given in Section~4.1, and per-method settings in Appendix~\ref{sec:app_implementation_details}.
    \item[] Guidelines:
    \begin{itemize}
        \item The answer \answerNA{} means that the paper does not include experiments.
        \item The experimental setting should be presented in the core of the paper to a level of detail that is necessary to appreciate the results and make sense of them.
        \item The full details can be provided either with the code, in appendix, or as supplemental material.
    \end{itemize}

\item {\bf Experiment statistical significance}
    \item[] Question: Does the paper report error bars suitably and correctly defined or other appropriate information about the statistical significance of the experiments?
    \item[] Answer: \answerNo{} 
    \item[] Justification: We have not reported error bars, but the main claims of the paper are supported as the overall performance on multiple metrics is consistently better than the compared methods.
    \item[] Guidelines:
    \begin{itemize}
        \item The answer \answerNA{} means that the paper does not include experiments.
        \item The authors should answer \answerYes{} if the results are accompanied by error bars, confidence intervals, or statistical significance tests, at least for the experiments that support the main claims of the paper.
        \item The factors of variability that the error bars are capturing should be clearly stated (for example, train/test split, initialization, random drawing of some parameter, or overall run with given experimental conditions).
        \item The method for calculating the error bars should be explained (closed form formula, call to a library function, bootstrap, etc.)
        \item The assumptions made should be given (e.g., Normally distributed errors).
        \item It should be clear whether the error bar is the standard deviation or the standard error of the mean.
        \item It is OK to report 1-sigma error bars, but one should state it. The authors should preferably report a 2-sigma error bar than state that they have a 96\% CI, if the hypothesis of Normality of errors is not verified.
        \item For asymmetric distributions, the authors should be careful not to show in tables or figures symmetric error bars that would yield results that are out of range (e.g., negative error rates).
        \item If error bars are reported in tables or plots, the authors should explain in the text how they were calculated and reference the corresponding figures or tables in the text.
    \end{itemize}

\item {\bf Experiments compute resources}
    \item[] Question: For each experiment, does the paper provide sufficient information on the computer resources (type of compute workers, memory, time of execution) needed to reproduce the experiments?
    \item[] Answer: \answerYes{} 
    \item[] Justification: We have provided the compute resources used for training in the implementation details section.
    \item[] Guidelines:
    \begin{itemize}
        \item The answer \answerNA{} means that the paper does not include experiments.
        \item The paper should indicate the type of compute workers CPU or GPU, internal cluster, or cloud provider, including relevant memory and storage.
        \item The paper should provide the amount of compute required for each of the individual experimental runs as well as estimate the total compute. 
        \item The paper should disclose whether the full research project required more compute than the experiments reported in the paper (e.g., preliminary or failed experiments that didn't make it into the paper). 
    \end{itemize}
    
\item {\bf Code of ethics}
    \item[] Question: Does the research conducted in the paper conform, in every respect, with the NeurIPS Code of Ethics \url{https://neurips.cc/public/EthicsGuidelines}?
    \item[] Answer: \answerYes{} 
    \item[] Justification: Our research fully complies with the NeurIPS Code of Ethics in all respects.
    \item[] Guidelines:
    \begin{itemize}
        \item The answer \answerNA{} means that the authors have not reviewed the NeurIPS Code of Ethics.
        \item If the authors answer \answerNo, they should explain the special circumstances that require a deviation from the Code of Ethics.
        \item The authors should make sure to preserve anonymity (e.g., if there is a special consideration due to laws or regulations in their jurisdiction).
    \end{itemize}

\item {\bf Broader impacts}
    \item[] Question: Does the paper discuss both potential positive societal impacts and negative societal impacts of the work performed?
    \item[] Answer: \answerYes{} 
    \item[] Justification: We have discussed the potential societal impacts of our work in the appendix. We mainly focus on the technical contribution of our work, and we do not foresee any direct societal impact of our work.
    \item[] Guidelines:
    \begin{itemize}
        \item The answer \answerNA{} means that there is no societal impact of the work performed.
        \item If the authors answer \answerNA{} or \answerNo, they should explain why their work has no societal impact or why the paper does not address societal impact.
        \item Examples of negative societal impacts include potential malicious or unintended uses (e.g., disinformation, generating fake profiles, surveillance), fairness considerations (e.g., deployment of technologies that could make decisions that unfairly impact specific groups), privacy considerations, and security considerations.
        \item The conference expects that many papers will be foundational research and not tied to particular applications, let alone deployments. However, if there is a direct path to any negative applications, the authors should point it out. For example, it is legitimate to point out that an improvement in the quality of generative models could be used to generate Deepfakes for disinformation. On the other hand, it is not needed to point out that a generic algorithm for optimizing neural networks could enable people to train models that generate Deepfakes faster.
        \item The authors should consider possible harms that could arise when the technology is being used as intended and functioning correctly, harms that could arise when the technology is being used as intended but gives incorrect results, and harms following from (intentional or unintentional) misuse of the technology.
        \item If there are negative societal impacts, the authors could also discuss possible mitigation strategies (e.g., gated release of models, providing defenses in addition to attacks, mechanisms for monitoring misuse, mechanisms to monitor how a system learns from feedback over time, improving the efficiency and accessibility of ML).
    \end{itemize}
    
\item {\bf Safeguards}
    \item[] Question: Does the paper describe safeguards that have been put in place for responsible release of data or models that have a high risk for misuse (e.g., pre-trained language models, image generators, or scraped datasets)?
    \item[] Answer: \answerNA{} 
    \item[] Justification: The paper poses no such risks.
    \item[] Guidelines:
    \begin{itemize}
        \item The answer \answerNA{} means that the paper poses no such risks.
        \item Released models that have a high risk for misuse or dual-use should be released with necessary safeguards to allow for controlled use of the model, for example by requiring that users adhere to usage guidelines or restrictions to access the model or implementing safety filters. 
        \item Datasets that have been scraped from the Internet could pose safety risks. The authors should describe how they avoided releasing unsafe images.
        \item We recognize that providing effective safeguards is challenging, and many papers do not require this, but we encourage authors to take this into account and make a best faith effort.
    \end{itemize}

\item {\bf Licenses for existing assets}
    \item[] Question: Are the creators or original owners of assets (e.g., code, data, models), used in the paper, properly credited and are the license and terms of use explicitly mentioned and properly respected?
    \item[] Answer: \answerYes{} 
    \item[] Justification: We cited all the works we mentioned or used and fully respect their license.
    \item[] Guidelines:
    \begin{itemize}
        \item The answer \answerNA{} means that the paper does not use existing assets.
        \item The authors should cite the original paper that produced the code package or dataset.
        \item The authors should state which version of the asset is used and, if possible, include a URL.
        \item The name of the license (e.g., CC-BY 4.0) should be included for each asset.
        \item For scraped data from a particular source (e.g., website), the copyright and terms of service of that source should be provided.
        \item If assets are released, the license, copyright information, and terms of use in the package should be provided. For popular datasets, \url{paperswithcode.com/datasets} has curated licenses for some datasets. Their licensing guide can help determine the license of a dataset.
        \item For existing datasets that are re-packaged, both the original license and the license of the derived asset (if it has changed) should be provided.
        \item If this information is not available online, the authors are encouraged to reach out to the asset's creators.
    \end{itemize}

\item {\bf New assets}
    \item[] Question: Are new assets introduced in the paper well documented and is the documentation provided alongside the assets?
    \item[] Answer: \answerYes{} 
    \item[] Justification: Our new assets (dataset, code, and pre-trained models) are documented in the repository and available at \url{https://github.com/ZZH-qwq/BRACE}.
    \item[] Guidelines:
    \begin{itemize}
        \item The answer \answerNA{} means that the paper does not release new assets.
        \item Researchers should communicate the details of the dataset\slash code\slash model as part of their submissions via structured templates. This includes details about training, license, limitations, etc. 
        \item The paper should discuss whether and how consent was obtained from people whose asset is used.
        \item At submission time, remember to anonymize your assets (if applicable). You can either create an anonymized URL or include an anonymized zip file.
    \end{itemize}

\item {\bf Crowdsourcing and research with human subjects}
    \item[] Question: For crowdsourcing experiments and research with human subjects, does the paper include the full text of instructions given to participants and screenshots, if applicable, as well as details about compensation (if any)? 
    \item[] Answer: \answerNA{} 
    \item[] Justification: The paper does not involve crowdsourcing nor research with human subjects.
    \item[] Guidelines:
    \begin{itemize}
        \item The answer \answerNA{} means that the paper does not involve crowdsourcing nor research with human subjects.
        \item Including this information in the supplemental material is fine, but if the main contribution of the paper involves human subjects, then as much detail as possible should be included in the main paper. 
        \item According to the NeurIPS Code of Ethics, workers involved in data collection, curation, or other labor should be paid at least the minimum wage in the country of the data collector. 
    \end{itemize}

\item {\bf Institutional review board (IRB) approvals or equivalent for research with human subjects}
    \item[] Question: Does the paper describe potential risks incurred by study participants, whether such risks were disclosed to the subjects, and whether Institutional Review Board (IRB) approvals (or an equivalent approval/review based on the requirements of your country or institution) were obtained?
    \item[] Answer: \answerNA{} 
    \item[] Justification: The paper does not involve crowdsourcing nor research with human subjects.
    \item[] Guidelines:
    \begin{itemize}
        \item The answer \answerNA{} means that the paper does not involve crowdsourcing nor research with human subjects.
        \item Depending on the country in which research is conducted, IRB approval (or equivalent) may be required for any human subjects research. If you obtained IRB approval, you should clearly state this in the paper. 
        \item We recognize that the procedures for this may vary significantly between institutions and locations, and we expect authors to adhere to the NeurIPS Code of Ethics and the guidelines for their institution. 
        \item For initial submissions, do not include any information that would break anonymity (if applicable), such as the institution conducting the review.
    \end{itemize}

\item {\bf Declaration of LLM usage}
    \item[] Question: Does the paper describe the usage of LLMs if it is an important, original, or non-standard component of the core methods in this research? Note that if the LLM is used only for writing, editing, or formatting purposes and does \emph{not} impact the core methodology, scientific rigor, or originality of the research, declaration is not required.
    \item[] Answer: \answerNA{} 
    \item[] Justification: LLM is used only for writing and editing.
    \item[] Guidelines:
    \begin{itemize}
        \item The answer \answerNA{} means that the core method development in this research does not involve LLMs as any important, original, or non-standard components.
        \item Please refer to our LLM policy in the NeurIPS handbook for what should or should not be described.
    \end{itemize}

\end{enumerate}

\end{document}